\documentclass[sigconf,screen]{acmart}

\acmConference[]{}{}{}
\setcopyright{none}
\usepackage[ruled,linesnumbered,vlined]{algorithm2e}
\SetCommentSty{mycommfont}

\usepackage{paralist}
\usepackage{wrapfig}
\usepackage{xcolor}
\def\Snospace~{\S{}}

\usepackage{subcaption}
\graphicspath{{fig/}{./}}

\renewcommand{\algorithmcfname}{ALGORITHM}

\usepackage{mathrsfs}
\usepackage{amsthm}
\usepackage{dsfont}
\usepackage[flushleft,para]{threeparttable}
\usepackage{booktabs}
\usepackage{adjustbox}
\usepackage{bm}

\usepackage{rotating}
\newcommand*{\headformat}[1]{#1}

\newlength{\maxlen}
\newcommand*{\shortheadformat}[1]{#1}

\newlength{\shortmaxlen}
\newcommand*{\head}[1]{%
    \begin{sideways}
      \makebox[\maxlen][l]{\headformat{#1}}
    \end{sideways}}
    
\newcommand*{\shorthead}[1]{%
    \begin{sideways}
      \makebox[\shortmaxlen][l]{\shortheadformat{#1}}
    \end{sideways}}

\usepackage{amsmath}
\DeclareMathOperator*{\argmax}{arg\,max}
\DeclareMathOperator*{\argmin}{arg\,min}

\newtheoremstyle{sig}
  {}
  {}
  {\itshape}
  {}
  {\scshape}
  {.}
  {.5em}
  {#1 #2\thmnote{\quad(#3)}}

\theoremstyle{sig}

\newtheorem{dfn}{Definition}

\newcommand{\sstitle}[1]{\noindent\textbf{#1.\/}}

\newcommand{\eg}{e.\,g.,\ }

\newcommand\Mark[1]{\textsuperscript#1}

\makeatletter
\newcommand{\RemoveAlgoNumber}{\renewcommand{\fnum@algocf}{\AlCapSty{\AlCapFnt\algorithmcfname}}}
\newcommand{\RevertAlgoNumber}{\algocf@resetfnum}
\makeatother

\def\Snospace~{\S{}}

\usepackage{tikz}
\usetikzlibrary{shapes, arrows.meta, positioning, fit}
\usepackage{multirow}

\usepackage{pifont}% http://ctan.org/pkg/pifont
\newcommand{\cmark}{\ding{51}}%
\newcommand{\xmark}{\ding{55}}%
\usepackage{microtype}

\newlength{\bibitemsep}
\newlength{\bibparskip}
\let\oldthebibliography\thebibliography
\renewcommand\thebibliography[1]{%
  \oldthebibliography{#1}%
  \setlength{\parskip}{\bibitemsep}%
  \setlength{\itemsep}{\bibparskip}%
}

\begin{document}

\setlength{\belowdisplayskip}{3pt}
\setlength{\belowdisplayshortskip}{3pt}
\setlength{\abovedisplayskip}{3pt}
\setlength{\abovedisplayshortskip}{3pt}

\title{A Survey of Machine Unlearning}

%\author{Thanh Tam Nguyen}
%\affiliation{
%\institution{Griffith University}
%\country{Australia}
%}
%\email{t.nguyen19@griffith.edu.au}
%
%\author{Thanh Trung Huynh}
%\affiliation{
%\institution{\'{E}cole Polytechnique F\'{e}d\'{e}rale de Lausanne}
%\country{Switzerland}
%}
%\email{thanh.huynh@epfl.ch}
%
%\author{Phi Le Nguyen}
%\affiliation{
%\institution{Hanoi University of Science and Technology}
%\country{Vietnam}
%}
%\email{lenp@soict.hust.edu.vn}
%
%\author{Alan Wee-Chung Liew}
%\affiliation{
%\institution{Griffith University}
%\country{Australia}
%}
%\email{a.liew@griffith.edu.au}
%
%\author{Hongzhi Yin}
%\affiliation{
%\institution{The University of Queensland}
%\country{Australia}
%}
%\email{h.yin1@uq.edu.au}
%
%\author{Quoc Viet Hung Nguyen}
%\affiliation{
%\institution{Griffith University}
%\country{Australia}
%}
%\email{quocviethung.nguyen@griffith.edu.au}

\author{Thanh Tam Nguyen\Mark{1},
Thanh Trung Huynh\Mark{2},
Zhao Ren\Mark{3},
Phi Le Nguyen\Mark{4},
Alan Wee-Chung Liew\Mark{1},
Hongzhi Yin\Mark{5},
Quoc Viet Hung Nguyen\Mark{1}
}
%\affiliation{\vspace{5em}\country{}}
\affiliation{%
  \institution{
  \Mark{1} Griffith University,
  \Mark{2} \'{E}cole Polytechnique F\'{e}d\'{e}rale de Lausanne,
  \Mark{3} University of Bremen,\\
  \Mark{4} Hanoi University of Science and Technology,
  \Mark{5} The University of Queensland
  }
%  \institution{\'{E}cole Polytechnique F\'{e}d\'{e}rale de Lausanne}
  \country{}
}

\renewcommand{\shortauthors}{Thanh Tam Nguyen, et al.}

\begin{abstract}

Today, computer systems hold large amounts of personal data. Yet while such an abundance of data allows breakthroughs in artificial intelligence, and especially machine learning, its existence can be a threat to user privacy, and it can weaken the bonds of trust between humans and AI. Recent regulations now require that, on request, private information about a user must be removed from both computer systems and from machine learning models -- this legislation is more colloquially called ``the right to be forgotten''). While removing data from back-end databases should be straightforward, it is not sufficient in the AI context as machine learning models often `remember' the old data. Contemporary adversarial attacks on trained models have proven that we can learn whether an instance or an attribute belonged to the training data. This phenomenon calls for a new paradigm, namely \emph{machine unlearning}, to make machine learning models forget about particular data. It turns out that recent works on machine unlearning have not been able to completely solve the problem due to the lack of common frameworks and resources. Therefore, this paper aspires to present a comprehensive examination of machine unlearning's concepts, scenarios, methods, and applications. Specifically, as a category collection of cutting-edge studies, the intention behind this article is to serve as a comprehensive resource for researchers and practitioners seeking an introduction to machine unlearning and its formulations, design criteria, removal requests, algorithms, and applications. In addition, we aim to highlight the key findings, current trends, and new research areas that have not yet featured the use of machine unlearning but could benefit greatly from it. We hope this survey serves as a valuable resource for machine learning researchers and those seeking to innovate privacy technologies. Our resources are publicly available at \url{https://github.com/tamlhp/awesome-machine-unlearning}.

\end{abstract}

%\begin{CCSXML}
%<ccs2012>
%   <concept>
%       <concept_id>10010147.10010257</concept_id>
%       <concept_desc>Computing methodologies~Machine learning</concept_desc>
%       <concept_significance>300</concept_significance>
%       </concept>
%      <concept>
%       <concept_id>10002951.10003227</concept_id>
%       <concept_desc>Information systems~Information systems applications</concept_desc>
%       <concept_significance>100</concept_significance>
%       </concept>
%      <concept>
%       <concept_id>10002978.10003029.10011150</concept_id>
%       <concept_desc>Security and privacy~Privacy protections</concept_desc>
%       <concept_significance>300</concept_significance>
%       </concept>
%   <concept>
%       <concept_id>10003456.10003462.10003477</concept_id>
%       <concept_desc>Social and professional topics~Privacy policies</concept_desc>
%       <concept_significance>300</concept_significance>
%       </concept>
%\end{CCSXML}
%
%\ccsdesc[300]{Computing methodologies~Machine learning}
%\ccsdesc[100]{Information systems~Information systems applications}
%\ccsdesc[300]{Security and privacy~Privacy protections}
%\ccsdesc[300]{Social and professional topics~Privacy policies}


\maketitle

\section{Introduction}
\label{sec:intro}

Computer systems today hold large amounts of personal data. Due to the great advancement in data storage and data transfer technologies, the amount of data being produced, recorded, and processed has exploded. For example, four billion YouTube videos are watched every day~\cite{sari2020learning}). These online personal data, including digital footprints made by (or about) netizens, reflects their behaviors, interactions, and communication patterns in real-world~\cite{nguyen2019debunking}. 
Other sources of personal data include the digital content that online users create to express their ideas and opinions, such as product reviews, blog posts (e.g. Medium), status seeking (e.g. Instagram), and knowledge sharing (e.g. Wikipedia)~\cite{nguyen2021judo}. 
More recently, personal data has also expanded to include data from wearable devices~\cite{ren2022prototype}. 
On the one hand, such an abundance of data has helped to advance artificial intelligence (AI). However, on the other hand, it threatens the privacy of users and has led to many data breaches~\cite{cao2015towards}. 
For this reason, some users may choose to have their data completely removed from a system, especially sensitive systems such as those do with finance or healthcare~\cite{ren2022prototype}. 
Recent regulations now compel organisations to give users ``the right to be forgotten'', i.e., the right to have all or part of their data deleted from a system on request~\cite{dang2021right,hawkins2023decision}.

While removing data from back-end databases satisfies the regulations, doing so is not sufficient in the AI context as machine learning models often `remember' the old data. Indeed, in machine learning systems, often millions, if not billions, of users' data have been processed during the model's training phase.
However, unlike humans who learn general patterns, machine learning models behave more like a lossy data compression mechanism~\cite{schelter2020amnesia}, and some are overfit against their training data.
The success of deep learning models in particular has been recently been attributed to the compression of training data~\cite{tishby2000information,tishby2015deep}.
This memorization behaviour can be further proven by existing works on adversarial attacks~\cite{ren2020generating,chang2022example,ren2020enhancing}, which have shown that it is possible to extract the private information within some target data from a trained model.
However, we also know that the parameters of a trained model do not tend to show any clear connection to the data that was used for training~\cite{shwartz2017opening}.
As a result, it can be challenging to remove information corresponding to a particular data item from a machine learning model. In other words, it can be difficult to make a machine learning model forget a user's data.

\begin{table*}[!h]
    \centering
        \vspace{-.5em}
    \caption{Comparison between existing surveys on machine unlearning}
    \label{tab:survey_comparison}
    \vspace{-1em}
    \footnotesize
    \scalebox{0.9}{
%    \begin{adjustbox}{max width=\textwidth}
 \begin{threeparttable}
    \begin{tabular}{c|llllll|lllll|lllll|llllll}
        \toprule
        \multirow{2}{*}{\textbf{Surveys}} & \multicolumn{6}{c|}{\textbf{Unlearning Framework}}   & \multicolumn{5}{c|}{\textbf{Unlearning Scenarios}} & \multicolumn{5}{c|}{\textbf{Unlearning Requests}} & \multicolumn{5}{c}{\textbf{Applications}} \\ 
        \cline{2-23}
        & \shorthead{Definitions} & \shorthead{Requirements} & \shorthead{Techniques}  & \shorthead{Verification} & \shorthead{Metrics} & \shorthead{Resources} & \shorthead{Exact} & \shorthead{Approximate} & \shorthead{Zero-glance} & \shorthead{Zero-shot} &  \shorthead{Few-shot} & \shorthead{Item} & \shorthead{Feature} & \shorthead{Class} & \shorthead{Task} & \shorthead{Stream} & \shorthead{Recommendation} & \shorthead{Fed Learning} & \shorthead{Graph Embedding} & \shorthead{Lifelong Learning} & \shorthead{LLM} & \shorthead{Generative Models} \\
%         & Parameter & Attack & Privacy & Augmentation & Influence & Model Update & Verification & Item & Feature & Class & Task   \\ 
         \midrule
         Ours & \cmark & \cmark & \cmark & \cmark & \cmark & \cmark & \cmark & \cmark & \cmark & \cmark & \cmark & \cmark & \cmark & \cmark & \cmark & \cmark & \cmark & \cmark & \cmark & \cmark & \cmark & \cmark\\
%        \cite{cao2015towards}  \\
%       \cite{shokri2017membership}  \\
       \cite{villaronga2018humans} & \cmark & \cmark & \cmark & \xmark & \xmark & \xmark & \xmark & \xmark & \xmark & \xmark & \xmark & \cmark & \xmark & \xmark & \xmark & \xmark & \xmark & \xmark & \xmark & \xmark & \xmark & \xmark \\
       \cite{veale2018algorithms} & \cmark & \cmark & \xmark & \cmark & \xmark & \xmark & \xmark & \xmark & \xmark & \xmark & \xmark & \xmark & \xmark & \xmark & \xmark & \xmark & \xmark & \xmark & \xmark & \xmark & \xmark & \xmark \\
       \cite{shintre2019making} & \cmark & \xmark & \cmark & \xmark & \xmark & \xmark & \xmark & \xmark & \xmark & \xmark & \xmark & \xmark & \xmark & \xmark & \xmark & \xmark & \xmark & \xmark & \xmark & \xmark & \xmark & \xmark \\
       \cite{schelter2020amnesia} & \cmark & \xmark & \cmark & \xmark & \cmark & \cmark & \cmark & \xmark & \xmark & \xmark & \xmark  & \cmark & \xmark & \xmark & \xmark & \xmark & \xmark & \xmark & \xmark &  \xmark & \xmark & \xmark \\
       \cite{mercuri2022unlearning} & \cmark & \xmark & \cmark & \xmark & \cmark & \xmark & \cmark & \cmark & \xmark & \xmark & \xmark  & \cmark & \xmark & \xmark & \xmark & \xmark & \xmark & \xmark & \xmark &  \xmark & \xmark & \xmark \\
       \cite{heng2023unlearningsurvey} & \cmark & \cmark & \cmark & \cmark & \cmark & \cmark & \cmark & \cmark & \xmark & \xmark & \xmark  & \cmark & \cmark & \cmark & \cmark & \cmark & \xmark &  \xmark & \xmark & \xmark & \xmark & \xmark\\
      \cite{xu2024machine} & \cmark & \xmark & \cmark & \xmark & \xmark & \xmark & \cmark & \cmark & \xmark & \xmark & \xmark & \cmark & \xmark & \xmark & \xmark & \xmark & \xmark & \xmark & \cmark & \xmark & \xmark & \xmark \\ 
      \cite{shaik2023exploring} & \cmark & \xmark & \cmark & \xmark & \cmark & \cmark & \cmark & \xmark & \xmark & \cmark & \cmark & \cmark & \xmark & \xmark & \xmark & \xmark & \cmark & \xmark & \xmark & \xmark & \xmark & \xmark \\
      \cite{mercuri2022introduction} & \cmark & \xmark & \cmark & \xmark & \cmark & \cmark & \cmark & \cmark & \xmark & \xmark & \xmark & \cmark & \xmark & \xmark & \xmark & \xmark & \xmark & \xmark & \xmark & \xmark & \xmark & \xmark \\
%       \cite{izzo2021approximate} \\
%       \cite{felps2020class} \\
%       \cite{garg2020formalizing} \\
         \bottomrule
    \end{tabular}
        \begin{tablenotes}
\item Further surveys on machine unlearning for specific domains can be found at~\url{https://github.com/tamlhp/awesome-machine-unlearning}
 \end{tablenotes}
     \end{threeparttable}
%    \end{adjustbox}
}
\end{table*}

The challenge of enabling users to fully delete their data from a machine learning model has led to the development of a new paradigm: \emph{machine unlearning}~\cite{nguyen2022markov,baumhauer2020machine,tahiliani2021machine}. Ideally, a machine unlearning mechanism would remove data from the model without requiring complete retraining~\cite{nguyen2022markov}. This approach upholds users' right to be forgotten while sparing model owners from frequent and costly retraining.

%This challenge of allowing users the possibility and flexibility to completely delete their data from a machine learning model calls for a new paradigm, namely \emph{machine unlearning}~\cite{nguyen2022markov,baumhauer2020machine,tahiliani2021machine}. Ideally, a machine unlearning mechanism would remove data from the model without needing to retrain it from scratch~\cite{nguyen2022markov}. To this end, a users' right to be forgotten would be observed and the model owner would be shielded from constant and expensive retraining exercises.

Researchers have already begun to study aspects of machine unlearning, such as removing part of the training data and analysing the subsequent model predictions~\cite{nguyen2022markov,thudi2021necessity}. 
However, it turns out that this problem cannot be completely solved due to a lack of common frameworks and resources~\cite{villaronga2018humans,veale2018algorithms,shintre2019making,schelter2020amnesia}. Hence, to begin building a foundation of works in this nascent area, we undertook a comprehensive survey of machine unlearning: its definitions, scenarios, mechanisms, and applications. Our resources are publicly available at~\footnote{\url{https://github.com/tamlhp/awesome-machine-unlearning}}.

\subsection{Reasons for Machine Unlearning}

There are many reasons for why a users may want to delete their data from a system. We have categorized these into four major groups: security, privacy, usability, and fidelity. Each reason is discussed in more detail next.

\sstitle{Security}
Recently, deep learning models have been shown to be vulnerable to external attacks, especially adversarial attacks~\cite{ren2020adversarial}. In an adversarial attack, the attacker generates adversarial data that are very similar to the original data to the extent that a human cannot distinguish between the real and fake data. This adversarial data is designed to force the deep learning models into outputting wrong predictions, which frequently results in serious problems. For example, in healthcare, a wrong prediction could lead to a wrong diagnosis, a non-suitable treatment, even a death. Hence, detecting and removing adversarial data is essential for ensuring the model's security and, once an attack is detected, the model needs to be able delete the adversarial data through a machine unlearning mechanism~\cite{cao2015towards,marchant2022hard}.

% \info{Security is another reason that users want data to be
% forgotten. Consider anomaly detection systems. The security of these systems hinges on the model of normal behaviors ex- tracted from the training data. By polluting1 the training data, attackers pollute the model, thus compromising security. For instance, Perdisci et al. [56] show that PolyGraph [55], a worm detection engine, fails to generate useful worm signatures if the training data is injected with well-crafted fake network flows. Once the polluted data is identified, the system must completely}
% \cite{cao2015towards}

% \cite{marchant2022hard}

\sstitle{Privacy}
Many privacy-preserving regulations have been enacted recently that involve the right to be forgotten''~\cite{bourtoule2021machine,dang2021right}, such as the European Union's General Data Protection Regulation (GDPR)~\cite{magdziarczyk2019right} and the California
Consumer Privacy Act~\cite{pardau2018california}. In this particular regulation, users must be given the right to have their data and related information deleted to protect their privacy. In part, this legislation has sprung up as a result of privacy leaks. For example,
cloud systems can leak user data due to multiple copies of data hold by different parties, backup policies, and replication strategies~\cite{singh2017data}.
In another case, machine learning approaches for genetic data processing were found to leak patients' genetic markers~\cite{fredrikson2014privacy,wang2009learning}. It is therefore not surprising that users would want to remove their data to avoid the risks of a data leak~\cite{cao2015towards}.

\sstitle{Usability}
People have difference preferences in online applications and/or services, especially recommender systems. An application will produce inconvenient recommendations if it cannot completely delete the incorrect data (\eg noise, malicious data, out-of-distribution data) related to a user. For example, one can accidentally search for an illegal product on his laptop, and find that he keeps getting this product recommendation on this phone, even after he cleared his web browser history~\cite{cao2015towards}. Such undesired usability by not forgetting data will not only produce wrong predictions, but also result in less users.

% \info{Usability is a third reason. Consider the recommendation
% or prediction system Google Now [7]. It infers a user's preferences from her search history, browsing history, and other analytics. It then pushes recommendations, such as news about a show, to the user. Noise or incorrect entries in analytics can seriously degrade the quality of the recommendation. One of our lab members experienced this problem first-hand. He loaned his laptop to a friend who searched for a TV show (``Jeopardy!'') on Google [1]. He then kept getting news about this show on his phone, even after he deleted the search record from his search history.}
% \cite{cao2015towards}

\sstitle{Fidelity}
Unlearning requests might come from biased machine learning models. Despite recent advances, machine learning models are still sensitive to bias that means their output can unfairly discriminate against a group of people~\cite{mehrabi2021survey}. For example, COMPAS, the software used by courts to decide parole cases, is more likely to consider African-American offenders to have higher risk scores than Caucasians, even though ethnicity information is not part of the input~\cite{zou2018ai}. Similar situations have been observed in beauty contest judged by AI, which was biased against contestants with darker skin tones, or facial recognition AI that wrongly recognized Asian facial features~\cite{feuerriegel2020fair}.

The source of these biases often originate from data. For example, AI systems that have been trained on public datasets that contain mostly white persons, such as ImageNet, are likely to make errors when processing images of black persons. Similarly, in an application screening system, inappropriate features, such as the gender or race of applicants, might be unintentionally learned by the machine learning model~\cite{dinsdale2021deep,dinsdale2020unlearning}. As a result, there is a need to unlearn these data, including the features and affected data items.

\subsection{Challenges in Machine Unlearning}
Before we can truly achieve machine unlearning, several challenges to removing specific parts of the training data need to be overcome. The challenges are summarized as follows.

\sstitle{Stochasticity of training}
We do not know the impact of each data point seen during training on the machine learning model due to the stochastic nature of the training procedure~\cite{bourtoule2021machine}. Neural networks, for example, are usually trained on random mini-batches containing a certain number of data samples. Further, the order of the training batches is also random~\cite{bourtoule2021machine}. This stochasticity raises difficulties for machine unlearning as the specific data sample to be removed would need to be removed from all batches.

% \info{We have a limited understanding of how each data point
% impacts the model. There exists no prior work that measures
% the influence of a particular training point on the parameters
% of a model.}
% \cite{bourtoule2021machine}

% \info{Stochasticity in training. A great deal of randomness exists
% in the training methods for complicated models such as DNNs;
% small batches of data (e.g., with 32 points) are randomly
% sampled from the dataset, and the ordering of batches varies
% between different epochs, i.e., passes of the algorithm through
% the dataset. Further, training is often parallelized without explicit synchronization, meaning the inherent random ordering
% of parallel threads may make the training non-deterministic.}
% \cite{bourtoule2021machine}

\sstitle{Incrementality of training}
A model's training procedure is an incremental process~\cite{bourtoule2021machine}. In other words, the model update on a given data sample will affect the model performance on data samples fed into the model after this data. A model's performance on this given data sample is also affected by prior data samples. Determining a way to erase the effect of the to-be-removed training sample on further model performance is a challenge for machine unlearning.

% \info{Training is incremental. Additionally, training is an incremental procedure where any given update reflects all updates
% that have occurred prior to it. For example, if a model is
% updated based on a particular training point (in a particular
% batch) at a particular epoch, all subsequent model updates will
% depend, in some implicit way, on that training point.}
% \cite{bourtoule2021machine}

% \info{Stochasticity in learning. Intuitively, learning algorithms are
% designed to search for an optimal hypothesis in a vast hypothesis space. In the case of neural networks, this space contains
% all models that can be defined by setting the weights of a fixed
% neural network architecture.}
% \cite{bourtoule2021machine}

\sstitle{Catastrophic unlearning}
In general, an unlearned model usually performs worse than the model retrained on the remaining data~\cite{nguyen2020variational,nguyen2022markov}. However, the degradation can be exponential when more data is unlearned. Such sudden degradation is often referred as catastrophic unlearning~\cite{nguyen2020variational}. While several studies~\cite{du2019lifelong,golatkar2020eternal} have explored ways to mitigate catastrophic unlearning by designing special loss functions, how to naturally prevent catastrophic unlearning is still an open question. 

% \info{We also show that MCU does not suffer from an important pitfall of
% catastrophic unlearning (Section 6) known to affect MU algorithms.}
% \cite{nguyen2022markov}

% \info{A trained ML model is said to experience catastrophic unlearning from the erased data when its resulting performance is considerably worse than that from retraining with the remaining data.
% Depending on its choice of loss function, machine unlearning may suffer from catastrophic unlearning that can render the model useless. For example, to mitigate this issue, the works of [9, 13] have to ``patch up'' their loss functions by additionally bounding the loss incurred by erased data with a rectified linear unit and injecting a regularization term to retain information of the remaining data, respectively. This begs the question whether there exists a loss function that can directly quantify the approximation gap and naturally prevent catastrophic unlearning.
% }
% \cite{nguyen2020variational}

\subsection{Contributions of this survey}
The aim of this paper is to supply a complete examination of research studies on machine unlearning as well as a discussion on potential new research directions in machine unlearning. The contributions of our survey can therefore be summarized as follows.

\begin{compactitem}
\item First, we show how to design an unlearning framework. We discuss the design requirements, different types of unlearning requests, and how to verify the unlearned model. The details can be found in \autoref{sec:framework}.
\item Second, we show how to define an unlearning problem in machine learning systems. This includes the formulation of exact unlearning and approximate unlearning as well as the definition of indistinguishability metrics to compare two given models (i.e., the unlearned model and the retrained model). The details are discussed in \autoref{sec:problem}.
\item Third, we discuss different scenarios of machine unlearning, including zero-glance unlearning, zero-shot unlearning, and few-shot unlearning. The details are provided in \autoref{sec:scenarios}
\item Fourth, we introduce a unified taxonomy that categorizes the machine unlearning approaches into three branches: model-agnostic methods, model-intrinsic methods, and data-driven methods. The details can be found in \autoref{sec:algorithms}.
\item Fifth, we compile a variety of regularly used datasets and open-source implementations to serve as a foundation for future machine unlearning research and benchmarking. The details are provided in \autoref{sec:resources}.
\item Finally, we highlight the findings, trends and the forthcoming according to our survey in \autoref{sec:discussion}. \autoref{sec:conclusion} then completes the paper.
\end{compactitem}

\subsection{Differences between this and previous surveys}

\autoref{tab:survey_comparison} summarizes the differences between our survey and existing efforts to unify the field. It is noteworthy that machine unlearning is different from data deletion~\cite{garg2020formalizing}. Some works focus on exploring theoretical foundations or subcategories of unlearning techniques~\cite{xu2024machine,shaik2023exploring}. Both topics concern the right to be forgotten legislated and exercised across the world~\cite{magdziarczyk2019right}. However, the latter focuses only on the data perspective following the General Data Protection Regulation (GDPR)~\cite{voigt2017eu}, while machine unlearning also addresses privacy problems from a model perspective.

There are some other concepts that might be mistaken as machine unlearning, such as data redaction that aims to poison the label information of the data to be forgotten inside the model~\cite{felps2020class}. In other words, it forces the model make wrong predictions about the forgotten data. Although applicable in some setting, this approach is not fully compatible with machine unlearning as the forgotten data has to be known a priori when the original model is trained~\cite{felps2020class}.

\section{Unlearning Framework}
\label{sec:framework}

\subsection{Unlearning Workflow}

%\info{Given a training data sample to forget, unlearning updates
%the system in two steps, following the learning process shown
%in \autoref{fig:unlearning_workflow}. First, it updates the set of selected features. 
%Second, unlearning updates the model.}
%\cite{cao2015towards}

\begin{figure*}[!h]
	\centering
%	\vspace{-1em}
	\includegraphics[width=0.9\linewidth]{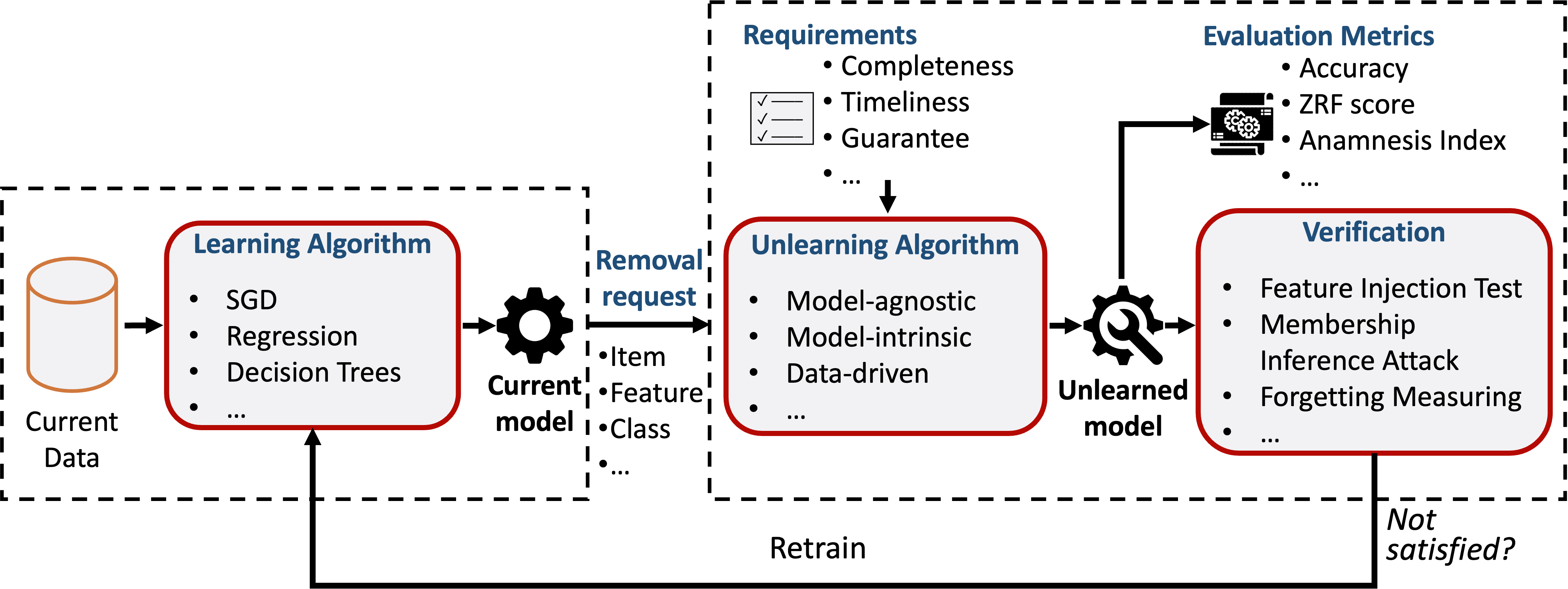}
%	\caption{\info{The common unlearning pipeline with the three stages of training, inference and unlearning. First, an initial model $w^*$ is trained on all data and used for inference. Subsequently, whenever a part Dm of the data is deleted, an updated model wu is obtained via machine unlearning. The pipeline restarts if the updated model is deemed inadequate.}\cite{mahadevan2022certifiable,mahadevan2021certifiable}}
%\vspace{-1em}
\caption{A Machine Unlearning Framework}
	\label{fig:unlearning_workflow}
%	\vspace{-1em}
\end{figure*}

The unlearning framework in \autoref{fig:unlearning_workflow} presents the typical workflow of a machine learning model in the presence of a data removal request. In general, a model is trained on some data and is then used for inference. Upon a removal request, the data-to-be-forgotten is unlearned from the model. The unlearned model is then verified against privacy criteria, and, if these criteria are not met, the model is retrained, i.e., if the model still leaks some information about the forgotten data. There are two main components to this process: the \emph{learning component} (left) and the \emph{unlearning component} (right). The learning component involves the current data, a learning algorithm, and the current model. In the beginning, the initial model is trained from the whole dataset using the learning algorithm. The unlearning component involves an unlearning algorithm, the unlearned model, optimization requirements, evaluation metrics, and a verification mechanism. Upon a data removal request, the current model will be processed by an unlearning algorithm to forget the corresponding information of that data inside the model. The unlearning algorithm might take several requirements into account such as completeness, timeliness, and privacy guarantees. The outcome is an unlearned model, which will be evaluated against different performance metrics (e.g., accuracy, ZRF score, anamnesis index). However, to provide a privacy certificate for the unlearned model, a verification (or audit) is needed to prove that the model actually forgot the requested data and that there are no information leaks. This audit might include a feature injection test, a membership inference attack, forgetting measurements, etc. 

If the unlearned model passes the verification, it becomes a new model for downstream tasks (e.g., inference, prediction, classification, recommendation). Otherwise, the remaining data, i.e., the original data excluding the data to be forgotten, needs to be used to retrain the model. Either way, the unlearning component will be called repeatedly upon a new removal request.

%Below, we describe how each typical unlearning method executes each stage of the pipeline, with a summary shown in \autoref{tab:workflow}.
%
%
%
%
%
%
%
%\sstitle{Training stage}
%This stage produces an ML model from the training dataset $D$. 
%
%\sstitle{Inference stage}
%Thi stage executes the unlearning algorithm so as to ``unlearn'' the removed data $D_m$ and produce an updated model ${w}_u$.
%
%\sstitle{Evaluation}
%During evaluation the updated model $w_u$ is assessed for certifiability as soon as it is produced using the test dataset $D_{test}$.

\subsection{Unlearning Requests}

\sstitle{Item Removal}
Requests to remove certain items/samples from the training data are the most common requests in machine unlearning~\cite{bourtoule2021machine}. The techniques used to unlearn these data are described in detail in \autoref{sec:algorithms}.
%In machine unlearning, users may want to remove some of or all of their data samples due to the consideration of privacy. 

\sstitle{Feature Removal}
In many scenarios, privacy leaks might not only originate from a single data item but also in a group of data with the similar features or labels~\cite{warnecke2021machine}. For example, a poisoned spam filter might misclassify malicious addresses that are present in thousands of emails. Thus, unlearning suspicious emails might not enough. Similarly, in an application screening system, inappropriate features, such as the gender or race of applicants, might need to be unlearned for thousands of affected applications.

In such cases, naively unlearning the affected data items sequentially is imprudent as repeated retraining is computationally expensive. Moreover, unlearning too many data items can inherently reduce the performance of the model, regardless of the unlearning mechanism used. Thus, there is a need for unlearning data at the feature or label level with an arbitrary number of data items.

Warnecke et al.~\cite{warnecke2021machine} proposed a technique for unlearning a group of training data based on influence functions. More precisely, the effect of training data on model parameter updates is estimated and formularized in closed-form. As a result of this formulation, influences of the learning sets act as a compact update instead of solving an optimisation problem iteratively (e.g., loss minimization). First-order and second-order derivatives are the keys to computing this update effectively~\cite{warnecke2021machine}. 

Guo et al.~\cite{guo2022efficient} proposed another technique to unlearn a feature based on disentangled representation. The core idea is to learn the correlation between features from the latent space as well as the effects of each feature on the output space. Using this information, certain features can be progressively detached from the learnt model upon request, while the remaining features are still preserved to maintain good accuracy. However, this method is mostly applicable to deep neural networks in the image domain, in which the deeper convolutional layers become smaller and can therefore identify abstract features that match real-world data attributes.

\sstitle{Class Removal}
There are many scenarios where the forgetting data belongs to single or multiple classes from a trained model. For example, in face recognition applications, each class is a person's face so there could potentially be thousands or millions of classes. However, when a user opts out of the system, their face information must be removed without using a sample of their face.

Similar to feature removal, class removal is more challenging than item removal because retraining solutions can incur many unlearning passes. Even though each pass might only come at a small computational cost due to data partitioning, the expense mounts up. However, partitioning data by class itself does not help the model's training in the first place, as learning the differences between classes is the core of many learning algorithms~\cite{tanha2020boosting}.
Although some of the above techniques for feature removal can be applied to class removal~\cite{warnecke2021machine}, it is not always the case as class information might be implicit in many scenarios.

Tarun et al.~\cite{tarun2021fast} proposed an unlearning method for class removal based on data augmentation. The basic concept is to introduce noise into the model such that the classification error is maximized for the target class(es). The model is updated by training on this noise without the need to access any samples of the target class(es). Since such impair step may disturb the model weights and degrade the classification performance for the remaining classes, a repair step is needed to train the model for one or a few more epochs on the remaining data. 
Their experiments show that the method can be efficient for large-scale multi-class problems (100 classes). Further, the method worked especially well with face recognition tasks because the deep neural networks were originally trained on triplet loss and negative samples so the difference between the classes was quite significant~\cite{masi2018deep}.

Baumhauer et al.~\cite{baumhauer2020machine} proposed an unlearning method for class removal based on a linear filtration operator that proportionally shifts the classification of the samples of the class to be forgotten to other classes. However, the approach is only applicable to class removal due to the characteristics of this operator.

\sstitle{Task Removal}
Today, machine learning models are not only trained for a single task but also for multiple tasks. This paradigm, aka continual learning or lifelong learning~\cite{parisi2019continual}, is motivated by the human brain, in which learning multiple tasks can benefit each other due to their correlations. This technique is also used overcome data sparsity or cold-start problems where there is not enough data to train a single task effectively. 

However, in these settings too, there can be a need to remove private data related to a specific task. For example, consider a robot that is trained to assist a patient at home during their medical treatment. This robot may be asked to forget this assistance behaviour after the patient has recovered~\cite{liu2022continual}. To this end, temporarily learning a task and forgetting it in the future has become a need for lifelong learning models. 

In general, unlearning a task is uniquely challenging as continual learning might depend on the order of the learned tasks. Therefore, removing a task might create a catastrophic unlearning effect, where the overall performance of multiple tasks is degraded in a domino-effect~\cite{liu2022continual}. Mitigating this problem requires the model to be aware of that the task may potentially be removed in future. Liu et al.~\cite{liu2022continual} explains that this requires users to explicitly define which tasks will be learned permanently and which tasks will be learned only temporarily.

\sstitle{Stream Removal}
Handling data streams where a huge amount of data arrives online requires some mechanisms to retain or ignore certain data while maintaining limited storage~\cite{nguyen2017retaining}. 
In the context of machine unlearning, however, handling data streams is more about dealing with a stream of removal requests.

Gupta et el.~\cite{gupta2021adaptive} proposed a streaming unlearning setting involving a sequence of data removal requests. This is motivated by the fact that many users can be involved in a machine learning system and decide to delete their data sequentially. Such is also the case when the training data has been poisoned in an adversarial attack and the data needs to be deleted gradually to recover the model's performance. These streaming requests can be either non-adaptive or adaptive. A non-adaptive request means that the removal sequence does not depend on the intermediate results of each unlearning request, whereas and adaptive request means that the data to be removed depends on the current unlearned model. In other words, after the poisonous data is detected, the model is unlearned gradually so as to decide which data item is most beneficial to unlearn next.

\subsection{Design Requirements}

\sstitle{Completeness (Consistency)}
A good unlearning algorithm should be complete~\cite{cao2015towards}, i.e. the unlearned model and the retrained model make the same predictions about any possible data sample (whether right or wrong). One way to measure this consistency is to compute the percentage of the same prediction results on a test data. This requirement can be designed as an optimization objective in an unlearning definition (\autoref{sec:exact_unlearning}) by formulating the difference between the output space of the two models. Many works on adversarial attacks can help with this formulation~\cite{sommer2022athena,chen2021machine}.

%\info{1) Completeness: Intuitively, completeness requires that
%once a data sample is removed, all its effects on the feature set
%and the model are also cleanly reversed. It essentially captures
%how consistent an unlearned system is with the system that
%has been retrained from scratch. If, for every possible sample,
%the unlearned system gives the same prediction result as the
%retrained system, then an attacker, operator, or user has no
%way of discovering that the unlearned data and its lineage
%existed in the system by feeding input samples to the unlearned
%system or even observing its features, model, and training
%data. Such unlearning is complete. To empirically measure
%completeness, we quantify the percentage of input samples that
%receive the same prediction results from both the unlearned
%and the retrained system using a representative test data set.
%The higher the percentage, the more complete the unlearning.
%Note that completeness does not depend on the correctness
%of prediction results: an incorrect but consistent prediction by
%both systems does not decrease completeness.}
%\cite{cao2015towards}

\sstitle{Timeliness}
In general, retraining can fully solve any unlearning problem. However, retraining is time-consuming, especially when the distribution of the data to be forgotten is unknown~\cite{cao2015towards,bourtoule2021machine}. As a result, there needs to be a trade-off between completeness and timeliness. Unlearning techniques that do not use retraining might be inherently not complete, i.e., they may lead to some privacy leaks, even though some provable guarantees are provided for special cases~\cite{GuoGHM20,marchant2022hard,neel2021descent}. To measure timeliness, we can measure the speed up of unlearning over retraining after an unlearning request is invoked.

It is also worth recognizing the cause of this trade-off between retraining and unlearning. When there is not much data to be forgotten, unlearning is generally more beneficial as the effects on model accuracy are small. However, when there is much forgetting data, retraining might be better as unlearning many times, even bounded, may catastrophically degrade the model's accuracy~\cite{cao2015towards}.

%\info{2) Timeliness: Timeliness in unlearning captures how much
%faster unlearning is than retraining at updating the features
%and the model in the system. The more timely the unlearning,
%the faster the system is at restoring privacy, security, and
%usability. Analytically, unlearning updates only a small number
%of summations and then runs a learning algorithm on these
%summations, whereas retraining runs the learning algorithm
%on the entire training data set, so unlearning is asymptotically
%faster by a factor of the training data size. To empirically measure timeliness, we quantify the speedup of unlearning over
%retraining. Unlearning does not replace retraining. Unlearning
%works better when the data to forget is small compared to the training set. This case is quite common. For instance, a single
%user's private data is typically small compared to the whole
%training data of all users. Similarly, an attacker needs only a
%small amount of data to pollute a learning system. When the data
%to forget becomes large, retraining may work better
%}
%\cite{cao2015towards}

\sstitle{Accuracy}
An unlearned model should be able to predict test samples correctly. Or at least its accuracy should be comparable to the retrained model. However, as retraining is computationally costly, retrained models are not always available for comparison. To address this issue, the accuracy of the unlearned model is often measured on a new test set, or it is compared with that of the original model before unlearning~\cite{he2021deepobliviate}.

\sstitle{Light-weight}
To prepare for unlearning process, many techniques need to store model checkpoints, historical model updates, training data, and other temporary data~\cite{he2021deepobliviate,bourtoule2021machine,liu2020federated}. A good unlearning algorithm should be light-weight and scale with big data. Any other computational overhead beside unlearning time and storage cost should be reduced as well~\cite{bourtoule2021machine}.

%\info{Intelligibility: Conceptually, the baseline strategy is very
%easy to understand and implement. Similarly, any unlearning strategy should be intelligible; this requirement
%ensures that the strategy is easy to debug by non-experts}
%\cite{bourtoule2021machine}

%\info{Comparable Accuracy: It is conceivable that the accuracy
%of the model degrades, even in the baseline, if (a) the
%fraction of training points that need to be unlearned}
%\cite{bourtoule2021machine}
%
%\info{Reduced Unlearning Time: The strategy should have
%provably lower time than the baseline for unlearning any
%number of points}
%\cite{bourtoule2021machine}

\sstitle{Provable guarantees}
With the exception of retraining, any unlearning process might be inherently approximate. It is practical for an unlearning method to provide a provable guarantee on the unlearned model. To this end, many works have designed unlearning techniques with bounded approximations on retraining~\cite{GuoGHM20,marchant2022hard,neel2021descent}. Nonetheless, these approaches are founded on the premise that models with comparable parameters will have comparable accuracy.

%\info{Provable Guarantees: Like the baseline, any new strategy
%should provide provable guarantees that any number of
%points have been unlearned (and do not influence model
%parameters). Additionally, such a guarantee should be
%intuitive and easy to understand for non-experts [31].}
%\cite{bourtoule2021machine}

\sstitle{Model-agnostic}
An unlearning process should be generic for different learning algorithms and machine learning models~\cite{bourtoule2021machine}, especially with provable guarantees as well. However, as machine learning models are different and have different learning algorithms as well, designing a model-agnostic unlearning framework could be challenging.

%\info{Model Agnostic: The new strategy for unlearning should
%be general i.e., should provide the aforementioned guarantees for models of varying nature and complexity.}
%\cite{bourtoule2021machine}

%\info{Limited Overhead: Any new unlearning strategy should
%not introduce additional overhead to what are already
%computationally-intense training procedures.}
%\cite{bourtoule2021machine}

\sstitle{Verifiability}
Beyond unlearning requests, another demand by users is to verify that the unlearned model now protects their privacy. To this end, a good unlearning framework should provide end-users with a verification mechanism. For example, backdoor attacks can be used to verify unlearning by injecting backdoor samples into the training data~\cite{sommer2020towards}. If the backdoor can be detected in the original model while not detected in the unlearned model, then verification is considered to be a success. However, such verification might be too intrusive for a trustworthy machine learning system and the verification might still introduce false positive due to the inherent uncertainty in backdoor detection.

\subsection{Unlearning Verification}

The goal of unlearning verification methods is to certify that one cannot easily distinguish between the unlearned models and their retrained counterparts~\cite{thudi2021necessity}. While the evaluation metrics (\autoref{sec:metrics}) are theoretical criteria for machine unlearning, unlearning verification can act as a certificate for an unlearned model. They also include best practices for validating the unlearned models efficiently. 

It is noteworthy that while unlearning metrics (in \autoref{sec:formulation}) and verification metrics share some overlaps, the big difference is that the former can be used for optimization or to provide a bounded guarantee, while the latter is used for evaluation only.

\sstitle{Feature Injection Test}
The goal of this test is to verify whether the unlearned model has adjusted the weights corresponding to the removed data samples based on data features/attributes~\cite{izzo2021approximate}. The idea is that if the set of data to be forgotten has a very distinct feature distinguishing it from the remaining set, it gives a strong signal for the model weights. However, this feature needs to be correlated with the labels of the set to be forgotten, otherwise the model might not learn anything from this feature. 

More precisely, an extra feature is added for each data item such that it is equal to zero for the remaining set and is perfectly correlated with the labels of the set to forget. Izzo et al.~\cite{izzo2021approximate} applied this idea with linear classifiers, where the weight associated with this extra feature is expected to be significantly different from zero after training. After the model is unlearned, this weight is expected to become zero. As a result, the difference of this weight can be plotted before and after unlearning as a measure of effectiveness of the unlearning process.

One limitation of this verification method is that the current solution is only applicable for linear and logistic models~\cite{izzo2021approximate}. This is because these models have explicit weights associated with the injected feature, whereas, for other models such as deep learning, injecting such a feature as a strong signal is non-trivial, even though the set to be forgotten is small. Another limitation to these types of methods is that an injected version of the data needs to be created so that the model can be learned (either from scratch or incrementally depending on the type of the model).

\sstitle{Forgetting Measuring}
Even after the data to be forgotten has been unlearned from the model, it is still possible for the model to carry detectable traces of those samples~\cite{jagielski2022measuring}. Jagielski et al.~\cite{jagielski2022measuring} proposed a formal way to measure the forgetfulness of a model via privacy attacks. More precisely, a model is said to $\alpha$-forget a training sample if a privacy attack (e.g., a membership inference) on that sample achieves no greater than success rate $\alpha$.  This definition is more flexible than differential privacy because a training algorithm is differentially private only if it immediately forgets every sample it learns. As a result, this definition allows a sample to be temporarily learned, and measures how long until it is forgotten by the model.

%\info{Rather than study forgetting through the lens of the model's accuracy on a specific example -- as done in the catastrophic forgetting literature -- we use an instance-specific notion of forgetting: we attempt to detect a specific example's presence in training. In catastrophic forgetting, the model's accuracy degrades on an entire sub-distribution. This does not necessarily have implications for forgetting of memorized information contained in individual examples. It is possible that, despite an accuracy decrease, the model still carries detectable traces of those examples, which is harmful for privacy. It is also possible that a high accuracy model, which has not yet forgotten the sub-distribution, generalizes well to the sub-distribution rather than memorizing the specifics of the training set. Thus, our definition of forgetting instead asks a more privacy-motivated question: whether it is possible to detect or extract an example in the training set. Hence, we draw from the literature on attacking the privacy of ML training data to define forgetting.}
%\cite{jagielski2022measuring}

\sstitle{Information Leakage}
Many machine learning models inherently leak information during the model updating process~\cite{chen2021machine}. Recent works have exploited this phenomenon by comparing the model before and after unlearning to measure the information leakage. More precisely, Salem et al.~\cite{salem2020updates} proposed an adversary attack in the image domain that could reconstruct a removed sample when a classifier is unlearned on a data sample. Brockschmidt et al.~\cite{zanella2020analyzing} suggested a similar approach for the text domain. Chen et al.~\cite{chen2021machine} introduced a membership inference attack to detect whether a removed sample belongs to the learning set. 
Compared to previous works~\cite{Salem0HBF019,shokri2017membership}, their approach additionally makes use of the posterior output distribution of the original model, besides that of the unlearned model.
Chen et al.~\cite{chen2021machine} also proposed two leakage metrics, namely the degradation count and the degradation rate.

\begin{compactitem}
\item The \emph{degradation count:} is defined as the ratio between the number of target samples whose membership can be inferred by the proposed attack with higher confidence compared to traditional attacks and the total number of samples.

\item The \emph{degradation rate:} is defined the average improvement rate of the confidence of the proposed attack compared to traditional attacks.
\end{compactitem}

%\info{
%There are recent studies aiming to quantify the information leakage in the model updating process. Salem et al.~\cite{salem2020updates} show that in the online learning applications, where an ML image classifier is updated by new data samples, the adversary can reconstruct the updated samples by exploiting information from two versions of the target ML model (before and after the updating). Brockschmidt et al.~\cite{zanella2020analyzing} show similar results in the natural language models as well as the data deletion scenario. This line of work is related to our attack in the sense that we all study the unintended information leakage in model updating processes.
%}

%\info{
%Existing works have proposed to use techniques such as membership inference [21] to verify the effectiveness of approximate unlearning [1, 12] to show that their approximately unlearned models cannot be easily distinguished from models that are not trained on the data points to be unlearned. Alternatively, others compare the similarity of the approximately unlearned models parameters to exactly unlearned models parameters [10, 11, 23, 25].
%}
%\cite{thudi2021necessity}

\sstitle{Membership Inference Attacks}
This kind of attack is designed to detect whether a target model leaks data~\cite{shokri2017membership,thudi2022bounding,chen2021machine}. Specifically, an inference model is trained to recognise new data samples from the training data used to optimize the target model. In~\cite{shokri2017membership}, a set of shallow models were trained on a new set of data items different from the one that the target model was trained on. The attack model was then trained to predict whether a data item belonged to the training data based on the predictions made by shallow models for training as well as testing data. The training set for the shallow and attack models share similar data distribution to the target model.
Membership inference attacks are helpful for detecting data leaks. Hence, they are useful for verifying the effectiveness of the machine unlearning~\cite{chen2021machine}.

\sstitle{Backdoor attacks}
Backdoor attacks were proposed to inject backdoors to the data for deceiving a machine learning model~\cite{wang2019neural}. The deceived model makes correct predictions with clean data, but with poison data in a target class as a backdoor trigger, it makes incorrect predictions. Backdoor attacks were used to verify the effectiveness of machine unlearning in~\cite{sommer2020towards,sommer2022athena}. Specifically, the setting begins with training a model that has a mixture of clean and poison data items across all users. Some of the users want their data deleted. If the users' data are not successfully deleted, the poison samples will be predicted as the target class. Otherwise, the model will not predict the poison samples as the target class. However, there is no absolute guarantee that this rule is always correct, although one can increase the number of poison samples to make this rule less likely to fail.

\sstitle{Slow-down attacks}
Some studies focus on the theoretical guarantee of indistinguishability between an unlearned and a retrained models. However, the practical bounds on computation costs are largely neglected in these papers~\cite{marchant2022hard}. As a result, a new threat has been introduced to machine unlearning where poisoning attacks are used to slow down the unlearning process. Formally, let $h_0 = A(D)$ be an initial model trained by a learning algorithm $A$ on a dataset $D$. The goal of the attacker is to poison a subset $D_{poison} \subset D$ such as to maximize the computation cost of removing $D_{poison}$ from $\hat{h}$ using an unlearning algorithm $U$. Marchant et al. ~\cite{marchant2022hard} defined and estimated an efficient computation cost for certifying removal methods. However, generalizing this computation cost for different unlearning methods is still an open research direction.

\sstitle{Interclass Confusion Test}
The idea of this test is to investigate whether information from the data to be forgotten can still be inferred from an unlearned model~\cite{goel2022evaluating}. Different from traditional approximate unlearning definitions that focus on the indistinguishability between unlearned and retrained models in the parameter space, this test focuses on the output space. More precisely, the test involves randomly selecting a set of samples $S \subset D$ from two chosen classes in the training data $D$ and then randomly swapping the label assignment between the samples of different classes to result in a confused set $S'$. Together $S'$ and $D \setminus S$ form a new training dataset $D'$, resulting in a new trained model. $S'$ is considered to be the forgotten data. From this, Goet et al.~\cite{goel2022evaluating} computes a forgetting score from a confusion matrix generated by the unlearned model. A lower forgetting score means a better unlearned model.

\sstitle{Federated verification}
Unlearning verification in federated learning is uniquely challenging. First, the participation of one or a few clients in the federation may subtly change the global model's performance, making verification in the output space challenging. Second, verification using adversarial attacks is not applicable in the federated setting because it might introduce new security threats to the infrastructure~\cite{gao2022verifi}. As a result, Gao et al.~\cite{gao2022verifi} proposes a verification mechanism that uses a few communication rounds for clients to verify their data in the global model. This approach is compatible with federated settings because the model is trained in the same way where the clients communicate with the server over several rounds.

\sstitle{Cryptographic proofs}
Since most of existing verification frameworks do not provide any theoretical guarantee, Eisenhofer et al.~\cite{eisenhofer2022verifiable} proposed a cryptography-informed protocol to compute two proofs, i.e. proof of update (the model was trained on a particular dataset $D$) and proof of unlearning (the forget item $d$ is not a member of $D$). The core idea of the proof of update is using SNARK~\cite{bitansky2012extractable} data structure to commit a hash whenever the model is updated (learned or unlearned) while ensuring that: (i) the model was obtained from the remaining data, (ii) the remaining data does not contain any forget items, (iii) the previous forget set is a subset of the current forget set, and (iv) the forget items are never re-added into the training data. The core idea of the proof of unlearning is using the Merkle tree to maintain the order of data items in the training data so that an unlearned item cannot be added to the training data again. While the approach is demonstrated on SISA (efficient retraining)~\cite{bourtoule2021machine}, it is applicable for any unlearning method.

\section{Unlearning Definition}
\label{sec:problem}

\subsection{Problem Formulation}
\label{sec:formulation}

While the application of machine unlearning can originate from security, usability, fidelity, and privacy reasons, it is often formulated as a privacy preserving problem where users can ask for the removal of their data from computer systems and machine learning models~\cite{sekhari2021remember,ginart2019making,bourtoule2021machine,garg2020formalizing}.
The forgetting request can be motivated by security and usability reasons as well. 
For example, the models can be attacked by adversarial data and produce wrong outputs. Once these types of attacks are detected, the corresponding adversarial data has to be removed as well without harming the model's predictive performance.

When fulfilling a removal request, the computer system needs to remove all user's data and `forget' any influence on the models that were trained on those data. As removing data from a database is considered trivial, the literature mostly concerns how to unlearn data from a model~\cite{GuoGHM20,izzo2021approximate,neel2021descent,ullah2021machine}.

To properly formulate an unlearning problem, we need to introduce a few concepts. First, let us denote $\mathcal{Z}$ as an example space, i.e., a space of data items or examples (called samples). Then, the set of all possible training datasets is denoted as $\mathcal{Z}^*$. One can argue that $\mathcal{Z}^* = 2^{\mathcal{Z}}$ but that is not important, as a particular training dataset $D 
\in Z^*$ is often given as input. Given $D$, we want to get a machine learning model from a hypothesis space $\mathcal{H}$. In general, the hypothesis space $\mathcal{H}$ covers the parameters and the meta-data of the models. Sometimes, it is modeled as $\mathcal{W} \times \Theta$, where $\mathcal{W}$ is the parameter space and $\Theta$ is the metadata/state space. The process of training a model on $D$ in the given computer system is enabled by a learning algorithm, denoted by a function $A: \mathcal{Z}^* \rightarrow \mathcal{H}$, with the trained model denoted as $A(D)$.

To support forgetting requests, the computer system needs to have an unlearning mechanism, denoted by a function $U$, that takes as input a training dataset $D \in Z^*$, a forget set $D_f \subset D$ (data to forget) and a model $A(D)$. It returns a sanitized (or unlearned) model $U(D, D_f, A(D)) \in \mathcal{H}$.
The unlearned model is expected to be the same or similar to a retrained model $A(D \setminus D_f)$ (i.e., a model as if it had been trained on the remaining data). Note that $A$ and $U$ are assumed to be randomized algorithms, i.e., the output is non-deterministic and can be modelled as a conditional probability distribution over the hypothesis space given the input data~\cite{marchant2022hard}. This assumption is reasonable as many learning algorithms are inherently stochastic (e.g., SGD) and some floating-point operations involve randomness in computer implementations~\cite{bourtoule2021machine}.
Another note is that we do not define the function $U$ precisely before-hand as its definition varies with different settings.

\autoref{tab:symbols} summarizes important notations.

\begin{table}
\centering
\caption{Important notations}
\label{tab:symbols}
\vspace{-1em}
\footnotesize
 \begin{adjustbox}{max width=0.45\textwidth}
\begin{tabular}{c l}
\toprule
\textbf{Symbols} & \textbf{Definition} \\
\midrule
$\mathcal{Z}$ & example space \\
$D$  & the training dataset \\
$D_f$  & forgetting set (the data to be forgotten) \\ 
$D_r = D \setminus D_f$  & retained set (the remaining data) \\ 
\midrule
$A(.)$ & a learning algorithm \\
$U(.)$ & an unlearning algorithm \\
\midrule
$\mathcal{H}$ & hypothesis space of models \\
%$M_0$ & the original model \\
%$M_t$ & the model after $t$-th unlearning request\\
%$w(M)$ or $w$ & parameters of the model \\
$w = A(D)$ & Parameters of the model trained on $D$ by $A$ \\
$w_r = A(D_r)$ & Parameters of the model trained on $D_r$ by $A$ \\
$w_u = U(.)$ & Parameters of the model unlearned by $U(.)$ \\
\bottomrule
\end{tabular}
\end{adjustbox}
\vspace{-1em}
\end{table}

\subsection{Exact Unlearning (Perfect Unlearning)}
\label{sec:exact_unlearning}

The core problem of machine unlearning involves the comparison between two distributions of machine learning models~\cite{thudi2022unrolling,bourtoule2021machine,brophy2021machine}. 
Let $Pr(A(D))$ define the distribution of all models trained on a dataset $D$ by a learning algorithm $A(.)$. 
Let $Pr(U(D, D_f, A(D)))$ be the distribution of unlearned models. The reason why the output of $U(.)$ is modelled as a distribution rather than a single point is that learning algorithms $A(.)$ and unlearning algorithms $U(.)$ are randomized as mentioned above.

\begin{dfn}[Exact unlearning - special case]
\label{def:exact_special}
Given a learning algorithm $A(.)$, a dataset $D$, and a forget set $D_f \subseteq D$, we say the process $U(.)$ is an exact unlearning process iff:
\begin{equation}
\label{eq:exact_special}
	Pr(A(D \setminus D_f)) = Pr(U(D, D_f, A(D)))
\end{equation}
\end{dfn}

Two key aspects can be drawn from this definition. First, the definition does not require that the model $A(D)$ be retrained from scratch on $D \setminus D_f$. Rather, it requires some evidence that it is likely to be a model that is trained from scratch on $D \setminus D_f$.
Second, two models trained with the same dataset should belong to the same distribution. However, defining this distribution is tricky. So to avoid the unlearning algorithm being specific to a particular training dataset, we have a more general definition~\cite{ginart2019making,brophy2021machine}:

\begin{dfn}[Exact unlearning - general case]
\label{def:exact_general}
Given a learning algorithm $A(.)$, we say the process $U(.)$ is an exact unlearning process iff $\forall \mathcal{T} \subseteq \mathcal{H}, D \in Z^*, D_f \subset D$:
\begin{equation}
\label{eq:exact_general}
	Pr(A(D \setminus D_f) \in \mathcal{T}) = Pr(U(D, D_f, A(D)) \in \mathcal{T})
\end{equation}
\end{dfn}

This definition allows us to define a metric space the models belong to (and consequently for the distributions). A model can be viewed either as just a mapping of inputs to outputs in which case $Pr(.)$ are distributions over a function space (i.e., continuous function with the supremum metric), or as the specific parameters $\bm{\theta}$ for a model architecture, in which case $Pr(.)$ are distributions over the weight space (e.g., some finite dimensional real vector space with the Euclidean norm). This ambiguity leads to two notions of exact unlearning:

\begin{itemize}
	\item \emph{Distribution of weights}: \autoref{eq:exact_general} implies the zero difference in the distribution of weights, i.e., $Pr(w_r) = Pr(w_u)$,
where the parameters of models $w_r$ learned by $A(D_r)$ and $w_u$ are the parameters of the models given by $U(.)$.
\item \emph{Distribution of outputs:} \autoref{eq:exact_general} implies zero difference in the distribution of outputs, i.e., $Pr(M(X; w_r))$ = $Pr(M(X; w_u))$, $\forall X \subseteq \mathcal{Z}$, where $M(.)$ is the parameterized mapping function from the input space $\mathcal{Z}$ to the output space (i.e., the machine learning model). This definition is sometimes referred to as \emph{weak unlearning}~\cite{baumhauer2020machine}.
\end{itemize}

If the unlearning mechanism $U(.)$ is implemented as retraining itself, equality is absolutely guaranteed. For this reason, retraining is sometimes considered to be the only exact unlearning method.
However, retraining inherently involves high computation costs, especially for large models~\cite{thudi2022unrolling}. Another disadvantage of retraining is that it cannot deal with batch settings, where multiple removal requests happen simultaneously or are grouped in a batch. 

There are many different metrics for comparing numerical distributions over the output space and the weight space. However, doing so is expensive (e.g., generating a sample in these distributions involves training the whole model). To mitigate this issue, some approaches design an alternative metric on a point basis to compute the distance between two models, either in the output space or in the weight space~\cite{shokri2017membership}.

\subsection{Approximate Unlearning (Bounded/Certified Unlearning)}

Approximate unlearning approaches attempt to address these cost-related constraints. In lieu of retraining, these strategies: perform computationally less costly actions on the final weights~\cite{GuoGHM20,graves2021amnesiac,sekhari2021remember}; modify the architecture~\cite{baumhauer2020machine}; or filter the outputs~\cite{baumhauer2020machine}. Approximate unlearning relaxes \autoref{def:exact_general} as follows~\cite{GuoGHM20}. 

\begin{definition}[$\epsilon$-Approximate Unlearning]
	Given $\epsilon > 0$, an unlearning mechanism $U$ performs $\epsilon$-certified removal for a learning algorithm $A$ if $\forall \mathcal{T} \subseteq \mathcal{H}, D \in Z^*, z \in D$:
	\begin{equation}
	\label{eq:approximate_epsilon}
e^{-\epsilon}\leq \frac{Pr( U(D, z, A(D)) \in \mathcal{T})}{Pr(A(D \setminus z) \in \mathcal{T})} \leq e^{\epsilon}
\end{equation}
where $z$ is the removed sample. 
\end{definition}
It is noteworthy that \autoref{eq:approximate_epsilon} defines the bounds on a single sample $z$ only. It is still an open question as to whether constant bounds can be provided for bigger subsets of $D$.
Moreover, the reason why we have the $[e^{-\epsilon}, e^{\epsilon}]$ bounds is that the probability distributions are often modeled by log functions, in which \autoref{eq:approximate_epsilon} is equivalent to:
\begin{equation}
	-\epsilon \leq \log \left[ Pr( U(D, z, A(D)) \in \mathcal{T}) - Pr(A(D \setminus z) \in \mathcal{T}) \right] \leq \epsilon
\end{equation}
or:
\begin{equation}
	\log || Pr( U(D, z, A(D)) \in \mathcal{T}) - Pr(A(D \setminus z) \in \mathcal{T}) || \leq \epsilon
\end{equation}
where $|| . ||$ is an absolute distance metric on the weight space or the output space.
A relaxed version of $\epsilon$-approximate unlearning is also defined in~\cite{neel2021descent}:

\begin{dfn}[($\epsilon$,$\delta$)-Approximate Unlearning]
	Given $\epsilon, \delta > 0$, an unlearning mechanism $U$ performs $\epsilon$-certified removal for a learning algorithm $A$ if $\forall \mathcal{T} \subseteq \mathcal{H}, D \in Z^*, z \in D$:
	\begin{equation}
Pr( U(D, z, A(D)) \in \mathcal{T}) \leq e^{\epsilon} {Pr(A(D \setminus z) \in \mathcal{T})} + \delta
\end{equation}
and
	\begin{equation}
{Pr(A(D \setminus z) \in \mathcal{T})} \leq e^{\epsilon}  Pr( U(D, z, A(D)) \in \mathcal{T}) + \delta
\end{equation}
\end{dfn}
In other words, $\delta$ upper bounds the probability for the max-divergence bound in \autoref{eq:approximate_epsilon} to fail.

%\info{Let $D$ be a fixed training dataset and let $A$ be a (randomized)
%learning algorithm that trains on $D$ and outputs a model
%$h \in H$, that is, $A : D \rightarrow H$. Randomness in $A$ induces a
%probability distribution over the models in the hypothesis
%set $H$. We would like to remove a training sample, $x \in D$,
%from the output of $A$.
%}

%\info{
%To this end, we define a data-removal mechanism $M$ that
%is applied to $A(D)$ and aims to remove the influence of
%x. If removal is successful, the output of $M$ should be
%difficult to distinguish from the output of $A$ applied on
%$D \setminus x$. Given $\epsilon > 0$, we say that removal mechanism $M$
%performs $\epsilon$-certified removal ($\epsilon$-CR) for learning algorithm
%$A$ if:
%}

\sstitle{Relationship to differential privacy}
Differential privacy states that:
\begin{equation}
	\forall \mathcal{T} \subseteq \mathcal{H}, D, D': e^{-\epsilon} \leq \frac{Pr(A(D) \in \mathcal{T})}{Pr(A(D \setminus z) \in \mathcal{T})} \leq e^\epsilon
\end{equation}
where $z$ is the removed sample. Differential privacy implies approximate unlearning: deleting the training data is not a concern if algorithm $A$ never memorises it in the first place~\cite{GuoGHM20}. However, this is exactly the contradiction between differential privacy and machine unlearning. If $A$ is differentially private for any data, then it does not learn anything from the data itself~\cite{bourtoule2021machine}. In other words, differential privacy is a very strong condition, and most differentially private models suffer a significant loss in accuracy even for large $\epsilon$~\cite{chaudhuri2011differentially,abadi2016deep}. 

%\info{where $D$ and $D'$ differ in only one sample. Since $D$ and
%$D \setminus x$ only differ in one sample, it is straightforward to
%see that differential privacy of $A$ is a sufficient condition
%for certified removal, viz., by setting removal mechanism
%$M$ to the identity function. Indeed, if algorithm $A$ never
%memorizes the training data in the first place, we need not
%worry about removing that data.
%}
%
%\info{Even though differential privacy is a sufficient condition,
%it is not a necessary condition for certified removal. For
%example, a nearest-neighbor classifier is not differentially
%private but it is trivial to certifiably remove a training sample
%in O(1) time with $\epsilon = 0$. We note that differential privacy
%is a very strong condition, and most differentially private models suffer a significant loss in accuracy even for large (Chaudhuri et al., 2011; Abadi et al., 2016). We therefore
%view the study of certified removal as analyzing the tradeoff between utility and removal efficiency, with re-training
%from scratch and differential privacy at the two ends of the
%spectrum, and removal in the middle.
%}
%\cite{GuoGHM20}
%
%
%\cite{neel2021descent}

\subsection{Indistinguishability Metrics}

To compare the two models in \autoref{def:exact_general}, we need to define a distance metric $d(.)$ between $Pr(A(D \setminus D_f) \in \mathcal{T})$ and $Pr(U(D, D_f, A(D)) \in \mathcal{T})$ ($\forall \mathcal{T} \subseteq \mathcal{H}$) in either the weight (parameter) space or the output space. To this end, several distance metrics have been studied:

\sstitle{$\ell_2$-distance}
Wu et al.~\cite{wu2020deltagrad} proposed using a Euclidean norm to compare the weights of $A(D_r)$ and the weights of $U(D, D_f, A(D))$. This is also termed as \emph{verification error}~\cite{wu2020deltagrad}. Despite being simple, this metric has several limitations:
(1) It is costly to compute this verification error as we need to also calculate $A(D_r)$ (through naive retraining). If the computational cost is cheap, machine unlearning is not necessary in the first place. 
(2) It is possible for two models having same training set and initialisation to have different weights~\cite{jia2021proof} due to training stochasticity and uncertainties in floating-point operations. Therefore, it is quite tricky to define a threshold for this error.

\sstitle{KL Divergence}
The Kullback-Leiber (KL) divergence or the Jensen-Shannon divergence is a popular distance metric between two distributions. Golatkar et al.~\cite{golatkar2020eternal} considered this divergence to measure the distance between two models in the parameter space. Although it might not require computing $A(D_r)$, it is necessary to have final distributions of models train on $D_r$ for computing the divergence. Distribution modeling is non-trivial and might involve sampling of many models trained on $D_r$ as well.

\sstitle{Weight Leakage}
Some studies measure privacy leaks of the removed sample $z$ from the parameters of the unlearned model~\cite{dwork2014algorithmic,sekhari2021remember,GuoGHM20}. These works assume that a model's weight distribution does not leak information about $z$ if it was not trained on $z$. However, measuring this privacy leakage is non-trivial and only well-defined for special classes of models. Guo et al.~\cite{GuoGHM20} proposed such a metric for linear classifiers via gradients.
More precisely, a model $w^*$ is trained on $D$ if the gradient $\nabla L(D; w^*) = 0$, where $L(.)$ is an empirical risk (loss) of a linear model. This is because $\argmin_w L(D; w)$ is uniquely determined for such $L(.)$. As a result, if the gradient $\nabla L(D \setminus z; w_u)$, where $w_u$ is the parameters of the model returned by $U$, is non-zero then the model either does not finish training or is not trained on $D \setminus z$. The former case can be safely ignored as we are not interested in non-converged models. However, the latter case indirectly implies that the model was trained on a dataset that included $z$, and thus it reveals some information about $z$. As a result, the gradient $\nabla L(D \setminus z; w_u)$ becomes an objective function for minimizing (or providing a bound) in those works~\cite{sekhari2021remember,GuoGHM20}.
Although this metric does provide an efficient way to verify the unlearning result, the above assumptions are not always correct from a numerical perspective, especially for non-linear models~\cite{gupta2021adaptive}. Using these metrics also requires access to the remaining data as well as the loss function, which are not always available.

\section{Other Unlearning Scenarios}
\label{sec:scenarios}

Beside the exact unlearning and approximate unlearning settings, there are other settings where access to the data, whether they are to be forgotten or retained, is restricted for security reasons.

\subsection{Zero-glance Unlearning}

Traditional privacy settings in machine unlearning assume there is access to all the training data before forgetting. Tarun et al. \cite{tarun2021fast} proposed to work in a stricter setting where once the user had made a request to forgetting their data (for example, their face in a facial recognition model), the organization could not use those samples even for the purposes of model weight manipulation.
Formally, the zero-glance setting means that the unlearning algorithm uses only the retained data:
\begin{equation}
w_u = U(D_{r\_sub}, A(D))
\end{equation}
where $w_u$ is the unlearned model and $D_{r\_sub} \subseteq D_r$ is a subset of samples drawn from the retained dataset $D_r = D \setminus D_f$. The smaller the $D_{r\_sub}$, the better the privacy.

\sstitle{Error-maximizing noise}
Unlearning without knowing the data to be forgotten is non-trivial. Tarun et al.~\cite{tarun2021fast} relaxed the setting somewhat by defining a set of classes to be forgotten, which should be completely removed from the organization. More precisely, given a sample $x$ and class label $y \in C$, and a set of classes $C = \{1, \ldots, K\}$, the training data $D$ is now a set of pairs $(x,y)$. Let $C_f$ denote the set of classes that needs to be forgotten by the model, and let $D_f$ be the data collection the classes to be forgotten belong to. The unlearned model is computed as:
$w_u = U(D_{r\_sub}, C_f, A(D))$.

Tarun et al.~\cite{tarun2021fast} proposed learning a noise matrix for the classes to be forgotten $C_f$ by maximizing the model's loss. In other words, a set of noise samples $\mathcal{N}$ are generated from $C_f$, $A(D)$ to approximate the data to be forgotten $D_f$. Then, the model $A(D)$ is trained for 1 epoch on $D_{r\_sub} 
\cup \mathcal{N}$ to damage the model parameters on the forgotten classes, thus inducing unlearning. After that, the impaired model is trained for 1 epoch on $D_{r\_sub}$ so as to repair any damage on the retained classes. Sometimes, it takes more epochs or a larger $D_{r\_sub}$ to enhance the unlearned model's performance. This is also a way to trade the advantages between retraining and unlearning (accuracy vs. time), which is not available in naive retraining.

In simple terms, existing methods for zero-glance unlearning try to transform the problem into its original form, where $w_u = U(D, D_f, A(D))$ by generating an approximate version of $D_f$.

\subsection{Zero-shot Unlearning}

If the training data is not accessible by an unlearning method $U$, the scenario becomes zero-shot unlearning. That is, the unlearning objective in \autoref{def:exact_general} becomes:
\begin{equation}
\label{eq:zero_shot}
Pr(A(D \setminus D_f) \in \mathcal{T}) \approx Pr(U(D_f, A(D)) \in \mathcal{T})
%w_u = U(D_f, A(D))
\end{equation}

However, it is non-trivial to solve this problem in generic cases. Similar to the zero-glance unlearning, Chundawat et al.~
\cite{chundawat2022zero} studied such a setting for classification with classes to be forgotten with the idea being that the outputs of the unlearned model should resemble those of a retrained model.
That is, the unlearning objective they want to achieve was: for some $X \subset \mathcal{Z}$:
\begin{equation}
M(X; w_u) \approx M(X; w_r)
\end{equation}
where $w_r = A(D \setminus D_f)$ and $w_u = U(C_f, A(D))$. Note that the data to be forgotten $D_f$ is not accessible to the unlearning algorithm $U(.)$. But, rather, as $C$ is a generally available information, $U(.)$ has access to the retained classes $C_r = C \setminus C_f$.

\sstitle{Error-minimization noise}
Chundawat et al.~\cite{chundawat2022zero} reused error-maximizing noise of \cite{tarun2021fast} to generate an impaired version of $D_f$ such that the model parameters of $A(D)$ are damaged when being trained on those noise samples. To handle the lack of access to retain data, Chundawat et al.~
\cite{chundawat2022zero} proposed error-minimizing noise to generate an approximate version of $D_r$ such that the impaired model could be trained to mitigate the performance degradation.

\sstitle{Gated knowledge transfer}
Unfortunately, the above error max-min approach yields poor unlearning outcomes as the generated noise is somewhat adhoc. Hence, inspired by~\cite{micaelli2019zero}, Chundawat et al.~\cite{chundawat2022zero} proposed a knowledge distillation mechanism with a crucial trick: a special type of `gate' is introduced to deny any knowledge of the classes to be forgotten $C_f$ so as to prevent a teacher model from passing knowledge to a student model.

\subsection{Few-shot Unlearning}

Few-shot unlearning is actually more similar to zero-glance setting rather than the zero-shot setting in that the unlearning algorithm only receives a small portion of the data to be forgotten $D_f$. More specifically, the unlearned model is computed as:
\begin{equation}
w_u = U(D, D_{f\_sub}, A(D))
\end{equation}
where $D_{f\_sub} \subseteq D_f$. This setting is useful for cases where the set to be forgotten $D_f$ contains mislabeled data or one wants to remove some data's malicious effects on a model. However, access to $D_f$ can be undermined ~\cite{yoon2022few} due to privacy regulations. Also, in most cases, $D_{f\_sub}$ is much smaller than $D_f$, i.e., $D_{f\_sub} \ll |D_f|$.

\sstitle{Model inversion}
Yoon et al.~\cite{yoon2022few} introduced a framework for few-shot unlearning on the basis of model inversion. First, a proxy for the learning set is retrieved from the given model via a model inversion mechanism. Second, a filtration method eliminates some of the retrieved data that may be responsible for the undesirable behaviour while interpolating the few target samples. Finally, the filtered data is used within a relearning process.

The proposed framework even works on stricter setting where the unlearning process can not access the original training data, i.e., $w_u = U(D_{f\_sub}, A(D))$~\cite{yoon2022few} . However, it is only applicable to classification models with a cross-entropy loss.

\sstitle{Influence approximation}
Peste et al.~\cite{peste2021ssse} proposed an unlearning process based on influence functions. The existing efficient influence-based unlearning process needs the calculation and the inverse of the Hessian matrix of loss from the model to determine the impact of any sample. However, those mentioned operations on the Hessian matrix are extremely costly, especially for high-dimensional models. Peste et al.~\cite{peste2021ssse} approximated it by using an empirical Fisher Information Matrix (FIM), which uses rank-one updates to allow easy computation and fast matrix inversion. As only the to-be-removed data is accessible and erasing the samples is simple via closed-form updates, those approximations gain a significant practical advantage.

\section{Unlearning Algorithms}
\label{sec:algorithms}

As mentioned in the Section~\ref{sec:intro}, machine unlearning can remove data and data linkages without retraining the machine learning model from scratch, saving time and computational resources~\cite{wang2022federated,chen2021machinegan}. The specific approaches of machine unlearning can be categorized into model-agnostic, model-intrinsic, and data-driven approaches.

\begin{table*}[!h]
\centering
\caption{Comparison of unlearning methods}
\label{tab:unlearning_comparison}
\vspace{-1em}
 \begin{adjustbox}{max width=0.9\textwidth}
    \begin{threeparttable}
    \begin{tabular}{l|ccccc|cccccc|ccccc}
        \toprule
	\textbf{Unlearning Methods} &  \multicolumn{5}{c|}{\textbf{Unlearning Scenarios}} &  \multicolumn{6}{c|}{\textbf{Design Requirements}} & \multicolumn{5}{c}{\textbf{Unlearning Requests}} \\
	& \head{Exact} & \head{Approximate} & \head{Zero-glance} & \head{Zero-shot} & \head{Few-shot} & \head{Completeness} & \head{Timeliness} & \head{Accuracy} & \head{Lightweight} & \head{Guarantees} & \head{Verifiability} & \head{Item} & \head{Feature} & \head{Class} & \head{Task} & \head{Stream}\\
	\midrule
	{\textbf{Model-agnostic}} & \multicolumn{9}{c}{}\\
	\hline
	Differential privacy~\cite{gupta2021adaptive,fraboni2024sifu} & \xmark  & \cmark & -- & -- & -- & \cmark & \cmark  & -- & \cmark & \cmark & -- & \cmark & \xmark & \xmark & \xmark & \cmark\\
	Certified removal~\cite{GuoGHM20,golatkar2020eternal,neel2021descent,ullah2021machine} & -- & \cmark & \xmark & \xmark & -- & -- & \cmark & \cmark & \cmark & \cmark & --   & \cmark  & \xmark & \xmark & \xmark & \cmark\\
	Statistical query learning~\cite{cao2015towards} & --  & \cmark  & \xmark & -- & \xmark & \cmark & \cmark & -- & \cmark & -- & -- & \cmark & \xmark & \xmark & \xmark & -- \\
	Decremental learning~\cite{ginart2019making,chen2019novel} & \xmark & --  & \xmark & -- & -- & \xmark & \cmark & \cmark & -- & -- & -- & \cmark & \xmark & \xmark & \xmark & \xmark \\
	Knowledge adaptation~\cite{chundawat2022can,kim2024layer,zhang2023machine,wang2023kga} & \xmark  & \cmark & -- & -- & -- & -- & -- & -- & \xmark & \xmark & -- & \cmark & -- & -- & -- & -- \\
	Parameter sampling~\cite{nguyen2022markov} & \xmark & \cmark & \cmark & \xmark & --  & -- & -- & -- & -- & -- & -- & \cmark & \xmark & \xmark & \xmark & \xmark \\
	\midrule
	{\textbf{Model-intrinsic}} & \multicolumn{9}{c}{}\\
	\hline
	Softmax classifiers~\cite{baumhauer2020machine} & \xmark & \cmark  & \cmark & \xmark & -- & \cmark & --   & --  &  \cmark & \xmark & \cmark & \xmark & \xmark & \cmark & \xmark & \xmark \\
  Linear models~\cite{izzo2021approximate,li2020online} & \cmark & \xmark & \xmark & -- & \xmark & --  & \cmark & -- & \cmark & \cmark & \cmark & \cmark & \xmark & \xmark & \xmark & \xmark \\
	Tree-based models~\cite{schelter2021hedgecut,wu2023deltaboost} & \xmark  & --  & \xmark & -- & \xmark & \xmark & -- & \cmark & \xmark & \xmark & \xmark & \cmark & \xmark & \xmark & \xmark & \xmark \\
	Bayesian models~\cite{nguyen2020variational} & -- & \cmark  & \xmark & \xmark & \xmark & -- & -- & \cmark & -- & \cmark & -- & \cmark & \xmark & \xmark & \xmark & \xmark \\
	DNN-based models~\cite{he2021deepobliviate,goyal2021revisiting,mehta2022deep,golatkar2021mixed,
	golatkar2020forgetting,basu2021influence,zhang2022machine} & \xmark & \cmark & \xmark & \xmark & \xmark & - & - & \cmark & - & - & - & \cmark & \xmark & \xmark & \xmark & \xmark \\
	\midrule
	{\textbf{Data-driven}} & \multicolumn{9}{c}{}\\
	\hline
	Data partition~\cite{bourtoule2021machine,aldaghri2021coded,dukler2023safe} & \cmark  & \xmark  & \cmark & \xmark & \cmark & \cmark & \xmark & -- & \xmark & \cmark & \cmark & \cmark & \xmark & -- & \xmark & -- \\
	Data augmentation~\cite{huang2021unlearnable,shan2020protecting,tarun2021fast,yu2021does} & \xmark & \cmark  & \xmark & \xmark & \xmark & -- & -- & --  & -- & \xmark & -- & -- & \xmark & \cmark & \xmark & \xmark \\
	Data influence~\cite{peste2021ssse,zeng2021learning,cao2022machine} & \xmark & \cmark & \xmark & \cmark & -- & -- & \cmark & -- & \cmark & \cmark & -- & \cmark & \xmark & \xmark & \xmark & \xmark \\
         \bottomrule
    \end{tabular}
    \begin{tablenotes}
\item \cmark: fully support
\item \xmark: no support
\item --: partially or indirectly support
\item[] []: representative citations
\end{tablenotes}
    \end{threeparttable}
    \end{adjustbox}
    
%\begin{adjustbox}{max width=\textwidth}
%\begin{tabular}{c p{3cm} p{4cm} p{3cm}}
%\toprule
%\textbf{Method} & \textbf{Training Algorithm} & \textbf{Unlearning Algorithm} & \textbf{Parameters} \\
%\midrule
%Fisher~\cite{golatkar2020eternal} & & & \\
%Influence~\cite{GuoGHM20} & & & \\
%DeltaGrad~\cite{wu2020deltagrad} & & & \\
%\bottomrule
%\end{tabular}
%\end{adjustbox}
\vspace{-1em}
\end{table*}

\subsection{Model-Agnostic Approaches}
\label{sec:model-agnostic}

Model-agnostic machine unlearning methodologies include unlearning processes or frameworks that are applicable to different models. However, in some cases, theoretical guarantees are only provided for a class of models (e.g., linear models). Nonetheless, they are still considered to be model-agnostic as their core ideas are applicable to complex models (e.g. deep neural networks) with practical results.

% \info{
% To remove the ith category from a model trained with an integrated dataset D,
% a vanilla way would be to retrain the model from scratch on the
% remaining data points D-Di
% . However, the time, computation, and
% energy costs of model retraining can be quite costly. To address
% this problem, machine unlearning has been recently studied, which
% typically produces an approximation of the fully retrained model at
% low cost.
% }
% \cite{wang2022federated}

\sstitle{Differential Privacy}
Differential privacy was first proposed to bound a data sample's influence on a machine learning model~\cite{dwork2008differential}. $\epsilon$-differential privacy unlearns a data sample by setting $\epsilon=0$, where $\epsilon$ bounds the level of change in any model parameters affected by that data sample~\cite{bourtoule2021machine,thudi2022unrolling}. However, Bourtoule et al.~\cite{bourtoule2021machine} notes that the algorithm cannot learn from the training data in such a case. 
Gupta et el.~\cite{gupta2021adaptive} proposed a differentially private unlearning mechanism for streaming data removal requests. These requests are adaptive as well, meaning the data to be removed depends on the current unlearned model. The idea, which is based on differential privacy, can be roughly formulated as:
\begin{equation}
	\Pr( U(D, s, A(D)) \in \mathcal{T}) \leq e^\epsilon Pr(A(D \setminus s) \in \mathcal{T}) + \beta
\end{equation}
for all adaptive removal sequences $s = (z_1, \ldots, z_k)$. One weakness of this condition is that it only guarantees the upper bound of the unlearning scheme compared to full retraining. However, its strength is that it supports a user's belief that the system has engaged in full retraining. Finally, an unlearning process is developed by a notion of differentially private publishing functions and a theoretical reduction from adaptive to non-adaptive sequences. Differentially private publishing functions guarantee that the model before and after an unlearning request do not differ too much~\cite{fraboni2024sifu}.

\sstitle{Certified Removal Mechanisms}
Unlearning algorithms falling into this category are the ones following the original approximate definition of machine unlearning~\cite{GuoGHM20,golatkar2020eternal}. While Guo et al.~\cite{GuoGHM20} focus on theoretical guarantees for linear models and convex losses, Golatkar et al.~\cite{golatkar2020eternal} introduce a computable upper bound for SGD-based learning algorithms, especially deep neural networks. The core idea is based on the notion of perturbation (noise) to mask the small residue incurred by the gradient-based update (e.g., a one-step Newton update~\cite{koh2017understanding}). The idea is applicable to other cases, although no theoretical guarantees are provided~\cite{bourtoule2021machine}.

More precisely, certified removal mechanisms mainly accommodate those linear models that minimize a standardized empirical risk, which is the total value of a convex loss function that measures the distance of the actual value from the expected one~\cite{marchant2022hard}. However, one has to rely on a customized learning algorithm that optimizes a perturbed version of the regularized empirical risk, where the added noise is drawn from a standard normal distribution. This normalized noise allows conventional convex optimization techniques to solve the learning problem with perturbation. As a result, the unlearning request can be done by computing the model perturbation towards the regularized empirical risk on the remaining data. The final trick is that this perturbation can be approximated by the influence function~\cite{koh2017understanding}, which is computed by inverting the Hessian on training data and the gradient of the data to be forgotten~\cite{marchant2022hard}. However, the error of model parameters in such a computation can be so large that the added noise cannot mask it. Therefore, if the provided theoretical upper bound exceeds a certain threshold, the unlearning algorithm resorts to retraining from scratch~\cite{marchant2022hard}.

Following this idea, Neel et al.~\cite{neel2021descent} provided further extensions, namely regularized perturbed gradient descent and distributed perturbed gradient descent, to support weak convex losses and provide theoretical guarantees on indistinguishability, accuracy, and unlearning times.

Ullah et al.~\cite{ullah2021machine} continued studying machine unlearning in the context of SGD and streaming removal requests. They define the notation of total variation stability for a learning algorithm:
\begin{equation}
	\sup_{D,D': |D \setminus D'| + |D' \setminus D|} \Delta (A(D), A(D')) \leq \rho
\end{equation} 
where $\Delta(.)$ is the largest possible difference between the two probabilities such that they can assign to the same event, aka total variance distance~\cite{verdu2014total}. This is also a special case of the optimal transportation cost between two probability distributions~
\cite{lei2019geometric}. In other words, a learning algorithm $A(.)$ is said to be $\rho$-TV-stable if given any two training datasets $D$ and $D'$, as long as they have 1 common data item, the cost of transporting from the model distribution $A(D)$ to $A(D')$ is bounded by $\rho$. For any $1/n \leq \rho < \infty$, Ullah et al.~\cite{ullah2021machine} proved that there exists an unlearning process that satisfies exact unlearning at any time in the streaming removal request while the model accuracy and the unlearning time are bounded w.r.t. $\rho$.

%\info{
%Other mechanisms relax the
%definition of differential privacy to provide certificates of data
%removal. This includes two concurrent proposals~\cite{GuoGHM20,golatkar2020eternal}.
%The mechanism by Guo et al.~\cite{GuoGHM20} uses a one-step Newton
%update~\cite{koh2017understanding}. While such a mechanism introduces a small
%residue, this is masked by adding noise (similar to approaches
%in differential privacy)
%}
%(copied from \cite{bourtoule2021machine})

%\todo{TODO: add more following the above references and}
%\cite{GuoGHM20,marchant2022hard,neel2021descent}

\sstitle{Statistical Query Learning} 
Statistical query learning is a form of machine learning that trains models by querying statistics on the training data rather than itself~\cite{cao2015towards}. In this form, a data sample can be forgotten efficiently by recomputing the statistics over the remaining data~\cite{bourtoule2021machine}.
More precisely, statistical query learning assumes that most of the learning algorithms can be represented as a sum of some efficiently computable transformations, called statistical queries~\cite{kearns1998efficient}. These statistical queries are basically requests to an oracle (e.g., a ground truth) to estimate a statistical function over all training data. Cao et al.~\cite{cao2015towards} showed that this formulation can generalize many algorithms for machine learning, such as the Chi-square test, naive Bayes, and linear regression. For example, in naive Bayes, these statistical queries are indicator functions that return 1 when the output is a target label and zero otherwise~\cite{cao2015towards}.
In the unlearning process, these queries are simply recomputed over the remaining data. The approach is efficient as these statistical functions are computationally efficient in the first place. Moreover, statistical query learning also supports adaptive statistical queries, which are computed based on the prior state of the learning models, including k-means, SVM, and gradient descent. Although this time, the unlearning update makes the model not convergent any more, only a few learning iterations (adaptive statistical queries) are needed since the model starts from an almost-converged state.
Moreover, if the old results of the summations are cached, say, via dynamic programming, then the speedup might be even higher.

The limitation of this approach is that it does not scale with complex models such as deep neural networks. Indeed, in complex models, the number of statistical queries could become exponentially large~\cite{bourtoule2021machine}, making both the unlearning and relearning steps less efficient.

In general, statistical query learning supports item removal and can be partially applied to stream removal~\cite{gupta2021adaptive} as well, although the streaming updates to the summations could be unbounded. It supports exact unlearning, but only partially when the statistical queries are non-adaptive. It also partially supports zero-shot unlearning, because only the statistics over the data need to be accessed, not the individual training data items.

% \info{More formally, the summation form follows statistical query
% (SQ) learning [48]. SQ learning forbids a learning algorithm
% from querying individual training data samples. Instead, it
% permits the algorithm to query only statistics about the training
% data through an oracle.}
% \cite{cao2015towards}

% \info{Cao et al. [15] model unlearning
% in the statistical query learning framework [16]. By doing so,
% they are able to unlearn a point when the learning algorithm
% queries data in an order decided prior to the start of learning.
% In this setting, it is possible to know exactly how individual
% training points contributed to model parameter updates. However, their approach is not general
% (G5) and does not easily
% scale to more complex models (such as those considered in
% this work). Indeed, these models are trained using adaptive
% statistical query algorithms which make queries that depend
% on all queries previously made. In this setting, the approach
% of Cao et al. [15] diverges in an unbounded way unless the
% number of queries made is small, which is not the case for
% the deep neural networks we experiment with.}
% \cite{bourtoule2021machine}

\sstitle{Decremental Learning}
Decremental learning algorithms were originally designed to remove redundant samples and reduce the training load on the processor for support vector machines (SVM)~\cite{chen2019novel,cauwenberghs2000incremental,tveit2003multicategory,tveit2003incremental,romero2007incremental,duan2007decremental} and linear classification~\cite{karasuyama2009multiple,karasuyama2010multiple,tsai2014incremental}.
As such, they focus on accuracy rather than the completeness of the machine unlearning.

Ginart et al.~\cite{ginart2019making} developed decremental learning solutions for $k$-means clustering based on quantization and data partition. The idea of quantization is to ensure that small changes in the data do not change the model. Quantization helps to avoid unnecessary unlearning so that accuracy is not catastrophically degraded. However, it is only applicable when there are few model parameters compared to the size of the dataset. The idea behind the data partitioning is to restrict the data's influence on the model parameters to only a few specific data partitions. This process helps to pinpoint the effects of unlearning to a few data features. But, again, the approach is only effective with a small number of features compared to the size of the dataset.
Notably, data privacy and data deletion are not completely correlative~\cite{ginart2019making}. Data privacy does not have to ensure data deletion (e.g., differential privacy), and data deletion does not have to ensure data privacy.

% \info{Ginart et al. [36] consider the
% problem from a data-protection regulation standpoint. They
% present a formal definition of complete data erasure which can
% be relaxed into a distance-bounded definition. Deletion time
% complexity bounds are provided. They note that the deletion
% and privacy problems are orthogonal, which means deletion
% capability does not imply privacy nor vice versa. However, it
% is unclear if the approach presented (Quantized k-Means) is
% applicable (G5) and scalable (G6) for all model classes}
% \cite{bourtoule2021machine}

\sstitle{Knowledge Adaptation (Knowledge Distillation)}
Knowledge adaptation selectively removes to-be-forgotten data samples~\cite{chundawat2022can}. In this approach~\cite{chundawat2022can}, one trains two neural networks as teachers (competent and incompetent) and one neural network as a student. The competent teacher is trained on the complete dataset, while the incompetent teacher is randomly initialised. The student is initialised with the competent teacher's model parameters. The student is trained to mimic both competent teacher and incompetent teacher by a loss function with KL-divergence evaluation values between the student and each of the two teachers. Notably, the competent teacher processes the retained data and the incompetent teacher deals with the forgotten data.

Beyond Chundwat et al.~\cite{chundawat2022can}, machine learning models have been quickly and accurately adapted by reconstructing the past gradients of knowledge-adaptation priors in~\cite{khan2021knowledge}. Ideas similar to knowledge-adaptation priors were also investigated in~\cite{ginart2019making,wu2020priu}. Lin et al.~\cite{lin2023erm} introduce ERM-KTP, a knowledge-level machine unlearning framework that leverages knowledge transfer to erase specific learned information while maintaining model performance. Wang et al.~\cite{wang2023kga} propose KGA, a general machine unlearning framework that aligns knowledge gaps between original and unlearned models for efficient data removal. Zhang et al.~\cite{zhang2023machine} present a stochastic teacher network method for unlearning, which trains the model to forget specific data points through stochastic processes. Kim et al.~\cite{kim2024layer} develop Layer Attack Unlearning, a fast and accurate method that uses layer-level attack and knowledge distillation to remove sensitive information from machine learning models. Lastly, Bonato et al.~\cite{bonato2024retain} investigate the impact of out-of-distribution images on unlearning models, proposing that retaining certain sets of data can help restore performance in unlearned models.
In general, knowledge adaptation is applicable to a wide range of unlearning requests and scenarios. However, it is difficult to provide a theoretical guarantee for this approach.

% \info{We aim to remove the information about the requested data-points by using two teachers (competent and incompetent) and one student. The student is initialized with knowledge about the complete data i.e., the parameters of the fully trained model. The idea is to selectively remove the information about the forget samples from this model. At the same time, the information pertaining to the retain set should not to be disturbed. Thus, the unlearning objec- tive is to remove the information about $D_f$ while retaining the information about $D_r$.
% }
% \cite{chundawat2022can}

% \info{[42] proposes to save all the intermediate weight parameters wt
% and gradients during training. Then, these information will be used to efficiently estimate
% the optimization path of strongly convex and smooth objective function after unlearning, which
% results in very limited applications. \cite{khan2021knowledge} proposes knowledge-adaptation priors to reduce the cost of
% retraining by enabling quick and accurate adaptation for a wide variety of tasks and models. Similar
% idea have been explored in \cite{ginart2019making} for K-means and \cite{wu2020priu} for logistic regression}
% \cite{conggrapheditor}

\sstitle{MCMC Unlearning (Parameter Sampling)}
Sampling-based machine unlearning has also been suggested as a way to train a standard machine learning model to forget data samples from the training data~\cite{nguyen2022markov}. The idea is to sample the distribution of model parameters using Markov chain Monte Carlo (MCMC). It is assumed that the forgetting set is often significantly smaller than the training data (otherwise retraining might be a better solution). Thus, the parameter distribution $Pr(w_r)$ of the retrained models should not differ much from that of the original model $Pr(w)$. In other words, the posterior density $Pr(w_r | D)$ should be sufficient large for sampling~\cite{nguyen2022markov}. 
More precisely, the posterior distribution from the retrained parameters can be defined as:
\begin{equation}
\label{eq:mcmc_unlearning}
	Pr(w_r | D) \approx Pr(w | D) \propto Pr(D | w) Pr(w)
\end{equation}
Here, the prior distribution $Pr(w)$ is often available from the learning algorithm, which means the stochasticity of learning via sampling can be estimated. The likelihood $Pr(D | w)$ is the prediction output of the model itself, which is also available after training. From \autoref{eq:mcmc_unlearning}, we only know that the density function of $Pr(w | D)$ is proportional to a function $f(w) = Pr(D | w) Pr(w)
$, which means $Pr(w | D)$ cannot be directly sampled. This is where MCMC comes into play, as it can still generate the next samples using a proposal density $g(w' | w)$~\cite{nguyen2022markov}. However, $g(w' | w)$ is assumed to be a Gaussian distribution centered on the current sample (the sampling process can be initialized with the original model).

As a result, a candidate set of model parameters $Pr(w_r | D)$ is constructed from the sampling, and the unlearning output is calculated by simply maximizing the posterior probability $Pr(w | D_r)$, i.e.:
\begin{equation}
	w_r = \argmax_w Pr(w | D_r)
\end{equation}
The benefit of such sampling-based unlearning is that no access to the forgetting set is required.

% \info{Let us consider the machine unlearning scenario where we would
% like to remove the effect of an erased dataset, }
% \cite{nguyen2022markov} 

%\todo{TODO: add more following the some of the above references}

\subsection{Model-Intrinsic Approaches}
The model-intrinsic approaches include unlearning methods designed for a specific type of models. Although they are model-intrinsic, their applications are not necessarily narrow, as many machine learning models can share the same type.

\sstitle{Unlearning for softmax classifiers (logit-based classifiers)}
Softmax (or logit-based) classifiers are classification models $M: \mathcal{Z} \rightarrow \mathds{R}^K$ that output a vector of logits $l \in \mathds{R}^k$, where $K$ is the number of classes, for each data sample $x \in \mathcal{Z}$. The core task of $M(x)$ is to estimate the probability distribution $Pr(X,Y)$, where $X$ is the random variable in $\mathcal{X}$, and $Y$ is the random variable in ${1, \ldots, K}$, such that:
\begin{equation}
	Pr(Y=i | X = x) \approx \sigma(l_i)
\end{equation}
Here, $\sigma(l_i) = \frac{\exp(l_i)}{\sum_{j=1..K} \exp l_j}$ is the softmax function. This formulation applies to logistic regression and deep neural networks with a densely connected output layer using softmax activations~\cite{baumhauer2020machine}.
Baumhauer et al.~\cite{baumhauer2020machine} proposed an unlearning method for softmax classifiers based on a linear filtration operator to proportionally shift the classification of the to-be-forgetten class samples to other classes. However, this approach is only works for class removal.

\sstitle{Unlearning for linear models}
Izzo et al.~\cite{izzo2021approximate} proposed an approximate unlearning method for linear and logistic models based on influence functions. They approximated a Hessian matrix computation with a project residual update~\cite{izzo2021approximate,cao2022machine} that combines gradient methods with synthetic data. It is suitable for forgetting small groups of points out of a learned model.
Some other studies consider an online setting for machine unlearning (aka online data deletion)~\cite{ginart2019making,li2020online}, in which the removal request is a sequence of entries that indicates which data item is to be unlearned. In general, this setting is more challenging than normal setting because indistinguishability must hold for any entry and for the end of the deletion sequence. The goal is to achieve a lower bound on amortized computation time~\cite{ginart2019making,li2020online}. 

Li et al.~\cite{li2020online} formulated a special case of the online setting where data is only accessible for a limited time so there is no full training process in the first place. More precisely, the system is allowed a constant memory to store historical data or a data sketch, and it has to make predictions within a bounded period of time. Although the data to be forgotten can be unlearned from a model on-the-fly using a regret scheme on the memory, this particular unlearning process is only applicable to ordinary linear regression~\cite{li2020online}.

%\info{The key components of this paper include introducing deletion efficient learning, based on an
%intuitive and operational notion of what it means to (efficiently) delete data from a (possibly stochastic) statistical model. We pose data deletion as an online problem, from which a notion of optimal deletion efficiency emerges from a natural lower bound on amortized computation time. We do a case-study on deletion efficient learning using the simple, yet perennial, k-means clustering problem}
%(copied from \cite{ginart2019making})
%
%\todo{TODO: rewrite and add more following the above references and}
%\cite{baumhauer2020machine,li2020online}

\sstitle{Unlearning for Tree-based Models}
Tree-based models are classification techniques that partition the feature space recursively, where the features and cut-off thresholds to split the data are determined by some criterion, such as information gain~\cite{schelter2021hedgecut,wu2023deltaboost}. There is a class of tree-based models, called extremely randomized trees~\cite{geurts2006extremely}, that are built by an ensemble of decision trees. These are very efficient because the candidate set of split features and cut-off thresholds are randomly generated. The best candidate is selected by a reduction in Gini impurity, which avoids the heavy computation of logarithms.

Schelter et al.~\cite{schelter2021hedgecut} proposed an unlearning solution for extremely randomized trees by measuring the robustness of the split decisions. A split decision is robust if removing $k$ data items does not reverse that split. Note that $k$ can be bounded, and it is often small as only one in ten-thousand users who wants to remove their data at a time~\cite{schelter2021hedgecut}). The learning algorithm is redesigned such that most of splits, especially the high-level ones, are robust. For the non-robust splits, all subtree variants are grown from all split candidates and maintained until a removal request would revise that split. When that happens, the split is switched to its variant with higher Gini gain. As a result, the unlearning process involves recalculating the Gini gains and updating the splits if necessary.

One limitation of this approach is that if the set to be forgotten is too large, there might be many non-robust splits. This would lead to high storage costs for the subtree variants. However, it does give a parameterized choice between unlearning and retraining. If there are many removal requests, retraining might be the best asymptotically. Alternatively, one might limit the maximum number of removal requests to be processed at a time. Moreover, tree-based models have a highly competitive performance for many predictive applications~\cite{schelter2021hedgecut}.

%\todo{TODO: add more following the references}

\sstitle{Unlearning for Bayesian Models}
Bayesian models are probabilistic models that approximate a posterior likelihood~\cite{fu2022knowledge,fu2021bayesian,jose2021unified,nguyen2020variational}. Also known as Bayesian inference, this process is particularly useful when a loss function is not well-defined or does not even exist. Bayesian models cover a wide range of machine learning algorithms, such as Bayesian neural networks, probabilistic graphical models, generative models, topic modeling, and probabilistic matrix factorization~\cite{zhang2020deep,roth2018bayesian,pearce2020uncertainty}. 

Unlearning for Bayesian models requires a special treatment, as the training already involves optimizing the posterior distribution of the model's parameters. It also often involves optimizing the Kullback-Leibler divergence between a prior belief and the posterior distribution~\cite{nguyen2020variational}. Nguyen et al.~\cite{nguyen2020variational} proposed the notion of \emph{exact Bayesian learning}:
\begin{equation}
	Pr(w | D_r) = Pr(w | D) Pr(D_f | D_r) / Pr(D_f | w) \propto Pr(w | D) / Pr(D_f | w)
\end{equation}
where $Pr (w | D_r)$ is the distribution of a retrained model (as if it were trained only on $D_r$). However, the posterior distribution $Pr(w | D_r)$ can only be sampled directly when the model parameters are discrete-valued (quantized) or the prior is conjugate~\cite{nguyen2020variational}. For non-conjugate priors, Nguyen et al.~\cite{nguyen2020variational} proved that we can approximate $Pr (w | D_r)$ by minimizing the KL divergence between $Pr(w | D)$ and $Pr (w | D_r)$. Since $Pr(w | D)$ is the original model's parameter distribution, this approximation prevents catastrophic unlearning. As such, the retained model performs significantly better than the unlearned model in terms of accuracy.

A notion of certified Bayesian unlearning has also been studied, where the KL divergence between the unlearned model and the retrained model is bounded~\cite{fu2022knowledge,fu2021bayesian,jose2021unified}:
\begin{equation}
	KL ( Pr(A(D_r)) , \mathop{\mathbb{E}}_{A(D)} Pr(U(D, D_f, A(D))) ) \leq \epsilon
\end{equation}
Here, the result of the unlearning process is an expectation over the parameter distribution of the original model $A(D) \sim Pr(w | D)$. This certification can be achieved for some energy functions when formulating the evidence lower bound (ELBO) in Bayesian models~\cite{fu2022knowledge,fu2021bayesian,jose2021unified}.

\sstitle{Unlearning for DNN-based Models}
Deep neural networks are advanced models that automatically learn features from data. As a result, it is very difficult to pinpoint the exact model update for each data item~\cite{golatkar2020forgetting,golatkar2020eternal,mehta2022deep,he2021deepobliviate,goyal2021revisiting}. Fortunately, deep neural networks consist of multiple layers. For layers with convex activation functions, existing unlearning methods such as certified removal mechanisms can be applied~\cite{GuoGHM20,neel2021descent,sekhari2021remember,cao2022machine}. For non-convex layers, Golatkar et al.~\cite{golatkar2021mixed,golatkar2020forgetting} proposed a caching approach that trains the model on data that are known a priori to be permanent. Then the model is fine-tuned on user data using some convex optimization.

%\info{Deterministic methods and methods that depend only on randomness that is sampled after the deletion request are already robust to adaptive deletion. This includes techniques that find an approximately optimal solution to a strongly convex problem and then perturb the solution to obscure the optimizer within a small radius e.g. Guo et al.~\cite{GuoGHM20}, Neel et al.~\cite{neel2021descent}, Sekhari et al.~\cite{sekhari2021remember}. It also includes the approach of Golatkar et al.~\cite{golatkar2021mixed,golatkar2020forgetting} which pre-trains a nonconvex model on data that will never be deleted and then does convex fine-tuning on user data on top of that.}
%(copied from \cite{gupta2021adaptive})

Sophisticated unlearning methods for DNNs rely primarily on influence functions~\cite{koh2017understanding,zhang2022machine}. Here, Taylor expansions are used to approximate the impact of a data item on the parameters of black-box models~\cite{zeng2021learning}. Some variants include DeltaGrad~\cite{wu2020deltagrad}, which stores the historical updates for each data item, and Fisher-based unlearning~\cite{golatkar2020eternal}, which we discussed under \autoref{sec:model-agnostic}). However, influence functions in deep neural networks are not stable with a large forget set~\cite{basu2021influence,mahadevan2021certifiable,mahadevan2022certifiable}.

More precisely, after the data to be forgotten has been deleted from database, Fisher-based unlearning~\cite{golatkar2020eternal} works on the remaining training data with the Newton's method, which uses a second-order gradient. To mitigate potential information leaks, noise is injected into the model's parameters~\cite{conggrapheditor}. 
As the Fisher-based method aims to approximate the model without the deleted data, there can be no guarantee that all the influence of the deleted data has been removed. Although injecting noise can help mitigate information leaks, the model's performance may be affected by the noise~\cite{conggrapheditor}.

Golatkar et al.~\cite{golatkar2020eternal} point out that the Hessian computation in certified removal mechanisms is too expensive for complex models like deep neural networks. Hence, they resorted to an approximation of Hessian via Levenberg-Marquardt semi-positive-definite approximation, which turns out to correspond with the Fisher Information Matrix~\cite{martens2020new}. Although it does not provide a concrete theoretical guarantee, Fisher-based unlearning could lead to further information-theoretic approaches to machine unlearning~\cite{guo2022efficient,golatkar2020forgetting}.

%\info{In deep neural networks, influence function~\cite{koh2017understanding} is proposed to measure the effect of a manipulated data point by using Taylor expansion to approximate model parameters. DeltaGrad~\cite{wu2020deltagrad} saves optimizer's update steps in order to more accurately approximate the removal of multiple sample points. Golatkar et al.~\cite{golatkar2020eternal} used the approximation of the Fisher Information Matrix for the remaining data sample to measure the update step to modify the model parameters. Although all techniques are capable of efficient model parameter approximation for a change with small number of samples~\cite{basu2021influence,mahadevan2021certifiable,mahadevan2022certifiable}, they are not scalable for situations when a large number of training points are added, deleted, or altered.}
%(copied from \cite{zeng2021learning})
%
%\todo{TODO: add more following the above references and}
%\cite{golatkar2020forgetting,golatkar2020eternal,mehta2022deep,he2021deepobliviate,goyal2021revisiting}

\subsection{Data-Driven Approaches}

%\subsubsection{Efficient Retraining (Smart Retraining)}

\begin{figure}[!h]
	\centering
%	\vspace{-1em}
	\includegraphics[width=1.0\linewidth]{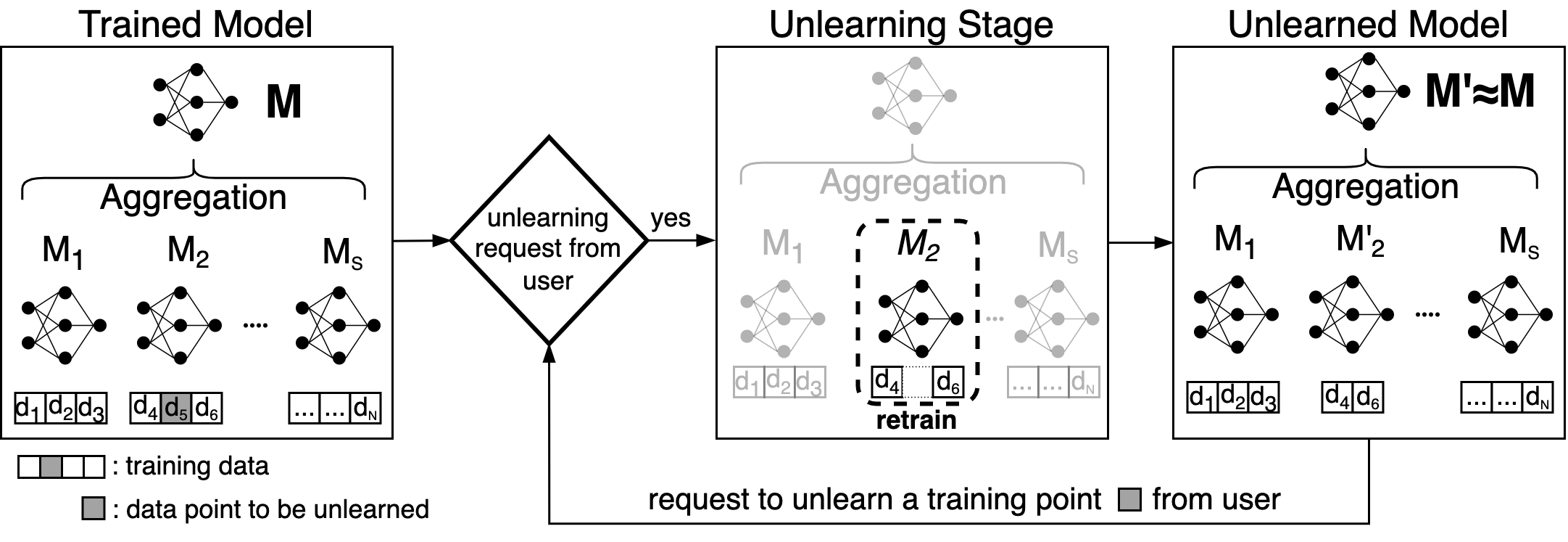}
%	\vspace{-1em}
	\caption{Efficient retraining for machine unlearning using data partition}
	\label{fig:partition}
%	\vspace{-1em}
\end{figure}

\sstitle{Data Partitioning (Efficient Retraining)}
The approaches falling into this category uses data partitioning mechanisms to speed up the retraining process. Alternatively, they partially retrain the model with some bounds on accuracy. Bourtoule et al.~\cite{bourtoule2021machine} proposed the well-known SISA framework (\autoref{fig:partition}) that partitions the data into shards and slices. Each shard has a single model, and the final output is an aggregation of multiple models over these shards. For each slice of a shard, a model checkpoint is stored during training so that a new model can be retrained from an intermediate state~\cite{bourtoule2021machine,aldaghri2021coded}.
Dukler et al.~\cite{dukler2023safe} partition the training data into disjoint shards and builds a graph structure to track dependencies between shards and model updates. When data is requested to be forgotten, the system efficiently removes the effects of specific shards by rolling back and recalculating model updates only for affected shards.

\sstitle{Data Augmentation (Error-manipulation noise)}
Data augmentation is the process of enriching or adding more data to support a model's training~\cite{yu2021does}. Such mechanisms can be used to support machine unlearning as well.
Huang et al.~\cite{huang2021unlearnable} proposed the idea of error-minimizing noise, which tricks a model into thinking that there is nothing to be learned from a given set of data (i.e., the loss does not change). However, it can only be used to protect a particular data item before the model is trained. A similar setting was also studied by Fawkes~\cite{shan2020protecting}, in which a targeted adversarial attack is used to ensure the model does not learn anything from a targeted data item.

Conversely, Tarun et al.~\cite{tarun2021fast} proposed error-maximizing noise to impair the model on a target class of data (to be forgotten). However, this tactic does not work on specific data items as it is easier to interfere with a model's prediction on a whole class as opposed to a specific data item of that class~\cite{tarun2021fast}.

%\sstitle{Error-Minimizing Noise}
%\info{In this paper, we consider a more challenging scenario where the goal of the defender is to make personal data completely unusable by unauthorized deep learning models. Fawkes~\cite{shan2020protecting} has made the first attempt towards this type of strict situation. By leveraging the targeted adversarial attack, Fawkes prevents unauthorized face tracker from tracking a person's identity. This work is similar to ours as we share a common objective that prevents unauthorized data usage. In contrast to the targeted adversarial attack, we propose a novel error-minimizing noise to produce unlearnable examples which can be used as a generic framework for a wide range of data protection tasks.}
%(copied from \cite{huang2021unlearnable})

%\info{We learn an error-maximizing noise for the respective unlearning classes. UNSIR is proposed to perform single-pass impair and single-pass repair b using very high learning rate. The impair step makes the network forget the unlearning data classes. The repair step stabilizes the network weights to better remember the re- maining tasks. The combination of both the steps al- low it to obtain excellent unlearning and retain accuracy.}
%(copied from \cite{tarun2021fast})

%\todo{TODO: rewrite and add more following the above references}

\sstitle{Data influence}
This group of unlearning approaches studies how a change in training data impacts a model's parameters~\cite{wu2022puma,conggrapheditor,cao2022machine}, where impact is computed using influence functions~\cite{mahadevan2022certifiable,chundawat2022zero}. However, influence functions depend on the current state of a learning algorithm~\cite{wu2022puma}. To mitigate this issue, several works store a training history of intermediate quantities (e.g., model parameters or gradients) generated by each step of model training~\cite{graves2021amnesiac,neel2021descent,wu2020deltagrad,wu2020priu}. Then, the unlearning process becomes one of subtracting these historical updates. However, the model's accuracy might degrade significantly due to catastrophic unlearning~\cite{nguyen2020variational} since the order in which the training data is fed matters to the learning model. Moreover, the influence itself does not verify whether the data to be forgotten is still included in the unlearned model~\cite{thudi2021necessity,thudi2022unrolling}.

Zeng et al.~\cite{zeng2021learning} suggested a new method of modeling data influence by adding regularization terms into the learning algorithm. Although this method is model-agnostic, it requires intervening in the original training process of the original model. Moreover, it is only applicable to convex learning problems and deep neural networks.
Peste et al.~\cite{peste2021ssse} closed this gap by introducing a new Fisher-based unlearning method, which can approximate the Hessian matrix. This method works for both shallow and deep models, and also convex and non-convex problems. The idea is to efficiently compute the matrix inversion of a Fisher Information Matrix using rank-one updates. However, as the whole process is approximate, there is no concrete guarantee on the unlearned model.

Yamasita et al.~\cite{yamashita2023one} propose a one-shot unlearning technique that removes the need for extra training by adding noise to the model's sensitive parameters, estimated using the Fisher information matrix (FIM). Unlike existing methods, this approach does not require retaining training data for FIM calculation. Instead, it utilizes class-specific synthetic signals, or mnemonic codes, during training. Each class is assigned a mnemonic code, which is incorporated into the training data as:
\begin{equation}
 \tilde{x}_{i}=(1-\lambda)x_{i}+\lambda\xi_{c}   
\end{equation}
where $\lambda\xi_{c}$ represents the mnemonic codes and $\lambda$ is the hyperparameter. This eliminates the need to retain training data for the unlearning process, reducing storage costs and enhancing practicality.

\section{Published Resources on Machine Unlearning}
\label{sec:resources}

\subsection{Published Unlearning Algorithms}

Some implementations of algorithms and models are available that have contributed to baseline experiments in machine unlearning. A summary of published implementations, including their language and platform details, the corresponding models, and URLs to their code repositories, are presented in \autoref{tab:algorithms}.

\begin{table*}[!h]
    \centering
    \vspace{-1em}
    \caption{Published Algorithms and Models} %\todo{TODO: any other column is welcomed}
    \label{tab:algorithms}
    \vspace{-1em}
    \begin{adjustbox}{max width=\textwidth}
    \begin{threeparttable}
    \begin{tabular}{c|c|c|c|l}
        \toprule
	\textbf{Unlearning Algorithms} & \textbf{Language} & \textbf{Platform} & \textbf{Applicable ML Models} &  \textbf{Code Repository} \\
	\midrule
	SISA~\cite{bourtoule2021machine} & Python &  - & Model-agnostic & \url{https://github.com/cleverhans-lab/machine-unlearning} \\
	Athena~\cite{sommer2020towards,sommer2022athena} & Python & - & Model-agnostic & \url{https://github.com/inspire-group/unlearning-verification} \\
	AmnesiacML~\cite{graves2021amnesiac} & Python & - & Model-agnostic & \url{https://github.com/lmgraves/AmnesiacML} \\
	Kpriors ~\cite{khan2021knowledge} & Python & Pytorch & Model-agnostic & \url{https://github.com/team-approx-bayes/kpriors} \\
	ERM ~\cite{neel2021descent} & Python & - & Model-agnostic & \url{https://github.com/ChrisWaites/descent-to-delete} \\
	ShallowAttack~\cite{chen2021machine} & Python & Pytorch & Model-agnostic & \url{https://github.com/MinChen00/UnlearningLeaks} \\
	UnrollingSGD~\cite{thudi2022unrolling} & Python & - & Model-agnostic & \url{https://github.com/cleverhans-lab/unrolling-sgd} \\
	DeltaGrad~\cite{wu2020deltagrad} & Python & - & Model-agnostic & \url{https://github.com/thuwuyinjun/DeltaGrad} \\
	Amnesia~\cite{schelter2020amnesia} & Rust & - & Model-agnostic & \url{https://github.com/schelterlabs/projects-amnesia} \\
	SCAR~\cite{bonato2024retain} & Python & - & Model-agnostic & \url{https://github.com/jbonato1/SCAR} \\
	KGA~\cite{wang2023kga} & Python & Pytorch & Model-agnostic & \url{https://github.com/Lingzhi-WANG/KGAUnlearn} \\
	MUPy~\cite{cao2015towards} & Python & LensKit & kNN & \url{https://github.com/theLauA/MachineUnlearningPy} \\
	DelKMeans~\cite{ginart2019making} & Python & - & kMeans & \url{https://github.com/tginart/deletion-efficient-kmeans} \\
	CertifiedRem~\cite{GuoGHM20}& Python & Pytorch & Linear models & \url{https://github.com/facebookresearch/certified-removal} \\
	CertAttack ~\cite{marchant2022hard} & Python & Tensorflow & Linear models & \url{https://github.com/ngmarchant/attack-unlearning} \\
	PRU ~\cite{izzo2021approximate} & Python & - & Linear models & \url{https://github.com/zleizzo/datadeletion} \\
	DeltaBoost~\cite{wu2023deltaboost} & Python & - & Tree-based models & \url{https://github.com/Xtra-Computing/DeltaBoost} \\
	HedgeCut~\cite{schelter2021hedgecut} & Python & - & Tree-based models & \url{https://github.com/schelterlabs/hedgecut} \\
	DaRE-RF~\cite{brophy2021machine} & Python & - & Tree-based models & \url{https://github.com/jjbrophy47/dare_rf}\\
	MCMC-Unlearning ~\cite{fu2022knowledge} & Python & Pytorch & Bayesian models & \url{https://github.com/fshp971/mcmc-unlearning} \\
	BIF~\cite{fu2021bayesian} & Python & Pytorch & Bayesian models & \url{https://github.com/fshp971/BIF} \\
	L-CODEC~\cite{mehta2022deep} & Python, Matlab & Pytorch & Deep learning & \url{https://github.com/vsingh-group/LCODEC-deep-unlearning} \\
	SelectiveForgetting~\cite{golatkar2020forgetting,golatkar2020eternal} & Python & - & Deep learning & \url{https://github.com/AdityaGolatkar/SelectiveForgetting} \\
	Neurons~\cite{DaiDHSCW22} & Python & - & Deep learning & \url{https://github.com/Hunter-DDM/knowledge-neurons} \\
	Unlearnable~\cite{huang2021unlearnable} & Python & - & Deep learning & \url{https://github.com/HanxunH/Unlearnable-Examples} \\
	DLMA~\cite{yu2021does} & Python & - & Deep learning & \url{https://github.com/AnonymousDLMA/MI_with_DA} \\
	ERM-KTP~\cite{lin2023erm} & Python & Pytorch & Deep learning & \url{https://github.com/RUIYUN-ML/ERM-KTP} \\
	GraphProjector~\cite{cong2023efficiently} & Python & - & Graph Learning & \url{https://github.com/CongWeilin/Projector} \\
	GraphEditor~\cite{conggrapheditor} & Python & - & Graph Learning & \url{https://anonymous.4open.science/r/GraphEditor-NeurIPS22-856E/README.md} \\
	FedLU~\cite{zhu2023heterogeneous} & Python & Pytorch & Graph Learning & \url{https://github.com/nju-websoft/FedLU/} \\
	GUIDE~\cite{wang2023inductive} & Python & - & Graph Learning & \url{https://github.com/Happy2Git/GUIDE} \\
	GNNDelete~\cite{cheng2023gnndelete} & Python & - & Graph Learning & \url{https://github.com/mims-harvard/GNNDelete} \\
	GraphEraser~\cite{chen2022graph} & Python & - & Graph Learning & \url{https://github.com/MinChen00/Graph-Unlearning} \\
	GST-Unlearn~\cite{pan2023unlearning} & Python & - & Graph Learning & \url{https://zenodo.org/records/7613150} \\
	RecEraser~\cite{chen2022recommendation} & Python, C++ & - & Recommender Systems & \url{https://github.com/chenchongthu/Recommendation-Unlearning} \\
	ADV-MULTVAE~\cite{ganhor2022unlearning} & Python & - & Recommender Systems & \url{https://github.com/CPJKU/adv-multvae} \\
	FedEraser~\cite{liu2021federaser} & Python & - & Federated Learning & \url{https://www.dropbox.com/s/1lhx962axovbbom/FedEraser-Code.zip?dl=0} \\
	RapidFed~\cite{liu2022right} & Python & - & Federated Learning & \url{https://github.com/yiliucs/federated-unlearning} \\
		SIFU~\cite{fraboni2024sifu} & Python & PyTorch & Federated Learning & \url{https://github.com/Accenture/Labs-Federated-Learning/tree/SIFU} \\
		Fast-FedUL~\cite{huynh2024fast} & Python & - & Federated Learning & \url{https://github.com/thanhtrunghuynh93/fastFedUL} \\
		FATS~\cite{tao2024communication} & Python & - & Federated Learning & \url{https://github.com/Happy2Git/FATS_supplement} \\
		EUL~\cite{chen2023unlearn} & Python & - & LLM & \url{https://github.com/SALT-NLP/Efficient_Unlearning} \\
		E2URec~\cite{wang2024towards} & Python & Pytorch & LLM, Recommender Systems & \url{https://github.com/justarter/E2URec} \\
		Ext-Sub~\cite{hu2024separate} & Python & - & LLM & \url{https://github.com/HITsz-TMG/Ext-Sub} \\
		SP~\cite{pochinkov2024dissecting} & Python & - & LLM & \url{https://github.com/nickypro/selective-pruning} \\
		Quark~\cite{lu2022quark} & Python & - & LLM & \url{https://github.com/GXimingLu/Quark} \\
		I2I-Unlearn~\cite{li2024machine} & Python & Pytorch & Generative Models & \url{https://github.com/jpmorganchase/i2i-generator-unlearning} \\
		AdvUnlearn~\cite{zhang2024defensive} & Python & - & Generative Models & \url{https://github.com/OPTML-Group/AdvUnlearn} \\
		FAST~\cite{panda2023fast} & Python & - & Generative Models & \url{https://github.com/Subhodip123/weak-unlearning-gan} \\
         \bottomrule
    \end{tabular}
    \begin{tablenotes}
\item -: No Dedicated Platforms.
\end{tablenotes}
    \end{threeparttable}
    \end{adjustbox}
    \vspace{-1em}
\end{table*}

\subsection{Published Datasets}

The most widely used datasets in machine unlearning are shown in \autoref{tab:datasets}. We have classified these into several groups based on their field of application, including image, tabular, text, sequence, and graph. Due to the space limit, the details of these datasets can be found in our technical report~\cite{nguyen2022survey}.

\begin{table*}[!h]
    \centering
    \caption{Published Datasets (URLs are embedded in dataset names).} %\todo{TODO: The most important column is REF}
    \label{tab:datasets}
    \vspace{-1em}
    \begin{adjustbox}{max width=0.8\textwidth}
    \begin{threeparttable}
    \begin{tabular}{c|c|c|c|c|c}
        \toprule
	\textbf{Category} & \textbf{Dataset} &  \textbf{\#Items} &  \textbf{Disk Size} & \textbf{Downstream Applications} & \textbf{REF} \\
	\midrule
	\multirow{3}{*}{Image} & \href{https://deepai.org/dataset/mnist}{MNIST} & 70K & 11MB &  Classification & 29+ papers (\cite{bourtoule2021machine,GuoGHM20,ginart2019making}, \ldots) \\
	& \href{https://www.cs.toronto.edu/~kriz/cifar.html}{CIFAR} & 60K & 163MB & Classification & 16+ papers (\cite{huang2021unlearnable,wang2022federated,shibata2021learning,graves2021amnesiac}, \ldots) \\
	& \href{http://ufldl.stanford.edu/housenumbers/}{SVHN} & 600K & 400MB+ & Classification & 8+ papers(\cite{GuoGHM20,huang2021mathsf,liu2020have,huang2021unlearnable,bourtoule2021machine}, \ldots)  \\ 
	& \href{https://www.yf.io/p/lsun}{LSUN}~\cite{yu2015lsun} & 69M+ & 1TB+ & Classification & \cite{GuoGHM20} \\ 
	& \href{https://www.image-net.org}{ImageNet}~\cite{deng2009imagenet} & 14M+ & 166GB & Classification & \cite{bourtoule2021machine,huang2021unlearnable,he2021deepobliviate,sommer2020towards,sommer2022athena,gao2022verifi}  \\ 
	\hline
	\multirow{3}{*}{Tabular} & \href{https://archive.ics.uci.edu/ml/datasets/adult}{Adult} & 48K+ & 10MB & Classification  & 8+ papers(\cite{liu2021federaser,brophy2021machine,chen2021machine,schelter2021hedgecut,shokri2017membership,liu2021revfrf}, \ldots) \\
	& \href{https://archive.ics.uci.edu/ml/datasets/breast+cancer}{Breast Cancer} & 569 & $<1MB$ & Classification & \cite{gao2022deletion,wu2022puma} \\
	& \href{https://archive.ics.uci.edu/ml/datasets/diabetes}{Diabetes} & 442 & $<1MB$ & Regression & \cite{warnecke2021machine,brophy2021machine,chen2019novel} \\ 
	\hline
	\multirow{3}{*}{Text} & \href{https://ai.stanford.edu/~amaas/data/sentiment/}{IMDB Review} & 50K & 66MB  & Sentiment Analysis  & \cite{liu2020learn} \\
	& \href{https://keras.io/api/datasets/reuters/}{Reuters} & 11K+ & 73MB  & Categorization  & \cite{liu2020learn} \\
	& \href{https://archive.ics.uci.edu/ml/datasets/Twenty+Newsgroups}{Newsgroup} & 20K & 1GB+  & Categorization  & \cite{liu2020learn} \\
	\hline
	\multirow{3}{*}{Sequence} & \href{https://archive.ics.uci.edu/ml/datasets/Epileptic%2BSeizure%2BRecognition}{Epileptic Seizure} & 11K+ & 7MB  & Timeseries Classification  & \cite{chundawat2022can} \\
	& \href{https://archive.ics.uci.edu/ml/datasets/human+activity+recognition+using+smartphones}{Activity Recognition} & 10K+ & 26MB & Timeseries Classification  & \cite{chundawat2022can} \\
	& \href{https://archive.ics.uci.edu/ml/datasets/detection_of_IoT_botnet_attacks_N_BaIoT}{Botnet} & 72M & 3GB+ & Clustering  & \cite{ginart2019making} \\
	\hline
	\multirow{3}{*}{Graph} & \href{https://ogb.stanford.edu/}{OGB} & 100M+ &  59MB & Classification  & \cite{congprivacy,chien2022certified} \\
	& \href{https://relational.fit.cvut.cz/dataset/CORA}{Cora} & 2K+ & 4.5MB  & Classification  & \cite{chen2021graph,chien2022certified,congprivacy} \\
	& \href{http://konect.cc/networks/}{MovieLens} & 1B+ & 3GB+ & Recommender Systems  & \cite{schelter2020amnesia} \\
         \bottomrule
    \end{tabular}
    \begin{tablenotes}
\item REF: Papers that run experiments on the dataset.
\end{tablenotes}
    \end{threeparttable}
    \end{adjustbox}
    \vspace{-1em}
\end{table*}

\subsection{Evaluation Metrics}
\label{sec:metrics}

The most often used metrics for measuring anomaly detection performance include accuracy, completeness, unlearn time, distance, and forgetting scores. Their formulas and common usage are summarized in \autoref{tab:metrics}. More detailed descriptions are given below.

\sstitle{Accuracy}
In machine unlearning, a model's accuracy needs to be compared on three different datasets: (1) The set to be forgotten. Since the expected behaviour of an unlearned model should mirror that of a retrained model, the accuracy on the remaining data should be similar to the retrained model. (2) The retained set. The retained set's accuracy should be close to that of the original model. (3) The test set. The unlearned model should still perform well on a separate test dataset compared to the retrained model.

\sstitle{Completeness}
The influence of the to-be-removed samples on the unlearned model must be completely eliminated. Completeness, hence, measures the degree to which an unlearned model is compatible with a retrained model~\cite{cao2015towards}. If the unlearned model gives similar predictions to a retrained model for all samples, the operation of feeding samples or observing the model's information is impractical to achieve the forgotten data and its lineage. The final metric is often calculated as the overlap of output space (e.g., the Jaccard distance) between the unlearned model and the retraining. However, computing this metric is often exhaustive.

\sstitle{Unlearning time and Retraining time}
Timeliness quantifies the time saved when using unlearning instead of retraining for model update. The quicker the system restores privacy, security, and usefulness, the more timely the unlearning process. In particular, retraining uses the whole training set to execute the learning algorithm, whereas unlearning executes the learning algorithm on a limited amount of summations; hence, the speed of unlearning is quicker due to the reduced size of the training data.

\sstitle{Relearn time}
Relearning time is an excellent proxy for measuring the amount of unlearned data information left in the model. If a model recovers its performance on unlearned data with just a few steps of retraining, it is extremely probable that the model has retained some knowledge of the unlearned data.

\sstitle{The layer-wise distance}
The weight difference between the original and unlearned neural networks helps when evaluating the unlearning impact on each layer~\cite{tarun2021fast}. The weight difference should be comparable to a retrained model given that a shorter distance indicates ineffective unlearning. Likewise, a much longer distance may point to a Streisand effect and possible information leaks.

\sstitle{Activation Distance}
The activation distance is the separation between the final activation of the scrubbed weights and the retrained model. A shorter activation distance indicates superior unlearning.

\sstitle{JS-Divergence}
When paired with the activation distance, the JS-Divergence between the predictions of the unlearned and retrained model provides a more full picture of unlearning. Less divergence results in better unlearning. The formula of JS-Divergence is 
\[\mathcal{JS}(M(x), T_d(x)) = 0.5*\mathcal{KL}(M(x)||m)+ 0.5*\mathcal{KL}(T_d(x)||m)\]
where $M$ is unlearned model, $T_d$ is a competent teacher, and $\mathcal{KL}$ is The Kullback-Leibler  divergence \cite{KLFormula}, $m= \frac{M(x)+T_d(x)}{2}$.

\sstitle{Membership Inference}
The membership inference metric leverages a membership inference attack to determine whether or not any information about the forgotten samples remains in the model~\cite{chen2021machine}. The set to be forgotten should have reduced the attack probability in the unlearned model. The chance of an inference attack should be reduced in the unlearned model compared to the original model for the forgotten class data.

\sstitle{ZRF (Zero Retrain Forgetting) score}
ZRF makes it possible to evaluate unlearning approaches independent of retraining~\cite{chundawat2022can}. The unpredictability of the model's predictions is measured by comparing them to an unskilled instructor. ZRF compares the set to be forgotten's output distribution to the output of a randomly initialised model, which in most situations is our lousy instructor. The ZRF score ranges between 0 and 1; it will be near to 1 if the model's behaviour with the forgotten samples is entirely random, and close to 0 if it exhibits a certain pattern. Formally, $\mathcal{ZRF} = 1 - \frac{1}{nf}\sum\limits_{i=0}^{n_f} \mathcal{JS}(M(x_i), T_d(x_i))$, where $x_i$
is the $i_{th}$ sample from the set to be forgotten with a total number of samples $n_f$.

\sstitle{Anamnesis Index (AIN)}
AIN values range between 0 and 1. The better the unlearning, the closer to 1. Instances where information from the classes to be forgotten are still preserved in the model correlate to AIN levels well below 1. A score closer to 0 also suggests that the unlearned model will rapidly relearn to generate correct predictions. This may be due to the fact that the last layers contain limited reversible modifications, which degrades the performance of the model on the forgotten classes. If an AIN score is much greater than 1, it may suggest that the approach causes parameter changes that are so severe that the unlearning itself may be detected (Streisand effect). This might be due to the fact that the model was pushed away from the original point and, as a result, is unable to retrieve previously learned knowledge about the forgotten class(es). The formula for calculating an AIN value is 
$AIN = \frac{r_t (M_u, M_{orig}, \alpha)}{r_t (M_s, M_{orig}, \alpha)}$, where $\alpha\%$ is a margin around the initial precision used to determine relearn time. $r_t(M, M_{orig}, \alpha )$ are mini-batches (or steps) to be achieved by the model $M$ on the classes to be forgotten within $\alpha\%$ of the precision compared to the original model $M_{orig}$.
$M_u$ and $M_s$ respectively represent the unlearned model and a model trained from scratch.

\begin{table*}[!h]
    \centering
%    \vspace{-2em}
    \caption{Evaluation Metrics}
    \label{tab:metrics}
    \vspace{-1em}
    \begin{adjustbox}{max width=0.85\textwidth}
    \begin{threeparttable}
    \begin{tabular}{p{2cm}|p{7cm}|p{7cm}|c}
        \toprule
	\textbf{Evaluation Metrics} & \textbf{Formula/Description} &  \textbf{Usage} & \textbf{REF}\\
	\midrule
	Accuracy & Accuracy on unlearned model on forget set and retrain set & Evaluating the predictive performance of unlearned model & \cite{tarun2021fast,golatkar2020eternal,golatkar2021mixed,chundawat2022zero} \\
	\hline
	Completeness & The overlapping (e.g. Jaccard distance) of output space
between the retrained and the unlearned model & Evaluating the indistinguishability between model outputs & \cite{cao2015towards}\\
\hline
	Unlearn Time & The amount of time of unlearning request & Evaluating the unlearning efficiency & \cite{cao2015towards} \\
	\hline
	Relearn Time & The epochs number required for the unlearned model to reach the accuracy of source model & Evaluating the unlearning efficiency (relearn with some data sample)& \cite{cao2015towards,tarun2021fast} \\
	\hline
	Layer-wise Distance & The weight difference between original model and retrain model &  Evaluate the indistinguishability between model parameters & \cite{tarun2021fast} \\
	\hline
	Activation Distance & An average of the L2-distance between the unlearned model and retrained model's predicted probabilities on the forget set & Evaluating the indistinguishability between model outputs & \cite{chundawat2022can,golatkar2021mixed} \\
	\hline
	JS-Divergence & Jensen-Shannon divergence between the predictions of the unlearned and retrained model: $\mathcal{JS}(M(x), T_d(x)) = 0.5*\mathcal{KL}(M(x)||m)+ 0.5*\mathcal{KL}(T_d(x)||m)$ & Evaluating the indistinguishability between model outputs & \cite{chundawat2022can} \\
	\hline
	Membership Inference Attack & Recall (\#detected items / \#forget items) & Verify the influence of forget data on the unlearned model & \cite{golatkar2021mixed,graves2021amnesiac} \\
	\hline
	ZRF score & $\mathcal{ZRF} = 1 - \frac{1}{nf}\sum\limits_{i=0}^{n_f} \mathcal{JS}(M(x_i), T_d(x_i))$  & The unlearned model should not intentionally give wrong output ($\mathcal{ZRF} = 0$) or random output ($\mathcal{ZRF} = 1$) on the forget item & \cite{chundawat2022can} \\
 \hline
	Anamnesis Index (AIN) & $AIN = \frac{r_t (M_u, M_{orig}, \alpha)}{r_t (M_s, M_{orig}, \alpha)}$ & Zero-shot machine unlearning & \cite{chundawat2022zero} \\
	\hline
	Epistemic Uncertainty & $
	\mbox{efficacy}(w;D) = \begin{cases}
		\frac{1}{i(w; D)}, \mbox{if i(w;D) > 0} \\
		\infty, \mbox{otherwise}
	\end{cases}
$ & How much information the model exposes & \cite{becker2022epistemic} \\
	\hline
	Model Inversion Attack & Visualization & Qualitative verifications and evaluations & \cite{graves2021amnesiac} \\
	\bottomrule
    \end{tabular}
    \begin{tablenotes}
\item REF: Highlighted papers that used/proposed the metric
\end{tablenotes}
    \end{threeparttable}
    \end{adjustbox}
%    \vspace{-1.5em}
\end{table*}

\sstitle{Epistemic Uncertainty}
Epistemic uncertainty is an uncertainty metric that measures how much we know about the optimal hypothesis in the parameter space~\cite{hullermeier2021aleatoric}. That is, are we certain that the current model parameters are optimal for a given dataset?. The influence functions are computed as the trace of Fischer Information Matrix (FIM)~\cite{becker2022epistemic}:
\begin{equation}
	i(w; D) = tr(\mathcal{I} (w; D))
\end{equation}
where $\mathcal{I} (w; D)$ is the FIM that determines the quantities of information that the model parameters $w$ hold regarding the dataset $D$~\cite{becker2022epistemic}, which can be approximate via Levenberg-Marquart~\cite{martens2020new}:
\begin{equation}
	\mathcal{I}(w; D) \approx \frac{1}{|D|} \sum_{x,y \in D} \left( \frac{\partial \log p(y|x;w}{\partial w} \right)^2
\end{equation}
Becker et al.~\cite{becker2022epistemic} proposed an efficacy score based on the epistemic uncertainty as follows.
\begin{equation}
	\mbox{efficacy}(w;D) = \begin{cases}
		\frac{1}{i(w; D)}, \mbox{if i(w;D) > 0} \\
		\infty, \mbox{otherwise}
	\end{cases}
\end{equation}
It measures how much information the model exposes. This metric is computationally efficient and does not require access to the retrained model~\cite{becker2022epistemic}. A better unlearning algorithm will produce the unlearned model with lower efficacy score. However, it only measures the overall information reduction, not the specific information related to the data to be forgotten. Moreover, it should be used along with an accuracy metric, as less exposing models do not necessarily have good predictive performance~\cite{becker2022epistemic}.

\sstitle{Model Inversion attack}
It has been shown that model inversion attacks may reproduce the records from trained machine learning models. It is possible to replicate samples of target points around a regression value if white-box access to a trained model is available. The data provided from the unlearned model following inversion should not include information about the forget class. This metric is qualitative and often used in image applications~\cite{graves2021amnesiac}.

\section{Unlearning Applications}

\subsection{Unlearning in Recommender Systems}

In the field of machine learning, recommender systems are used to predict what a user might want to buy or watch. They often use collaborative filtering to learn a user's preferences based on their past behavior. However, a recommender system may be required to forget private training points and its complete impact on a model in order to protect user privacy or comply with explicit user removal requests. Utility is another reason for unlearning requests. For example, the accuracy of a recommendation can be degraded due to out-of-distribution data or poisoning attacks~\cite{marchant2022hard,du2019lifelong}. In the latter case, data that is detected as poisoned will need to be removed, while in the former case, old data may need to be removed so that the system keeps up with the new data distribution.

Unlearning techniques for machine learning in general cannot be used directly on recommender systems~\cite{chen2022recommendation,wang2022efficiently}. For example, collaborative filtering recommendation uses the information of similarity across users-item interaction; and thus, arbitrarily partitioning the training sets could break such coupling information~\cite{bourtoule2021machine}. Some researchers have developed unlearning methods for graph data~\cite{chen2021graph} and while recommendation data can be modelled as graphs, their user-item interactions are not uniform~\cite{chen2022recommendation}.

To overcome these challenges, Chen et al.~\cite{chen2022recommendation} proposed a partition-based retraining process, called smart retraining, to unlearn the model from the removed  user behavior data. The idea is to develop a strategy to partition the data with regard to the resemblance between users and items while maintaining the balance between different partitions for retraining. Next, the output of the submodels is combined using an attention-based method, each of which is associated with each disjoint partition.

In other settings, Yuan et al.~\cite{yuan2023federated} propose FRU, a method to erase user contributions from federated learning systems using a rollback mechanism. AltEraser~\cite{liu2022forgetting} is a fast unlearning technique for neural recommender systems that employs second-order optimization to efficiently remove unwanted data without full retraining. Sinha et al.~\cite{sinha2024multi} examine unlearning in multi-modal recommendation systems, addressing the complexities of handling different data types and reducing legal risks while optimizing computational efficiency. Liu et al.~\cite{li2023making} explore attribute-wise unlearning to make users indistinguishable in recommender systems, enhancing privacy by eliminating sensitive user attributes.

\subsection{Unlearning Federated Learning}

Recently, federated learning has become popular in the field of machine learning~\cite{mcmahan2017communication}. One typical federated learning scenario is building a machine learning model from healthcare information. Due to privacy regulations, the medical record data cannot leave the clients' devices. Here, the clients could be hospitals or personal computers and are assumed to have machine learning environments. The server does not transmit the actual data to the global model. Rather, there is a communication protocol between the clients and servers that governs collaborative model training~\cite{wu2022federated}. In the literature, the communication protocol Federated Average (FedAvg)~\cite{mcmahan2017communication} is typically used for model training. It consists of multiple rounds. Each round, the current global model weights are transmitted to the clients. Based on these weights, each client uses stochastic gradient descent to adjust their local model. Then the local model's weights are forwarded to the server. In the final phase of each loop, the server aggregates the received weights by (weighted) averaging to prepare the global model for the next round~\cite{halimi2022federated}.

Given such training protocols, machine unlearning cannot be extended easily to the federated learning setting~\cite{fraboni2024sifu,huynh2024fast,tao2024communication}. This is because the global weights are computed by aggregations rather than raw gradients. These are especially mixed up when many clients participate~\cite{gao2022verifi}. Moreover, these clients might have some overlapping data, making it difficult to quantify the impact of each training item on the model weights~\cite{liu2020learn}. Using classic unlearning methods by gradient manipulation may even lead to severe accuracy degradation or new privacy threats~\cite{liu2021federaser}.

Additionally, current studies on federated unlearning tend to assume that the data to be removed belongs wholly to one client~\cite{liu2021federaser,wang2022federated,wu2022federated,liu2020learn}. With this assumption, the historical contributions of particular clients to the global model's training can be logged and erased easily. However, erasing historical parameter updates might still damage the global model, but there are many strategies for overcoming this issue. For example, Liu et al.~\cite{liu2021federaser} proposed calibration training to separate the individual contributions of clients as much as possible. This mechanism does not work well for deep neural networks, but it does work well with shallow architectures such as a 2-layer CNN or a network with two fully-connected layers. In addition, there is a trade-off between scalability and precision due to the cost of storing historical information on the federated server. Wu et al.~\cite{wu2022federated} put forward a knowledge distillation strategy that uses a prime global model to train the unlearned model on the remaining data. However, as the clients' data is not accessible by the server, some unlabeled (synthetic) data that follows the distribution of the whole dataset needs to be sampled and extra rounds of information exchange are needed between the clients and server. As a result, the whole process is costly and approximate. Also, it might be further offset when the data is non-IID~\cite{liu2022right}. On another spectrum, Liu et al.~\cite{liu2022right} proposed a smart retraining method for federated unlearning without communication protocols. The approach uses the L-BFGS algorithm~\cite{berahas2016multi,bollapragada2018progressive} to efficient solve a Hessian approximation with historical parameter updates for global model retraining. However, this method is only applicable to small models  ($\leq$ 10K parameters). Plus, it involves storing old model snapshots (including historical gradients and parameters), which poses some privacy threats.

Recently, Shaik et al.~\cite{shaik2024framu} propose a framework, namely FRAMU, that uses reinforcement learning to help decentralized agents make optimal data unlearning decisions based on real-time feedback. By leveraging attention mechanisms and balancing exploration and exploitation, the framework ensures continual model adaptation, improving convergence, privacy preservation, and decision-making in dynamic, multi-modal environments~\cite{lee2024learning}. Further studies on federated unlearning can be found at~\cite{liu2023survey,9964015,qiu2023fedcio,shao2024federated}.

\subsection{Unlearning in Graph Embedding}

So far the data in machine unlearning settings is assumed to be independent. However, there are many cases where the data samples are relational, such as is the case with graph data. Graph representation learning is a well-established research direction in machine learning, specifically in deep learning~\cite{hamilton2020graph,chen2020graph,lee2024sfgcn,lotfi2021detection}. However, applying machine unlearning to graph representation learning is arguably more challenging. First, the data is correlated, and it is non-trivial to partition the data, even uniformly. Second, the unlearning requests can happen upon a node or an edge. Third, the graph data itself might be non-uniform due to unbalanced connected components in the graph. Therefore, existing graph partition methods might lead to unbalanced data partitions, making the retraining process non-uniform.

To mitigate these problems, Chen et al.~\cite{chen2021graph} proposed a new graph partitioning strategies especially for machine unlearning. The general idea is based on the notion of assignment preference that represents the benefit of a node assigned to a shard (i.e., a data partition). Such node-shard pairs are further fine-tuned with neighbor counts, which track down the number of neighbors of a node belonging to the same target shard. The authors also proposed an aggregation method to combine different partition strategies. Further, the retraining process is based on message passing in graph neural networks, which facilitates fast retraining.

Unlearning without retraining is also possible for graph embedding. However, several challenges need to be overcome. The interdependency of graph data, especially across different subgraphs, is non-trivial for model training. A removal of node and edge could not only cause an impact on its neighbor but also on multi-hops.
Cong et al.~\cite{congprivacy,conggrapheditor} proposed a one-shot unlearning solution that only requires access to the data to be forgotten. The idea is inspired by the architecture of a linear graph neural network (GNN), in which non-linearities in a typical GNN are replaced by a single weight matrix between consecutive convolutional layers. Despite its linear span over all input node features, such linear-GNNs have shown competent performance, e.g. SGC~\cite{wu2019simplifying} and APPNP~\cite{gasteiger2018combining}. Using this property, Cong et al.~\cite{congprivacy,conggrapheditor} proposed an exact unlearning process at the algorithmic level based on linear operations such as projection and recombination.

Existing graph unlearning methods have two main limitations: they either degrade model performance by altering weights shared across all nodes or fail to effectively delete edges due to reliance on local neighborhoods. To solve this, Cheng et al.~\cite{cheng2023gnndelete} introduce GNNDELETE, which focuses on two components: Deleted Edge Consistency and Neighborhood Influence. Deleted Edge Consistency ensures that the impact of removed elements is eliminated from both model weights and adjacent representations. Neighborhood Influence retains residual knowledge even after elements are removed, allowing GNNDELETE to modify representations to delete nodes and edges while preserving the remaining knowledge.

Recently, Li et al.~\cite{li2024towards} propose a mutual evolution framework for general graph unlearning, which iteratively updates both graph structures and model parameters to improve unlearning effectiveness. Zhu et al.~\cite{zhu2023heterogeneous} introduce a method for heterogeneous federated knowledge graph embedding learning and unlearning, enabling privacy-preserving updates in distributed knowledge graph systems. Wang et al.~\cite{wang2023inductive} present inductive graph unlearning, which removes nodes and edges from graph models while maintaining model integrity and scalability in dynamic environments. Finally, Pan et al.~\cite{pan2023unlearning} investigate unlearning in graph classifiers with limited data, developing techniques that minimize the impact of unlearning on model performance despite constrained resources.

\subsection{Unlearning in Lifelong Learning}

Unlearning is not always a bad thing for the accuracy of a machine learning model. Machine unlearning has been researched as a countermeasure against catastrophic forgetting in deep neural networks~\cite{du2019lifelong,parne2021machine,liu2022continual}. Catastrophic forgetting is a phenomenon where deep neural networks perform badly after learning too many tasks~\cite{kirkpatrick2017overcoming}. One naive solution to this problem is training the model on the historical data again. Clearly, this solution is impractical not only due to computational cost but also because there is no guarantee that the model will converge and nor is there a guarantee that the forgetting will not happen again~\cite{parisi2019continual}. Du et al.~\cite{du2019lifelong} suggested a solution based on unlearning to prevent catastrophic forgetting. The core idea is to unlearn harmful samples (e.g., false negatives/positives) and then update the model so that its performance before the forgetting effect is maintained. 

Unlearning has also been used to handle exploding losses in machine learning. Here, the term loss involves the computation of $-\log Pr(x)$, and, when $Pr(x)$ is approximately zero, the loss may be arbitrarily significant. The problem is more severe in anomaly detection where normal samples can have very small $Pr(x)$ (abnormal samples have very large $Pr(x)$ and their sum of probabilities is one). Du et al.~\cite{du2019lifelong} hence proposed an unlearning method to mitigate this problem with an unlearning loss that regularizes those extreme cases.

Unlearning has been studied for other lifelong settings as well~\cite{parne2021machine}. These setting use incremental models, such as decision tree and naive Bayes, which allows the model to unlearn data samples on-the-fly. Liu et al.~\cite{liu2022continual} considered requests on unlearning specific tasks for lifelong models. In particular, there are three types of requests in lifelong learning: (i) to learn a task permanently, (ii) to learn a task temporarily and forget it later upon a privacy request, and (iii) to forget a task. Different from traditional machine unlearning, unlearning in lifelong learning needs to maintain knowledge transfer between tasks but also preserve all knowledge for the remaining tasks. Moreover, the setting is more challenging as it depends on the order of tasks as the tasks are learnt online during the model lifetime. Additionally, the model cannot keep all previous data (zero-glance unlearning), making the unlearning process more challenging. Liu et al.~\cite{liu2022continual} proposed a solution inspired by SISA, the data partitioning mechanism for smart retraining~\cite{bourtoule2021machine}. It creates an isolated temporary model for each task and merges the isolated models into the main model.

%\info{Third, since the lifelong anomaly detection problem will run in
%fashion that the model will keep being updated over time, it may forget previously observed examples. This issue is typically referred to as catastrophic forgetting in the deep learning literature [12]. A naive solution to this problem is to retrain the model with all previously observed examples. However, this naive approach is not practical, since the data set will be ever-growing over time, and retraining will soon become too costly. To tackle this issue, we develop an incremental learning approach to leverage a maintained important memory set to make the model not forgetting important past examples}
%\cite{du2019lifelong}

%\cite{parne2021machine,liu2022continual}

\subsection{Unlearning in Large Language Models}

Unlearning in large language models (LLMs) has become essential due to increasing privacy concerns and the need to comply with regulations like GDPR~\cite{magdziarczyk2019right} and CCPA~\cite{pardau2018california}. As LLMs are trained on vast datasets that may include sensitive information, there is a risk of these models inadvertently memorising and reproducing private data~\cite{kadhe2023fairsisa,zhu2024decoupling,wang2024unlearning}. Chen et al.~\cite{chen2023unlearn} introduce an efficient framework for unlearning in LLMs that integrates lightweight unlearning layers into transformer models, allowing for the selective removal of specific data without retraining the entire model. They also introduce a fusion mechanism to combine these layers when multiple unlearning requests are made. Kassem~\cite{kassem2023preserving} propose another unlearning approach called ``DeMem'', which uses a reinforcement learning feedback loop with a negative similarity score as a reward. This approach incentivises the language model to paraphrase memorised content, thereby reducing the risk of sensitive information being exposed. Wang et al.~\cite{wang2024towards} introduce E2URec, an unlearning method for LLM-based recommender systems. It addresses forgetting user data while preserving model performance by updating a small set of parameters using low-rank adaptation modules and employing a teacher-student framework.

Some works, use machine unlearning as a way to improve the truthfulness and detoxification of LLMs. Hu et al.~\cite{hu2024separate}  propose the ``Extraction-before-Subtraction'' (Ext-Sub) method with parameter-efficient modules (PEMs) to isolate and remove undesirable features like untruthfulness or toxicity, while preserving the model's core capabilities. The method extracts deficiency capabilities from ``anti-expert'' PEMs and subtracts them from ``expert'' PEMs, enhancing performance without compromising fundamental abilities. Pochinkov et al.~\cite{pochinkov2024dissecting} introduce another method by selectively pruning neurons responsible for specific behaviours, such as coding or toxic language, while maintaining overall performance. 
%The approach efficiently identifies and removes these neurons, effectively reducing the targeted behaviour without significantly impacting the model's general capabilities, offering a practical solution for managing unwanted behaviours in LLMs.
Lu et al.~\cite{lu2022quark} presents a framework called Quark that uses reinforcement learning as a machine unlearning approach to mitigate undesirable text generation behaviors, such as toxicity and repetition, while preserving language quality. 
%By quantizing text based on reward scores and applying a KL-divergence penalty, Quark effectively improves control over text generation, outperforming other RL methods like PPO in both fluency and diversity.

\subsection{Unlearning in Generative Models}

Machine unlearning has been studied for generative models, such as diffusion models and adversarial networks. Zhang et al.~\cite{zhang2024defensive} propose a method that ensures robust concept erasure in diffusion models. This approach uses adversarial training to defend against unwanted model behavior while selectively erasing learned concepts, maintaining the model's overall functionality. Li et al.~\cite{li2024machine} develop techniques to remove specific data representations from image generators. Unlearncanvas~\cite{zhang2024unlearncanvas}, a stylised image dataset to benchmark unlearning performance in diffusion models, has been developed. This dataset provides a standard for evaluating the effectiveness of various unlearning techniques in visual data, emphasizing the need for robust testing frameworks. Panda et al.~\cite{panda2023fast} present FAST, a weak unlearning method for black-box generative models that uses feature similarity to remove specific learned patterns without significantly impacting model accuracy. Tiwary et al.~\cite{tiwary2023adapt} explore a strategy that exploits parameter space semantics in GANs to improve unlearning efficiency, enabling models to adapt to new data while effectively forgetting outdated or undesired information.

\section{Discussion and Future Prospects}
\label{sec:discussion}

In this section, we analyze the current and potential developments in machine unlearning and summarize our findings. In addition, we identify a number of unanswered research topics that could be addressed to progress the foundation of machine unlearning.

\subsection{Summary and Trends}

\sstitle{Influence functions are dominant methods}
Understanding the impact of a given data item on a model's parameters or model performance is the key to machine unlearning~\cite{koh2017understanding,basu2021influence}. Such insights will speed-up the unlearning process immensely by simply reversing the model updates associated with the target data. Although there could be some offset in doing so, promising results have shown that this offset can be bounded~\cite{mahadevan2021certifiable,
mahadevan2022certifiable,chundawat2022zero}.

\sstitle{Reachability of model parameters}
Existing works define unlearning as obtaining a new model with an accuracy as good as if the model had retrained without the data to be forgotten ~\cite{golatkar2020forgetting,golatkar2020eternal,graves2021amnesiac,thudi2021necessity}. We argue that such a parameter-space assumption should be taken into serious considerations. As model parameters can be reachable with or without some given data, is there any case where the original and unlearned models share the same parameters?~\cite{thudi2021necessity} Although some studies use parameter distribution to bound the problem~\cite{GuoGHM20}, there could still be false positive cases, where some effects from the forgotten data still exist in the unlearned model.

\sstitle{Unlearning verification (Data auditing) is needed}
Unlearning verification (or data auditing) is the process of determining whether specific data have been eliminated from a model.
To fully enable regulations over the right to be forgotten, the unlearning effect should be independently verified. There have been only a few works on unlearning verification~\cite{gao2022verifi,huang2021mathsf}. However, the definition of a successful verification is still controversial as different unlearning solutions use different evaluation metrics, especially when the cutting threshold of a verification metric still depends on the application domain~\cite{thudi2021necessity}.

\sstitle{Federated unlearning is emerging}
Federated learning brings about a unique setting for machine unlearning research~~\cite{liu2021federaser,wang2022federated,wu2022federated,liu2020learn}. It has separate clients participating in the federated training process. As a result, removing a client out of the federation could be done precisely using historical updates. The rational behind this is that the user data on a client mostly helps to make correct predictions about that user. This locality helps to avoid a catastrophic unlearning phenomenon in a traditional machine learning setting. However, we all need to be aware that there are many cases in federated learning where the data is non-IID or the removal request only covers part of the client data.

\sstitle{Model repair via unlearning}
Machine learning models can be poisoned by adversarial attacks~\cite{wang2019neural,liu2022backdoor}. Intuitively, if the poisonous data is detected and removed and then the model is retrained, the new model should be poison-free. However, the retraining would be too expensive. This is indeed similar to the unlearning setting. Compared to existing defence methods, the models in machine learning determine then update the inner problematic weights through influence functions.

A similar application is to remove bias from the model due to some biased feature in the data~\cite{dinsdale2021deep,dinsdale2020unlearning}. Status quo studies on fairness and de-biasing learning mostly focus on learning a fair and unbiased feature representation~\cite{ramaswamy2021fair,nam2020learning,singh2022anatomizing,
wang2020towards}, where, machine unlearning, e.g. feature unlearning~\cite{guo2022efficient}, would ensure the biased features are deleted properly but the model's quality would still be maintained.

In another setting, machine unlearning can be used to repair overtrained deep neural networks by actively unlearning useless, obsolete, or redundant data samples that could cause catastrophic forgetting~\cite{du2019lifelong,golatkar2020eternal}. Moreover, machine unlearning might be used to boost the model's accuracy as well~\footnote{\url{https://insights.daffodilsw.com/blog/machine-unlearning-what-it-is-all-about}}, e.g. as forgetting is similar to compressing in information bottleneck theory~\cite{shwartz2017opening,tishby2000information,tishby2015deep}~\footnote{\url{https://github.com/ZIYU-DEEP/Awesome-Information-Bottleneck}}.

\subsection{Open Research Questions}

There are several open research questions that future studies can address. This section will list and discuss those fundamental topics in machine unlearning.

\sstitle{Unified Design Requirements}
Among the current unlearning approaches, there is no absolute winner that satisfies all design requirements.
Most unlearning algorithms focus on approximate unlearning scenarios and data item removal (\autoref{tab:unlearning_comparison}). However, there are other types of practical unlearning scenarios that need to be considered, such as zero-glance, zero-shot, few-shot learning. Likewise, there are other types of  removal requests that must be handled, e.g., feature removal, class removal, task removal, stream removal, and so on. Moreover, satisfying all design requirements -- completeness, timeliness, accuracy, etc. -- would make unlearning solutions more applicable to industry-grade systems.

\sstitle{Unified Benchmarking}
Although there have been many works on machine unlearning recently, not many of them have a common setting for benchmarking comparisons. In particular, there is not a lot of published source code (\autoref{tab:algorithms}) and each of them targets different learning algorithms or different applications (e.g. recommender systems, graph embedding). Schelter et al.~\cite{schelter2020amnesia} undertook an empirical study but the benchmark was limited to decremental learning methods and focused only on efficiency.

\sstitle{Adversarial Machine Unlearning}
More studies have focused on attacking ML systems to improve our understanding and protection of these systems~\cite{veale2018algorithms,wang2019neural,ren2020adversarial}. Adversarial machine unlearning examines attacks on unlearning algorithms to better certify unlearned models~\cite{gao2022deletion,marchant2022hard}. Unlike machine unlearning, which mitigates adversarial attacks~\cite{liu2022backdoor}, adversarial machine unlearning is stricter, addressing not only model accuracy but also privacy guarantees. For instance, it may lack knowledge of learning algorithms but still access the unlearning process.

\sstitle{Interpretable Machine Unlearning}
In the future, explanations for machine unlearning can be used to increase confidence in human-AI interactions and enable unlearning verification or removed data auditing~\cite{gao2022verifi,huang2021mathsf}. However, the inverted nature of machine unlearning might pose problems for explanation methods to be applicable at all. Devising techniques aimed at explaining the unlearning process (e.g. using influence functions) is still an unsolved task~\cite{koh2017understanding,basu2021influence}.

\sstitle{Machine Unlearning in Evolving Data Streams}
Evolving data streams pose problems to machine learning models, especially neural networks, due to shifts in the data distributions and the model predictions~\cite{haug2021learning}. Although there are ways to overcome this limitation~\cite{duda2020training}, they rely on the changes in the model parameters to detect concept drift~\cite{haug2021learning}. However, such detection might be not reliably correct in unlearning settings, where the changes in model parameters are approximate. Consequently, it is expected that machine learning for streaming removal request may attract more attention in the next few years. It is noteworthy that unlearning might be used to repair obsolete models by forgetting old data that contradicts with the detected concept drift. However, this requires a contradiction analysis between old and new data~\cite{hu2021distilling}.

A similar setting is the consideration of out-of-distribution (OOD) data in the forget set. For those settings, the unlearning is imbalanced. Some data might have no impact while the others have great influence on the model parameters. There are studies on learning algorithms for OOD data in federated learning~\cite{sattler2021fedaux}. Hence, it may be worthwhile investigating novel unlearning algorithms tailored to OOD data.

%\sstitle{Attacking Unlearned Models: strong attacks give strong certificates}

%\sstitle{Unlearning as model repair}

%\sstitle{Unlearn with Concept Drifts}

%\sstitle{Unlearn with OOD Data}

%\sstitle{Unlearning in Federated Learning}

%\sstitle{Unlearning with Multiple Optimization Objectives}

\sstitle{Causality in Machine Unlearning}
There are cases where a large amount of data need to be removed from an machine learning system, even though the portion of data to be forgotten is insignificant in comparison to all the data. For example, a data pollution attack might affect millions of data items, but only a few of them can be detected by human experts or SOTA detection methods~\cite{cao2018efficient}. Causality analysis~\cite{guo2020survey} could become a useful tool to automatically unlearn the polluted data in this setting and guarantee the non-existence of the polluted information in the final model.

\section{Conclusion}
\label{sec:conclusion}

This survey is the first to investigate machine unlearning techniques in a systematic manner. In this paper, we addressed the primary difficulties and research advancements in conceptualizing, planning, and solving the problems of machine unlearning. In addition, we presented a unified taxonomy that divides machine unlearning strategies into three approaches: model-agnostic methods, model-intrinsic methods, and data-driven methods. We hope that our taxonomy can help categorize future studies and gain deeper insight into methodologies as well as address the difficulties in machine unlearning. Also, we expect this survey can assist researchers in identifying the most optimal unlearning strategies for different applications. The survey provides clear summaries and comparisons between various unlearning methodologies, giving a comprehensive and general view of current work as well as the current process of machine unlearning.

%Furthermore, we also show how to design a machine unlearning framework with removal requests, design requirements, and verification procedures that need to be considered. To advance future studies in this field, we provide a foundation for structured benchmarking by assembling various popular datasets and open-source applications. We also highlight the current trends and future prospects based on our survey results. 
%
%Due to the importance of privacy, security, usability, and fidelity to machine learning systems in both practical applications and academia, innovative concepts in machine unlearning are in high demand and have the potential to influence the field substantially. With this survey, we expect many applications from various domains to profit from machine unlearning in the coming years.


% \begin{acks}
% We would like to thank 	
% \end{acks}

%\vspace{-1em}


%\bibliographystyle{ACM-Reference-Format}
%\bibliography{../ref_short,../ref2,../ref_b}

\begin{thebibliography}{220}

%%% the Web address.

\ifx \showCODEN    \undefined \def \showCODEN     #1{\unskip}     \fi
\ifx \showDOI      \undefined \def \showDOI       #1{#1}\fi
\ifx \showISBNx    \undefined \def \showISBNx     #1{\unskip}     \fi
\ifx \showISBNxiii \undefined \def \showISBNxiii  #1{\unskip}     \fi
\ifx \showISSN     \undefined \def \showISSN      #1{\unskip}     \fi
\ifx \showLCCN     \undefined \def \showLCCN      #1{\unskip}     \fi
\ifx \shownote     \undefined \def \shownote      #1{#1}          \fi
\ifx \showarticletitle \undefined \def \showarticletitle #1{#1}   \fi
\ifx \showURL      \undefined \def \showURL       {\relax}        \fi
% invisible to TeX
\providecommand\bibfield[2]{#2}
\providecommand\bibinfo[2]{#2}
\providecommand\natexlab[1]{#1}
\providecommand\showeprint[2][]{arXiv:#2}

\bibitem[\protect\citeauthoryear{Abadi, Chu, Goodfellow, McMahan, Mironov,
  Talwar, and Zhang}{Abadi et~al\mbox{.}}{2016}]%
        {abadi2016deep}
\bibfield{author}{\bibinfo{person}{Martin Abadi}, \bibinfo{person}{Andy Chu},
  \bibinfo{person}{Ian Goodfellow}, \bibinfo{person}{H~Brendan McMahan},
  \bibinfo{person}{Ilya Mironov}, \bibinfo{person}{Kunal Talwar}, {and}
  \bibinfo{person}{Li Zhang}.} \bibinfo{year}{2016}\natexlab{}.
\newblock \showarticletitle{Deep learning with differential privacy}. In
  \bibinfo{booktitle}{\emph{SIGSAC}}. \bibinfo{pages}{308--318}.
\newblock


\bibitem[\protect\citeauthoryear{Aldaghri, Mahdavifar, et~al\mbox{.}}{Aldaghri
  et~al\mbox{.}}{2021}]%
        {aldaghri2021coded}
\bibfield{author}{\bibinfo{person}{Nasser Aldaghri}, \bibinfo{person}{Hessam
  Mahdavifar}, {et~al\mbox{.}}} \bibinfo{year}{2021}\natexlab{}.
\newblock \showarticletitle{Coded machine unlearning}.
\newblock \bibinfo{journal}{\emph{IEEE Access}}  \bibinfo{volume}{9}
  (\bibinfo{year}{2021}), \bibinfo{pages}{88137--88150}.
\newblock


\bibitem[\protect\citeauthoryear{Basu, Pope, and Feizi}{Basu
  et~al\mbox{.}}{2021}]%
        {basu2021influence}
\bibfield{author}{\bibinfo{person}{Samyadeep Basu}, \bibinfo{person}{Phil
  Pope}, {and} \bibinfo{person}{Soheil Feizi}.}
  \bibinfo{year}{2021}\natexlab{}.
\newblock \showarticletitle{Influence Functions in Deep Learning Are Fragile}.
  In \bibinfo{booktitle}{\emph{ICLR}}.
\newblock


\bibitem[\protect\citeauthoryear{Baumhauer, Sch{\"o}ttle, and
  Zeppelzauer}{Baumhauer et~al\mbox{.}}{2020}]%
        {baumhauer2020machine}
\bibfield{author}{\bibinfo{person}{Thomas Baumhauer}, \bibinfo{person}{Pascal
  Sch{\"o}ttle}, {and} \bibinfo{person}{Matthias Zeppelzauer}.}
  \bibinfo{year}{2020}\natexlab{}.
\newblock \showarticletitle{Machine unlearning: Linear filtration for
  logit-based classifiers}.
\newblock \bibinfo{journal}{\emph{arXiv preprint arXiv:2002.02730}}
  (\bibinfo{year}{2020}).
\newblock


\bibitem[\protect\citeauthoryear{Becker and Liebig}{Becker and Liebig}{2022}]%
        {becker2022epistemic}
\bibfield{author}{\bibinfo{person}{Alexander Becker} {and}
  \bibinfo{person}{Thomas Liebig}.} \bibinfo{year}{2022}\natexlab{}.
\newblock \showarticletitle{Evaluating Machine Unlearning via Epistemic
  Uncertainty}.
\newblock  (\bibinfo{year}{2022}).
\newblock


\bibitem[\protect\citeauthoryear{Berahas, Nocedal, et~al\mbox{.}}{Berahas
  et~al\mbox{.}}{2016}]%
        {berahas2016multi}
\bibfield{author}{\bibinfo{person}{Albert~S Berahas}, \bibinfo{person}{Jorge
  Nocedal}, {et~al\mbox{.}}} \bibinfo{year}{2016}\natexlab{}.
\newblock \showarticletitle{A multi-batch L-BFGS method for machine learning}.
\newblock \bibinfo{journal}{\emph{NIPS}}  \bibinfo{volume}{29}
  (\bibinfo{year}{2016}).
\newblock


\bibitem[\protect\citeauthoryear{Bitansky, Canetti, Chiesa, and
  Tromer}{Bitansky et~al\mbox{.}}{2012}]%
        {bitansky2012extractable}
\bibfield{author}{\bibinfo{person}{Nir Bitansky}, \bibinfo{person}{Ran
  Canetti}, \bibinfo{person}{Alessandro Chiesa}, {and} \bibinfo{person}{Eran
  Tromer}.} \bibinfo{year}{2012}\natexlab{}.
\newblock \showarticletitle{From extractable collision resistance to succinct
  non-interactive arguments of knowledge, and back again}. In
  \bibinfo{booktitle}{\emph{ITCS}}. \bibinfo{pages}{326--349}.
\newblock


\bibitem[\protect\citeauthoryear{Bollapragada, Nocedal, Mudigere, Shi, and
  Tang}{Bollapragada et~al\mbox{.}}{2018}]%
        {bollapragada2018progressive}
\bibfield{author}{\bibinfo{person}{Raghu Bollapragada}, \bibinfo{person}{Jorge
  Nocedal}, \bibinfo{person}{Dheevatsa Mudigere}, \bibinfo{person}{Hao-Jun
  Shi}, {and} \bibinfo{person}{Ping Tak~Peter Tang}.}
  \bibinfo{year}{2018}\natexlab{}.
\newblock \showarticletitle{A progressive batching L-BFGS method for machine
  learning}. In \bibinfo{booktitle}{\emph{ICML}}. \bibinfo{pages}{620--629}.
\newblock


\bibitem[\protect\citeauthoryear{Bonato, Cotogni, and Sabetta}{Bonato
  et~al\mbox{.}}{2024}]%
        {bonato2024retain}
\bibfield{author}{\bibinfo{person}{Jacopo Bonato}, \bibinfo{person}{Marco
  Cotogni}, {and} \bibinfo{person}{Luigi Sabetta}.}
  \bibinfo{year}{2024}\natexlab{}.
\newblock \showarticletitle{Is Retain Set All You Need in Machine Unlearning?
  Restoring Performance of Unlearned Models with Out-Of-Distribution Images}.
\newblock \bibinfo{journal}{\emph{arXiv preprint arXiv:2404.12922}}
  (\bibinfo{year}{2024}).
\newblock


\bibitem[\protect\citeauthoryear{Bourtoule, Chandrasekaran, Choquette-Choo,
  Jia, Travers, Zhang, Lie, and Papernot}{Bourtoule et~al\mbox{.}}{2021}]%
        {bourtoule2021machine}
\bibfield{author}{\bibinfo{person}{Lucas Bourtoule}, \bibinfo{person}{Varun
  Chandrasekaran}, \bibinfo{person}{Christopher~A Choquette-Choo},
  \bibinfo{person}{Hengrui Jia}, \bibinfo{person}{Adelin Travers},
  \bibinfo{person}{Baiwu Zhang}, \bibinfo{person}{David Lie}, {and}
  \bibinfo{person}{Nicolas Papernot}.} \bibinfo{year}{2021}\natexlab{}.
\newblock \showarticletitle{Machine unlearning}. In
  \bibinfo{booktitle}{\emph{SP}}. \bibinfo{pages}{141--159}.
\newblock


\bibitem[\protect\citeauthoryear{Brophy and Lowd}{Brophy and Lowd}{2021}]%
        {brophy2021machine}
\bibfield{author}{\bibinfo{person}{Jonathan Brophy} {and}
  \bibinfo{person}{Daniel Lowd}.} \bibinfo{year}{2021}\natexlab{}.
\newblock \showarticletitle{Machine unlearning for random forests}. In
  \bibinfo{booktitle}{\emph{ICML}}. \bibinfo{pages}{1092--1104}.
\newblock


\bibitem[\protect\citeauthoryear{Cao and Yang}{Cao and Yang}{2015}]%
        {cao2015towards}
\bibfield{author}{\bibinfo{person}{Yinzhi Cao} {and} \bibinfo{person}{Junfeng
  Yang}.} \bibinfo{year}{2015}\natexlab{}.
\newblock \showarticletitle{Towards making systems forget with machine
  unlearning}. In \bibinfo{booktitle}{\emph{2015 IEEE Symposium on Security and
  Privacy}}. \bibinfo{pages}{463--480}.
\newblock


\bibitem[\protect\citeauthoryear{Cao, Yu, Aday, Stahl, Merwine, and Yang}{Cao
  et~al\mbox{.}}{2018}]%
        {cao2018efficient}
\bibfield{author}{\bibinfo{person}{Yinzhi Cao},
  \bibinfo{person}{Alexander~Fangxiao Yu}, \bibinfo{person}{Andrew Aday},
  \bibinfo{person}{Eric Stahl}, \bibinfo{person}{Jon Merwine}, {and}
  \bibinfo{person}{Junfeng Yang}.} \bibinfo{year}{2018}\natexlab{}.
\newblock \showarticletitle{Efficient repair of polluted machine learning
  systems via causal unlearning}. In \bibinfo{booktitle}{\emph{ASIACCS}}.
  \bibinfo{pages}{735--747}.
\newblock


\bibitem[\protect\citeauthoryear{Cao, Wang, Si, Huang, and Xiao}{Cao
  et~al\mbox{.}}{2022}]%
        {cao2022machine}
\bibfield{author}{\bibinfo{person}{Zihao Cao}, \bibinfo{person}{Jianzong Wang},
  \bibinfo{person}{Shijing Si}, \bibinfo{person}{Zhangcheng Huang}, {and}
  \bibinfo{person}{Jing Xiao}.} \bibinfo{year}{2022}\natexlab{}.
\newblock \showarticletitle{Machine Unlearning Method Based On Projection
  Residual}. In \bibinfo{booktitle}{\emph{DSAA}}. \bibinfo{pages}{1--8}.
\newblock


\bibitem[\protect\citeauthoryear{Cauwenberghs et~al\mbox{.}}{Cauwenberghs
  et~al\mbox{.}}{2000}]%
        {cauwenberghs2000incremental}
\bibfield{author}{\bibinfo{person}{Gert Cauwenberghs} {et~al\mbox{.}}}
  \bibinfo{year}{2000}\natexlab{}.
\newblock \showarticletitle{Incremental and decremental support vector machine
  learning}.
\newblock \bibinfo{journal}{\emph{NIPS}}  \bibinfo{volume}{13}
  (\bibinfo{year}{2000}).
\newblock


\bibitem[\protect\citeauthoryear{Chang, Ren, Nguyen, Nejdl, and Schuller}{Chang
  et~al\mbox{.}}{2022}]%
        {chang2022example}
\bibfield{author}{\bibinfo{person}{Yi Chang}, \bibinfo{person}{Zhao Ren},
  \bibinfo{person}{Thanh~Tam Nguyen}, \bibinfo{person}{Wolfgang Nejdl}, {and}
  \bibinfo{person}{Bj{\"o}rn~W Schuller}.} \bibinfo{year}{2022}\natexlab{}.
\newblock \showarticletitle{Example-based Explanations with Adversarial Attacks
  for Respiratory Sound Analysis}. In \bibinfo{booktitle}{\emph{INTERSPEECH}}.
\newblock


\bibitem[\protect\citeauthoryear{Chaudhuri, Monteleoni, and Sarwate}{Chaudhuri
  et~al\mbox{.}}{2011}]%
        {chaudhuri2011differentially}
\bibfield{author}{\bibinfo{person}{Kamalika Chaudhuri}, \bibinfo{person}{Claire
  Monteleoni}, {and} \bibinfo{person}{Anand~D Sarwate}.}
  \bibinfo{year}{2011}\natexlab{}.
\newblock \showarticletitle{Differentially private empirical risk
  minimization.}
\newblock \bibinfo{journal}{\emph{JMLR}} \bibinfo{volume}{12},
  \bibinfo{number}{3} (\bibinfo{year}{2011}).
\newblock


\bibitem[\protect\citeauthoryear{Chen, Sun, Zhang, and Ding}{Chen
  et~al\mbox{.}}{2022a}]%
        {chen2022recommendation}
\bibfield{author}{\bibinfo{person}{Chong Chen}, \bibinfo{person}{Fei Sun},
  \bibinfo{person}{Min Zhang}, {and} \bibinfo{person}{Bolin Ding}.}
  \bibinfo{year}{2022}\natexlab{a}.
\newblock \showarticletitle{Recommendation unlearning}. In
  \bibinfo{booktitle}{\emph{WWW}}. \bibinfo{pages}{2768--2777}.
\newblock


\bibitem[\protect\citeauthoryear{Chen, Wang, Wang, et~al\mbox{.}}{Chen
  et~al\mbox{.}}{2020}]%
        {chen2020graph}
\bibfield{author}{\bibinfo{person}{Fenxiao Chen}, \bibinfo{person}{Yun-Cheng
  Wang}, \bibinfo{person}{Bin Wang}, {et~al\mbox{.}}}
  \bibinfo{year}{2020}\natexlab{}.
\newblock \showarticletitle{Graph representation learning: a survey}.
\newblock \bibinfo{journal}{\emph{ATSIP}}  \bibinfo{volume}{9}
  (\bibinfo{year}{2020}).
\newblock


\bibitem[\protect\citeauthoryear{Chen and Yang}{Chen and Yang}{2023}]%
        {chen2023unlearn}
\bibfield{author}{\bibinfo{person}{Jiaao Chen} {and} \bibinfo{person}{Diyi
  Yang}.} \bibinfo{year}{2023}\natexlab{}.
\newblock \showarticletitle{Unlearn What You Want to Forget: Efficient
  Unlearning for {LLM}s}. In \bibinfo{booktitle}{\emph{EMNLP}},
  \bibfield{editor}{\bibinfo{person}{Houda Bouamor}, \bibinfo{person}{Juan
  Pino}, {and} \bibinfo{person}{Kalika Bali}} (Eds.).
  \bibinfo{pages}{12041--12052}.
\newblock


\bibitem[\protect\citeauthoryear{Chen, Huang, et~al\mbox{.}}{Chen
  et~al\mbox{.}}{2021a}]%
        {chen2021machinegan}
\bibfield{author}{\bibinfo{person}{Kongyang Chen}, \bibinfo{person}{Yao Huang},
  {et~al\mbox{.}}} \bibinfo{year}{2021}\natexlab{a}.
\newblock \showarticletitle{Machine unlearning via GAN}.
\newblock \bibinfo{journal}{\emph{arXiv preprint arXiv:2111.11869}}
  (\bibinfo{year}{2021}).
\newblock


\bibitem[\protect\citeauthoryear{Chen, Zhang, Wang, Backes, Humbert, and
  Zhang}{Chen et~al\mbox{.}}{2021b}]%
        {chen2021graph}
\bibfield{author}{\bibinfo{person}{Min Chen}, \bibinfo{person}{Zhikun Zhang},
  \bibinfo{person}{Tianhao Wang}, \bibinfo{person}{Michael Backes},
  \bibinfo{person}{Mathias Humbert}, {and} \bibinfo{person}{Yang Zhang}.}
  \bibinfo{year}{2021}\natexlab{b}.
\newblock \showarticletitle{Graph unlearning}.
\newblock \bibinfo{journal}{\emph{arXiv preprint arXiv:2103.14991}}
  (\bibinfo{year}{2021}).
\newblock


\bibitem[\protect\citeauthoryear{Chen, Zhang, Wang, Backes, Humbert, and
  Zhang}{Chen et~al\mbox{.}}{2021c}]%
        {chen2021machine}
\bibfield{author}{\bibinfo{person}{Min Chen}, \bibinfo{person}{Zhikun Zhang},
  \bibinfo{person}{Tianhao Wang}, \bibinfo{person}{Michael Backes},
  \bibinfo{person}{Mathias Humbert}, {and} \bibinfo{person}{Yang Zhang}.}
  \bibinfo{year}{2021}\natexlab{c}.
\newblock \showarticletitle{When machine unlearning jeopardizes privacy}. In
  \bibinfo{booktitle}{\emph{SIGSAC}}. \bibinfo{pages}{896--911}.
\newblock


\bibitem[\protect\citeauthoryear{Chen, Zhang, Wang, Backes, Humbert, and
  Zhang}{Chen et~al\mbox{.}}{2022b}]%
        {chen2022graph}
\bibfield{author}{\bibinfo{person}{Min Chen}, \bibinfo{person}{Zhikun Zhang},
  \bibinfo{person}{Tianhao Wang}, \bibinfo{person}{Michael Backes},
  \bibinfo{person}{Mathias Humbert}, {and} \bibinfo{person}{Yang Zhang}.}
  \bibinfo{year}{2022}\natexlab{b}.
\newblock \showarticletitle{Graph unlearning}. In
  \bibinfo{booktitle}{\emph{CCS}}. \bibinfo{pages}{499--513}.
\newblock


\bibitem[\protect\citeauthoryear{Chen, Xiong, Xu, and Zuo}{Chen
  et~al\mbox{.}}{2019}]%
        {chen2019novel}
\bibfield{author}{\bibinfo{person}{Yuantao Chen}, \bibinfo{person}{Jie Xiong},
  \bibinfo{person}{Weihong Xu}, {and} \bibinfo{person}{Jingwen Zuo}.}
  \bibinfo{year}{2019}\natexlab{}.
\newblock \showarticletitle{A novel online incremental and decremental learning
  algorithm based on variable support vector machine}.
\newblock \bibinfo{journal}{\emph{Cluster Computing}} \bibinfo{volume}{22},
  \bibinfo{number}{3} (\bibinfo{year}{2019}), \bibinfo{pages}{7435--7445}.
\newblock


\bibitem[\protect\citeauthoryear{Cheng, Dasoulas, He, Agarwal, and
  Zitnik}{Cheng et~al\mbox{.}}{2023}]%
        {cheng2023gnndelete}
\bibfield{author}{\bibinfo{person}{Jiali Cheng}, \bibinfo{person}{George
  Dasoulas}, \bibinfo{person}{Huan He}, \bibinfo{person}{Chirag Agarwal}, {and}
  \bibinfo{person}{Marinka Zitnik}.} \bibinfo{year}{2023}\natexlab{}.
\newblock \showarticletitle{{GNND}elete: A General Strategy for Unlearning in
  Graph Neural Networks}. In \bibinfo{booktitle}{\emph{ICLR}}.
\newblock


\bibitem[\protect\citeauthoryear{Chien, Pan, et~al\mbox{.}}{Chien
  et~al\mbox{.}}{2022}]%
        {chien2022certified}
\bibfield{author}{\bibinfo{person}{Eli Chien}, \bibinfo{person}{Chao Pan},
  {et~al\mbox{.}}} \bibinfo{year}{2022}\natexlab{}.
\newblock \showarticletitle{Certified Graph Unlearning}.
\newblock \bibinfo{journal}{\emph{arXiv preprint arXiv:2206.09140}}
  (\bibinfo{year}{2022}).
\newblock


\bibitem[\protect\citeauthoryear{Chundawat, Tarun, Mandal, and
  Kankanhalli}{Chundawat et~al\mbox{.}}{2022a}]%
        {chundawat2022can}
\bibfield{author}{\bibinfo{person}{Vikram~S Chundawat},
  \bibinfo{person}{Ayush~K Tarun}, \bibinfo{person}{Murari Mandal}, {and}
  \bibinfo{person}{Mohan Kankanhalli}.} \bibinfo{year}{2022}\natexlab{a}.
\newblock \showarticletitle{Can Bad Teaching Induce Forgetting? Unlearning in
  Deep Networks using an Incompetent Teacher}.
\newblock \bibinfo{journal}{\emph{arXiv preprint arXiv:2205.08096}}
  (\bibinfo{year}{2022}).
\newblock


\bibitem[\protect\citeauthoryear{Chundawat, Tarun, Mandal, and
  Kankanhalli}{Chundawat et~al\mbox{.}}{2022b}]%
        {chundawat2022zero}
\bibfield{author}{\bibinfo{person}{Vikram~S Chundawat},
  \bibinfo{person}{Ayush~K Tarun}, \bibinfo{person}{Murari Mandal}, {and}
  \bibinfo{person}{Mohan Kankanhalli}.} \bibinfo{year}{2022}\natexlab{b}.
\newblock \showarticletitle{Zero-shot machine unlearning}.
\newblock \bibinfo{journal}{\emph{arXiv preprint arXiv:2201.05629}}
  (\bibinfo{year}{2022}).
\newblock


\bibitem[\protect\citeauthoryear{Cong and Mahdavi}{Cong and Mahdavi}{2022a}]%
        {conggrapheditor}
\bibfield{author}{\bibinfo{person}{Weilin Cong} {and} \bibinfo{person}{Mehrdad
  Mahdavi}.} \bibinfo{year}{2022}\natexlab{a}.
\newblock \showarticletitle{GRAPHEDITOR: An Efficient Graph Representation
  Learning and Unlearning Approach}.
\newblock
  \bibinfo{journal}{\emph{\url{https://congweilin.github.io/CongWeilin.io/}}}
  (\bibinfo{year}{2022}).
\newblock


\bibitem[\protect\citeauthoryear{Cong and Mahdavi}{Cong and Mahdavi}{2022b}]%
        {congprivacy}
\bibfield{author}{\bibinfo{person}{Weilin Cong} {and} \bibinfo{person}{Mehrdad
  Mahdavi}.} \bibinfo{year}{2022}\natexlab{b}.
\newblock \showarticletitle{Privacy Matters! Efficient Graph Representation
  Unlearning with Data Removal Guarantee}.
\newblock
  \bibinfo{journal}{\emph{\url{https://congweilin.github.io/CongWeilin.io/}}}
  (\bibinfo{year}{2022}).
\newblock


\bibitem[\protect\citeauthoryear{Cong and Mahdavi}{Cong and Mahdavi}{2023}]%
        {cong2023efficiently}
\bibfield{author}{\bibinfo{person}{Weilin Cong} {and} \bibinfo{person}{Mehrdad
  Mahdavi}.} \bibinfo{year}{2023}\natexlab{}.
\newblock \showarticletitle{Efficiently forgetting what you have learned in
  graph representation learning via projection}. In
  \bibinfo{booktitle}{\emph{AISTATS}}. \bibinfo{pages}{6674--6703}.
\newblock


\bibitem[\protect\citeauthoryear{Dai, Dong, et~al\mbox{.}}{Dai
  et~al\mbox{.}}{2022}]%
        {DaiDHSCW22}
\bibfield{author}{\bibinfo{person}{Damai Dai}, \bibinfo{person}{Li Dong},
  {et~al\mbox{.}}} \bibinfo{year}{2022}\natexlab{}.
\newblock \showarticletitle{Knowledge Neurons in Pretrained Transformers}. In
  \bibinfo{booktitle}{\emph{ACL}}. \bibinfo{pages}{8493--8502}.
\newblock


\bibitem[\protect\citeauthoryear{Dang}{Dang}{2021}]%
        {dang2021right}
\bibfield{author}{\bibinfo{person}{Quang-Vinh Dang}.}
  \bibinfo{year}{2021}\natexlab{}.
\newblock \showarticletitle{Right to Be Forgotten in the Age of Machine
  Learning}. In \bibinfo{booktitle}{\emph{ICADS}}. \bibinfo{pages}{403--411}.
\newblock


\bibitem[\protect\citeauthoryear{Deng, Dong, Socher, et~al\mbox{.}}{Deng
  et~al\mbox{.}}{2009}]%
        {deng2009imagenet}
\bibfield{author}{\bibinfo{person}{Jia Deng}, \bibinfo{person}{Wei Dong},
  \bibinfo{person}{Richard Socher}, {et~al\mbox{.}}}
  \bibinfo{year}{2009}\natexlab{}.
\newblock \showarticletitle{Imagenet: A large-scale hierarchical image
  database}. In \bibinfo{booktitle}{\emph{CVPR}}. \bibinfo{pages}{248--255}.
\newblock


\bibitem[\protect\citeauthoryear{Dinsdale, Jenkinson, et~al\mbox{.}}{Dinsdale
  et~al\mbox{.}}{2020}]%
        {dinsdale2020unlearning}
\bibfield{author}{\bibinfo{person}{Nicola~K Dinsdale}, \bibinfo{person}{Mark
  Jenkinson}, {et~al\mbox{.}}} \bibinfo{year}{2020}\natexlab{}.
\newblock \showarticletitle{Unlearning scanner bias for mri harmonisation}. In
  \bibinfo{booktitle}{\emph{MICCAI}}. \bibinfo{pages}{369--378}.
\newblock


\bibitem[\protect\citeauthoryear{Dinsdale, Jenkinson, and Namburete}{Dinsdale
  et~al\mbox{.}}{2021}]%
        {dinsdale2021deep}
\bibfield{author}{\bibinfo{person}{Nicola~K Dinsdale}, \bibinfo{person}{Mark
  Jenkinson}, {and} \bibinfo{person}{Ana~IL Namburete}.}
  \bibinfo{year}{2021}\natexlab{}.
\newblock \showarticletitle{Deep learning-based unlearning of dataset bias for
  MRI harmonisation and confound removal}.
\newblock \bibinfo{journal}{\emph{NeuroImage}}  \bibinfo{volume}{228}
  (\bibinfo{year}{2021}), \bibinfo{pages}{117689}.
\newblock


\bibitem[\protect\citeauthoryear{Du, Chen, et~al\mbox{.}}{Du
  et~al\mbox{.}}{2019}]%
        {du2019lifelong}
\bibfield{author}{\bibinfo{person}{Min Du}, \bibinfo{person}{Zhi Chen},
  {et~al\mbox{.}}} \bibinfo{year}{2019}\natexlab{}.
\newblock \showarticletitle{Lifelong anomaly detection through unlearning}. In
  \bibinfo{booktitle}{\emph{SIGSAC}}. \bibinfo{pages}{1283--1297}.
\newblock


\bibitem[\protect\citeauthoryear{Duan, Li, He, and Zeng}{Duan
  et~al\mbox{.}}{2007}]%
        {duan2007decremental}
\bibfield{author}{\bibinfo{person}{Hua Duan}, \bibinfo{person}{Hua Li},
  \bibinfo{person}{Guoping He}, {and} \bibinfo{person}{Qingtian Zeng}.}
  \bibinfo{year}{2007}\natexlab{}.
\newblock \showarticletitle{Decremental learning algorithms for nonlinear
  langrangian and least squares support vector machines}. In
  \bibinfo{booktitle}{\emph{OSB}}. \bibinfo{pages}{358--366}.
\newblock


\bibitem[\protect\citeauthoryear{Duda, Jaworski, Cader, and Wang}{Duda
  et~al\mbox{.}}{2020}]%
        {duda2020training}
\bibfield{author}{\bibinfo{person}{Piotr Duda}, \bibinfo{person}{Maciej
  Jaworski}, \bibinfo{person}{Andrzej Cader}, {and} \bibinfo{person}{Lipo
  Wang}.} \bibinfo{year}{2020}\natexlab{}.
\newblock \showarticletitle{On training deep neural networks using a streaming
  approach}.
\newblock \bibinfo{journal}{\emph{JAISCR}}  \bibinfo{volume}{10}
  (\bibinfo{year}{2020}).
\newblock


\bibitem[\protect\citeauthoryear{Dukler, Bowman, Achille, Golatkar,
  Swaminathan, and Soatto}{Dukler et~al\mbox{.}}{2023}]%
        {dukler2023safe}
\bibfield{author}{\bibinfo{person}{Yonatan Dukler}, \bibinfo{person}{Benjamin
  Bowman}, \bibinfo{person}{Alessandro Achille}, \bibinfo{person}{Aditya
  Golatkar}, \bibinfo{person}{Ashwin Swaminathan}, {and}
  \bibinfo{person}{Stefano Soatto}.} \bibinfo{year}{2023}\natexlab{}.
\newblock \showarticletitle{Safe: Machine unlearning with shard graphs}. In
  \bibinfo{booktitle}{\emph{ICCV}}. \bibinfo{pages}{17108--17118}.
\newblock


\bibitem[\protect\citeauthoryear{Dwork}{Dwork}{2008}]%
        {dwork2008differential}
\bibfield{author}{\bibinfo{person}{Cynthia Dwork}.}
  \bibinfo{year}{2008}\natexlab{}.
\newblock \showarticletitle{Differential privacy: A survey of results}. In
  \bibinfo{booktitle}{\emph{TAMC}}. \bibinfo{pages}{1--19}.
\newblock


\bibitem[\protect\citeauthoryear{Dwork, Roth, et~al\mbox{.}}{Dwork
  et~al\mbox{.}}{2014}]%
        {dwork2014algorithmic}
\bibfield{author}{\bibinfo{person}{Cynthia Dwork}, \bibinfo{person}{Aaron
  Roth}, {et~al\mbox{.}}} \bibinfo{year}{2014}\natexlab{}.
\newblock \showarticletitle{The algorithmic foundations of differential
  privacy}.
\newblock \bibinfo{journal}{\emph{Foundations and Trends{\textregistered} in
  Theoretical Computer Science}} \bibinfo{volume}{9}, \bibinfo{number}{3--4}
  (\bibinfo{year}{2014}), \bibinfo{pages}{211--407}.
\newblock


\bibitem[\protect\citeauthoryear{Eisenhofer, Riepel, Chandrasekaran, Ghosh,
  Ohrimenko, and Papernot}{Eisenhofer et~al\mbox{.}}{2022}]%
        {eisenhofer2022verifiable}
\bibfield{author}{\bibinfo{person}{Thorsten Eisenhofer},
  \bibinfo{person}{Doreen Riepel}, \bibinfo{person}{Varun Chandrasekaran},
  \bibinfo{person}{Esha Ghosh}, \bibinfo{person}{Olga Ohrimenko}, {and}
  \bibinfo{person}{Nicolas Papernot}.} \bibinfo{year}{2022}\natexlab{}.
\newblock \showarticletitle{Verifiable and Provably Secure Machine Unlearning}.
\newblock \bibinfo{journal}{\emph{arXiv preprint arXiv:2210.09126}}
  (\bibinfo{year}{2022}).
\newblock


\bibitem[\protect\citeauthoryear{Felps, Schwickerath, Williams, Vuong, Briggs,
  Hunt, Sakmar, Saranchak, and Shumaker}{Felps et~al\mbox{.}}{2020}]%
        {felps2020class}
\bibfield{author}{\bibinfo{person}{Daniel~L Felps}, \bibinfo{person}{Amelia~D
  Schwickerath}, \bibinfo{person}{Joyce~D Williams}, \bibinfo{person}{Trung~N
  Vuong}, \bibinfo{person}{Alan Briggs}, \bibinfo{person}{Matthew Hunt},
  \bibinfo{person}{Evan Sakmar}, \bibinfo{person}{David~D Saranchak}, {and}
  \bibinfo{person}{Tyler Shumaker}.} \bibinfo{year}{2020}\natexlab{}.
\newblock \showarticletitle{Class Clown: Data Redaction in Machine Unlearning
  at Enterprise Scale}.
\newblock \bibinfo{journal}{\emph{arXiv preprint arXiv:2012.04699}}
  (\bibinfo{year}{2020}).
\newblock


\bibitem[\protect\citeauthoryear{Feuerriegel, Dolata, and Schwabe}{Feuerriegel
  et~al\mbox{.}}{2020}]%
        {feuerriegel2020fair}
\bibfield{author}{\bibinfo{person}{Stefan Feuerriegel},
  \bibinfo{person}{Mateusz Dolata}, {and} \bibinfo{person}{Gerhard Schwabe}.}
  \bibinfo{year}{2020}\natexlab{}.
\newblock \showarticletitle{Fair AI}.
\newblock \bibinfo{journal}{\emph{Business \& information systems engineering}}
  \bibinfo{volume}{62}, \bibinfo{number}{4} (\bibinfo{year}{2020}),
  \bibinfo{pages}{379--384}.
\newblock


\bibitem[\protect\citeauthoryear{Fraboni, Van~Waerebeke, Scaman, Vidal, Kameni,
  and Lorenzi}{Fraboni et~al\mbox{.}}{2024}]%
        {fraboni2024sifu}
\bibfield{author}{\bibinfo{person}{Yann Fraboni}, \bibinfo{person}{Martin
  Van~Waerebeke}, \bibinfo{person}{Kevin Scaman}, \bibinfo{person}{Richard
  Vidal}, \bibinfo{person}{Laetitia Kameni}, {and} \bibinfo{person}{Marco
  Lorenzi}.} \bibinfo{year}{2024}\natexlab{}.
\newblock \showarticletitle{SIFU: Sequential Informed Federated Unlearning for
  Efficient and Provable Client Unlearning in Federated Optimization}. In
  \bibinfo{booktitle}{\emph{AISTATS}}. \bibinfo{pages}{3457--3465}.
\newblock


\bibitem[\protect\citeauthoryear{Fredrikson, Lantz, Jha, Lin, Page, and
  Ristenpart}{Fredrikson et~al\mbox{.}}{2014}]%
        {fredrikson2014privacy}
\bibfield{author}{\bibinfo{person}{Matthew Fredrikson}, \bibinfo{person}{Eric
  Lantz}, \bibinfo{person}{Somesh Jha}, \bibinfo{person}{Simon Lin},
  \bibinfo{person}{David Page}, {and} \bibinfo{person}{Thomas Ristenpart}.}
  \bibinfo{year}{2014}\natexlab{}.
\newblock \showarticletitle{Privacy in pharmacogenetics: An $\{$End-to-End$\}$
  case study of personalized warfarin dosing}. In
  \bibinfo{booktitle}{\emph{USENIX Security}}. \bibinfo{pages}{17--32}.
\newblock


\bibitem[\protect\citeauthoryear{Fu, He, et~al\mbox{.}}{Fu
  et~al\mbox{.}}{2022}]%
        {fu2022knowledge}
\bibfield{author}{\bibinfo{person}{Shaopeng Fu}, \bibinfo{person}{Fengxiang
  He}, {et~al\mbox{.}}} \bibinfo{year}{2022}\natexlab{}.
\newblock \showarticletitle{Knowledge Removal in Sampling-based Bayesian
  Inference}. In \bibinfo{booktitle}{\emph{ICLR}}.
\newblock


\bibitem[\protect\citeauthoryear{Fu, He, Xu, and Tao}{Fu et~al\mbox{.}}{2021}]%
        {fu2021bayesian}
\bibfield{author}{\bibinfo{person}{Shaopeng Fu}, \bibinfo{person}{Fengxiang
  He}, \bibinfo{person}{Yue Xu}, {and} \bibinfo{person}{Dacheng Tao}.}
  \bibinfo{year}{2021}\natexlab{}.
\newblock \showarticletitle{Bayesian inference forgetting}.
\newblock \bibinfo{journal}{\emph{arXiv preprint arXiv:2101.06417}}
  (\bibinfo{year}{2021}).
\newblock


\bibitem[\protect\citeauthoryear{Ganh{\"o}r, Penz, Rekabsaz, Lesota, and
  Schedl}{Ganh{\"o}r et~al\mbox{.}}{2022}]%
        {ganhor2022unlearning}
\bibfield{author}{\bibinfo{person}{Christian Ganh{\"o}r},
  \bibinfo{person}{David Penz}, \bibinfo{person}{Navid Rekabsaz},
  \bibinfo{person}{Oleg Lesota}, {and} \bibinfo{person}{Markus Schedl}.}
  \bibinfo{year}{2022}\natexlab{}.
\newblock \showarticletitle{Unlearning protected user attributes in
  recommendations with adversarial training}. In
  \bibinfo{booktitle}{\emph{Proceedings of the 45th International ACM SIGIR
  Conference on Research and Development in Information Retrieval}}.
  \bibinfo{pages}{2142--2147}.
\newblock


\bibitem[\protect\citeauthoryear{Gao, Garg, Mahmoody, and Vasudevan}{Gao
  et~al\mbox{.}}{2022a}]%
        {gao2022deletion}
\bibfield{author}{\bibinfo{person}{Ji Gao}, \bibinfo{person}{Sanjam Garg},
  \bibinfo{person}{Mohammad Mahmoody}, {and} \bibinfo{person}{Prashant~Nalini
  Vasudevan}.} \bibinfo{year}{2022}\natexlab{a}.
\newblock \showarticletitle{Deletion Inference, Reconstruction, and Compliance
  in Machine (Un) Learning}.
\newblock \bibinfo{journal}{\emph{Proc. Priv. Enhancing Technol.}}
  \bibinfo{volume}{2022}, \bibinfo{number}{3} (\bibinfo{year}{2022}),
  \bibinfo{pages}{415--436}.
\newblock


\bibitem[\protect\citeauthoryear{Gao, Ma, Wang, Sun, Li, Ji, Cheng, and
  Chen}{Gao et~al\mbox{.}}{2022b}]%
        {gao2022verifi}
\bibfield{author}{\bibinfo{person}{Xiangshan Gao}, \bibinfo{person}{Xingjun
  Ma}, \bibinfo{person}{Jingyi Wang}, \bibinfo{person}{Youcheng Sun},
  \bibinfo{person}{Bo Li}, \bibinfo{person}{Shouling Ji}, \bibinfo{person}{Peng
  Cheng}, {and} \bibinfo{person}{Jiming Chen}.}
  \bibinfo{year}{2022}\natexlab{b}.
\newblock \showarticletitle{VeriFi: Towards Verifiable Federated Unlearning}.
\newblock \bibinfo{journal}{\emph{arXiv preprint arXiv:2205.12709}}
  (\bibinfo{year}{2022}).
\newblock


\bibitem[\protect\citeauthoryear{Garg, Goldwasser, and Vasudevan}{Garg
  et~al\mbox{.}}{2020}]%
        {garg2020formalizing}
\bibfield{author}{\bibinfo{person}{Sanjam Garg}, \bibinfo{person}{Shafi
  Goldwasser}, {and} \bibinfo{person}{Prashant~Nalini Vasudevan}.}
  \bibinfo{year}{2020}\natexlab{}.
\newblock \showarticletitle{Formalizing data deletion in the context of the
  right to be forgotten}. In \bibinfo{booktitle}{\emph{EUROCRYPT}}.
  \bibinfo{pages}{373--402}.
\newblock


\bibitem[\protect\citeauthoryear{Gasteiger, Bojchevski, and
  G{\"u}nnemann}{Gasteiger et~al\mbox{.}}{2019}]%
        {gasteiger2018combining}
\bibfield{author}{\bibinfo{person}{Johannes Gasteiger},
  \bibinfo{person}{Aleksandar Bojchevski}, {and} \bibinfo{person}{Stephan
  G{\"u}nnemann}.} \bibinfo{year}{2019}\natexlab{}.
\newblock \showarticletitle{Combining Neural Networks with Personalized
  PageRank for Classification on Graphs}. In \bibinfo{booktitle}{\emph{ICLR}}.
\newblock


\bibitem[\protect\citeauthoryear{Geurts, Ernst, et~al\mbox{.}}{Geurts
  et~al\mbox{.}}{2006}]%
        {geurts2006extremely}
\bibfield{author}{\bibinfo{person}{Pierre Geurts}, \bibinfo{person}{Damien
  Ernst}, {et~al\mbox{.}}} \bibinfo{year}{2006}\natexlab{}.
\newblock \showarticletitle{Extremely randomized trees}.
\newblock \bibinfo{journal}{\emph{Machine learning}} \bibinfo{volume}{63},
  \bibinfo{number}{1} (\bibinfo{year}{2006}), \bibinfo{pages}{3--42}.
\newblock


\bibitem[\protect\citeauthoryear{Ginart, Guan, Valiant, and Zou}{Ginart
  et~al\mbox{.}}{2019}]%
        {ginart2019making}
\bibfield{author}{\bibinfo{person}{Antonio Ginart}, \bibinfo{person}{Melody
  Guan}, \bibinfo{person}{Gregory Valiant}, {and} \bibinfo{person}{James~Y
  Zou}.} \bibinfo{year}{2019}\natexlab{}.
\newblock \showarticletitle{Making ai forget you: Data deletion in machine
  learning}.
\newblock \bibinfo{journal}{\emph{NIPS}}  \bibinfo{volume}{32}
  (\bibinfo{year}{2019}).
\newblock


\bibitem[\protect\citeauthoryear{Goel, Prabhu, and Kumaraguru}{Goel
  et~al\mbox{.}}{2022}]%
        {goel2022evaluating}
\bibfield{author}{\bibinfo{person}{Shashwat Goel}, \bibinfo{person}{Ameya
  Prabhu}, {and} \bibinfo{person}{Ponnurangam Kumaraguru}.}
  \bibinfo{year}{2022}\natexlab{}.
\newblock \showarticletitle{Evaluating Inexact Unlearning Requires Revisiting
  Forgetting}.
\newblock \bibinfo{journal}{\emph{arXiv preprint arXiv:2201.06640}}
  (\bibinfo{year}{2022}).
\newblock


\bibitem[\protect\citeauthoryear{Golatkar, Achille, Ravichandran, Polito, and
  Soatto}{Golatkar et~al\mbox{.}}{2021}]%
        {golatkar2021mixed}
\bibfield{author}{\bibinfo{person}{Aditya Golatkar},
  \bibinfo{person}{Alessandro Achille}, \bibinfo{person}{Avinash Ravichandran},
  \bibinfo{person}{Marzia Polito}, {and} \bibinfo{person}{Stefano Soatto}.}
  \bibinfo{year}{2021}\natexlab{}.
\newblock \showarticletitle{Mixed-privacy forgetting in deep networks}. In
  \bibinfo{booktitle}{\emph{CVPR}}. \bibinfo{pages}{792--801}.
\newblock


\bibitem[\protect\citeauthoryear{Golatkar, Achille, and Soatto}{Golatkar
  et~al\mbox{.}}{2020a}]%
        {golatkar2020eternal}
\bibfield{author}{\bibinfo{person}{Aditya Golatkar},
  \bibinfo{person}{Alessandro Achille}, {and} \bibinfo{person}{Stefano
  Soatto}.} \bibinfo{year}{2020}\natexlab{a}.
\newblock \showarticletitle{Eternal sunshine of the spotless net: Selective
  forgetting in deep networks}. In \bibinfo{booktitle}{\emph{CVPR}}.
  \bibinfo{pages}{9304--9312}.
\newblock


\bibitem[\protect\citeauthoryear{Golatkar, Achille, and Soatto}{Golatkar
  et~al\mbox{.}}{2020b}]%
        {golatkar2020forgetting}
\bibfield{author}{\bibinfo{person}{Aditya Golatkar},
  \bibinfo{person}{Alessandro Achille}, {and} \bibinfo{person}{Stefano
  Soatto}.} \bibinfo{year}{2020}\natexlab{b}.
\newblock \showarticletitle{Forgetting outside the box: Scrubbing deep networks
  of information accessible from input-output observations}. In
  \bibinfo{booktitle}{\emph{ECCV}}. \bibinfo{pages}{383--398}.
\newblock


\bibitem[\protect\citeauthoryear{Goyal, Hassija, and Albuquerque}{Goyal
  et~al\mbox{.}}{2021}]%
        {goyal2021revisiting}
\bibfield{author}{\bibinfo{person}{Adit Goyal}, \bibinfo{person}{Vikas
  Hassija}, {and} \bibinfo{person}{Victor Hugo C~de Albuquerque}.}
  \bibinfo{year}{2021}\natexlab{}.
\newblock \showarticletitle{Revisiting Machine Learning Training Process for
  Enhanced Data Privacy}. In \bibinfo{booktitle}{\emph{IC3}}.
  \bibinfo{pages}{247--251}.
\newblock


\bibitem[\protect\citeauthoryear{Graves, Nagisetty, and Ganesh}{Graves
  et~al\mbox{.}}{2021}]%
        {graves2021amnesiac}
\bibfield{author}{\bibinfo{person}{Laura Graves}, \bibinfo{person}{Vineel
  Nagisetty}, {and} \bibinfo{person}{Vijay Ganesh}.}
  \bibinfo{year}{2021}\natexlab{}.
\newblock \showarticletitle{Amnesiac machine learning}. In
  \bibinfo{booktitle}{\emph{AAAI}}, Vol.~\bibinfo{volume}{35}.
  \bibinfo{pages}{11516--11524}.
\newblock


\bibitem[\protect\citeauthoryear{Guo, Goldstein, Hannun, and van~der
  Maaten}{Guo et~al\mbox{.}}{2020b}]%
        {GuoGHM20}
\bibfield{author}{\bibinfo{person}{Chuan Guo}, \bibinfo{person}{Tom Goldstein},
  \bibinfo{person}{Awni~Y. Hannun}, {and} \bibinfo{person}{Laurens van~der
  Maaten}.} \bibinfo{year}{2020}\natexlab{b}.
\newblock \showarticletitle{Certified Data Removal from Machine Learning
  Models}. In \bibinfo{booktitle}{\emph{ICML}}, Vol.~\bibinfo{volume}{119}.
  \bibinfo{pages}{3832--3842}.
\newblock


\bibitem[\protect\citeauthoryear{Guo, Cheng, Li, Hahn, and Liu}{Guo
  et~al\mbox{.}}{2020a}]%
        {guo2020survey}
\bibfield{author}{\bibinfo{person}{Ruocheng Guo}, \bibinfo{person}{Lu Cheng},
  \bibinfo{person}{Jundong Li}, \bibinfo{person}{P~Richard Hahn}, {and}
  \bibinfo{person}{Huan Liu}.} \bibinfo{year}{2020}\natexlab{a}.
\newblock \showarticletitle{A survey of learning causality with data: Problems
  and methods}.
\newblock \bibinfo{journal}{\emph{CSUR}} \bibinfo{volume}{53},
  \bibinfo{number}{4} (\bibinfo{year}{2020}), \bibinfo{pages}{1--37}.
\newblock


\bibitem[\protect\citeauthoryear{Guo, Guo, Zhang, Xu, and Wang}{Guo
  et~al\mbox{.}}{2022}]%
        {guo2022efficient}
\bibfield{author}{\bibinfo{person}{Tao Guo}, \bibinfo{person}{Song Guo},
  \bibinfo{person}{Jiewei Zhang}, \bibinfo{person}{Wenchao Xu}, {and}
  \bibinfo{person}{Junxiao Wang}.} \bibinfo{year}{2022}\natexlab{}.
\newblock \showarticletitle{Efficient Attribute Unlearning: Towards Selective
  Removal of Input Attributes from Feature Representations}.
\newblock \bibinfo{journal}{\emph{arXiv preprint arXiv:2202.13295}}
  (\bibinfo{year}{2022}).
\newblock


\bibitem[\protect\citeauthoryear{Gupta, Jung, Neel, Roth, Sharifi-Malvajerdi,
  and Waites}{Gupta et~al\mbox{.}}{2021}]%
        {gupta2021adaptive}
\bibfield{author}{\bibinfo{person}{Varun Gupta}, \bibinfo{person}{Christopher
  Jung}, \bibinfo{person}{Seth Neel}, \bibinfo{person}{Aaron Roth},
  \bibinfo{person}{Saeed Sharifi-Malvajerdi}, {and} \bibinfo{person}{Chris
  Waites}.} \bibinfo{year}{2021}\natexlab{}.
\newblock \showarticletitle{Adaptive machine unlearning}.
\newblock \bibinfo{journal}{\emph{NIPS}}  \bibinfo{volume}{34}
  (\bibinfo{year}{2021}), \bibinfo{pages}{16319--16330}.
\newblock


\bibitem[\protect\citeauthoryear{Halimi, Kadhe, Rawat, and Baracaldo}{Halimi
  et~al\mbox{.}}{2022}]%
        {halimi2022federated}
\bibfield{author}{\bibinfo{person}{Anisa Halimi}, \bibinfo{person}{Swanand
  Kadhe}, \bibinfo{person}{Ambrish Rawat}, {and} \bibinfo{person}{Nathalie
  Baracaldo}.} \bibinfo{year}{2022}\natexlab{}.
\newblock \showarticletitle{Federated Unlearning: How to Efficiently Erase a
  Client in FL?}
\newblock \bibinfo{journal}{\emph{arXiv preprint arXiv:2207.05521}}
  (\bibinfo{year}{2022}).
\newblock


\bibitem[\protect\citeauthoryear{Hamilton}{Hamilton}{2020}]%
        {hamilton2020graph}
\bibfield{author}{\bibinfo{person}{William~L Hamilton}.}
  \bibinfo{year}{2020}\natexlab{}.
\newblock \showarticletitle{Graph representation learning}.
\newblock \bibinfo{journal}{\emph{Synthesis Lectures on Artifical Intelligence
  and Machine Learning}} \bibinfo{volume}{14}, \bibinfo{number}{3}
  (\bibinfo{year}{2020}), \bibinfo{pages}{1--159}.
\newblock


\bibitem[\protect\citeauthoryear{Haug and Kasneci}{Haug and Kasneci}{2021}]%
        {haug2021learning}
\bibfield{author}{\bibinfo{person}{Johannes Haug} {and}
  \bibinfo{person}{Gjergji Kasneci}.} \bibinfo{year}{2021}\natexlab{}.
\newblock \showarticletitle{Learning parameter distributions to detect concept
  drift in data streams}. In \bibinfo{booktitle}{\emph{ICPR}}.
  \bibinfo{pages}{9452--9459}.
\newblock


\bibitem[\protect\citeauthoryear{Hawkins, Alhuwaish, Belguith, Vranaki, and
  Charlesworth}{Hawkins et~al\mbox{.}}{2023}]%
        {hawkins2023decision}
\bibfield{author}{\bibinfo{person}{Katie Hawkins}, \bibinfo{person}{Nora
  Alhuwaish}, \bibinfo{person}{Sana Belguith}, \bibinfo{person}{Asma Vranaki},
  {and} \bibinfo{person}{Andrew Charlesworth}.}
  \bibinfo{year}{2023}\natexlab{}.
\newblock \showarticletitle{A Decision-Making Process to Implement the `Right
  to Be Forgotten'in Machine Learning}. In \bibinfo{booktitle}{\emph{Annual
  Privacy Forum}}. \bibinfo{pages}{20--38}.
\newblock


\bibitem[\protect\citeauthoryear{He, Meng, Chen, He, and Hu}{He
  et~al\mbox{.}}{2021}]%
        {he2021deepobliviate}
\bibfield{author}{\bibinfo{person}{Yingzhe He}, \bibinfo{person}{Guozhu Meng},
  \bibinfo{person}{Kai Chen}, \bibinfo{person}{Jinwen He}, {and}
  \bibinfo{person}{Xingbo Hu}.} \bibinfo{year}{2021}\natexlab{}.
\newblock \showarticletitle{Deepobliviate: a powerful charm for erasing data
  residual memory in deep neural networks}.
\newblock \bibinfo{journal}{\emph{arXiv preprint arXiv:2105.06209}}
  (\bibinfo{year}{2021}).
\newblock


\bibitem[\protect\citeauthoryear{Hu, Li, Hu, Zheng, Liu, and Zhang}{Hu
  et~al\mbox{.}}{2024}]%
        {hu2024separate}
\bibfield{author}{\bibinfo{person}{Xinshuo Hu}, \bibinfo{person}{Dongfang Li},
  \bibinfo{person}{Baotian Hu}, \bibinfo{person}{Zihao Zheng},
  \bibinfo{person}{Zhenyu Liu}, {and} \bibinfo{person}{Min Zhang}.}
  \bibinfo{year}{2024}\natexlab{}.
\newblock \showarticletitle{Separate the wheat from the chaff: Model deficiency
  unlearning via parameter-efficient module operation}. In
  \bibinfo{booktitle}{\emph{AAAI}}, Vol.~\bibinfo{volume}{38}.
  \bibinfo{pages}{18252--18260}.
\newblock


\bibitem[\protect\citeauthoryear{Hu, Tang, Miao, Hua, and Zhang}{Hu
  et~al\mbox{.}}{2021}]%
        {hu2021distilling}
\bibfield{author}{\bibinfo{person}{Xinting Hu}, \bibinfo{person}{Kaihua Tang},
  \bibinfo{person}{Chunyan Miao}, \bibinfo{person}{Xian-Sheng Hua}, {and}
  \bibinfo{person}{Hanwang Zhang}.} \bibinfo{year}{2021}\natexlab{}.
\newblock \showarticletitle{Distilling causal effect of data in
  class-incremental learning}. In \bibinfo{booktitle}{\emph{CVPR}}.
  \bibinfo{pages}{3957--3966}.
\newblock


\bibitem[\protect\citeauthoryear{Huang, Ma, Erfani, Bailey, and Wang}{Huang
  et~al\mbox{.}}{2021b}]%
        {huang2021unlearnable}
\bibfield{author}{\bibinfo{person}{Hanxun Huang}, \bibinfo{person}{Xingjun Ma},
  \bibinfo{person}{Sarah~Monazam Erfani}, \bibinfo{person}{James Bailey}, {and}
  \bibinfo{person}{Yisen Wang}.} \bibinfo{year}{2021}\natexlab{b}.
\newblock \showarticletitle{Unlearnable Examples: Making Personal Data
  Unexploitable}. In \bibinfo{booktitle}{\emph{ICLR}}.
\newblock


\bibitem[\protect\citeauthoryear{Huang, Li, et~al\mbox{.}}{Huang
  et~al\mbox{.}}{2021a}]%
        {huang2021mathsf}
\bibfield{author}{\bibinfo{person}{Yangsibo Huang}, \bibinfo{person}{Xiaoxiao
  Li}, {et~al\mbox{.}}} \bibinfo{year}{2021}\natexlab{a}.
\newblock \showarticletitle{EMA: Auditing Data Removal from Trained Models}. In
  \bibinfo{booktitle}{\emph{MICCAI}}. \bibinfo{pages}{793--803}.
\newblock


\bibitem[\protect\citeauthoryear{H{\"u}llermeier and Waegeman}{H{\"u}llermeier
  and Waegeman}{2021}]%
        {hullermeier2021aleatoric}
\bibfield{author}{\bibinfo{person}{Eyke H{\"u}llermeier} {and}
  \bibinfo{person}{Willem Waegeman}.} \bibinfo{year}{2021}\natexlab{}.
\newblock \showarticletitle{Aleatoric and epistemic uncertainty in machine
  learning: An introduction to concepts and methods}.
\newblock \bibinfo{journal}{\emph{Machine Learning}} \bibinfo{volume}{110},
  \bibinfo{number}{3} (\bibinfo{year}{2021}), \bibinfo{pages}{457--506}.
\newblock


\bibitem[\protect\citeauthoryear{Huynh, Nguyen, Nguyen, Nguyen, Weidlich,
  Nguyen, and Aberer}{Huynh et~al\mbox{.}}{2024}]%
        {huynh2024fast}
\bibfield{author}{\bibinfo{person}{Thanh~Trung Huynh},
  \bibinfo{person}{Trong~Bang Nguyen}, \bibinfo{person}{Phi~Le Nguyen},
  \bibinfo{person}{Thanh~Tam Nguyen}, \bibinfo{person}{Matthias Weidlich},
  \bibinfo{person}{Quoc Viet~Hung Nguyen}, {and} \bibinfo{person}{Karl
  Aberer}.} \bibinfo{year}{2024}\natexlab{}.
\newblock \showarticletitle{Fast-fedul: A training-free federated unlearning
  with provable skew resilience}. In \bibinfo{booktitle}{\emph{ECML-PKDD}}.
  \bibinfo{pages}{55--72}.
\newblock


\bibitem[\protect\citeauthoryear{Izzo, Smart, Chaudhuri, and Zou}{Izzo
  et~al\mbox{.}}{2021}]%
        {izzo2021approximate}
\bibfield{author}{\bibinfo{person}{Zachary Izzo}, \bibinfo{person}{Mary~Anne
  Smart}, \bibinfo{person}{Kamalika Chaudhuri}, {and} \bibinfo{person}{James
  Zou}.} \bibinfo{year}{2021}\natexlab{}.
\newblock \showarticletitle{Approximate data deletion from machine learning
  models}. In \bibinfo{booktitle}{\emph{AISTAT}}. \bibinfo{pages}{2008--2016}.
\newblock


\bibitem[\protect\citeauthoryear{Jagielski, Thakkar, Tram{\`e}r, Ippolito, Lee,
  Carlini, Wallace, Song, Thakurta, Papernot, et~al\mbox{.}}{Jagielski
  et~al\mbox{.}}{2022}]%
        {jagielski2022measuring}
\bibfield{author}{\bibinfo{person}{Matthew Jagielski}, \bibinfo{person}{Om
  Thakkar}, \bibinfo{person}{Florian Tram{\`e}r}, \bibinfo{person}{Daphne
  Ippolito}, \bibinfo{person}{Katherine Lee}, \bibinfo{person}{Nicholas
  Carlini}, \bibinfo{person}{Eric Wallace}, \bibinfo{person}{Shuang Song},
  \bibinfo{person}{Abhradeep Thakurta}, \bibinfo{person}{Nicolas Papernot},
  {et~al\mbox{.}}} \bibinfo{year}{2022}\natexlab{}.
\newblock \showarticletitle{Measuring Forgetting of Memorized Training
  Examples}.
\newblock \bibinfo{journal}{\emph{arXiv preprint arXiv:2207.00099}}
  (\bibinfo{year}{2022}).
\newblock


\bibitem[\protect\citeauthoryear{Jia, Yaghini, Choquette-Choo, Dullerud, Thudi,
  Chandrasekaran, and Papernot}{Jia et~al\mbox{.}}{2021}]%
        {jia2021proof}
\bibfield{author}{\bibinfo{person}{Hengrui Jia}, \bibinfo{person}{Mohammad
  Yaghini}, \bibinfo{person}{Christopher~A Choquette-Choo},
  \bibinfo{person}{Natalie Dullerud}, \bibinfo{person}{Anvith Thudi},
  \bibinfo{person}{Varun Chandrasekaran}, {and} \bibinfo{person}{Nicolas
  Papernot}.} \bibinfo{year}{2021}\natexlab{}.
\newblock \showarticletitle{Proof-of-learning: Definitions and practice}. In
  \bibinfo{booktitle}{\emph{SP}}. \bibinfo{pages}{1039--1056}.
\newblock


\bibitem[\protect\citeauthoryear{Jose and Simeone}{Jose and Simeone}{2021}]%
        {jose2021unified}
\bibfield{author}{\bibinfo{person}{Sharu~Theresa Jose} {and}
  \bibinfo{person}{Osvaldo Simeone}.} \bibinfo{year}{2021}\natexlab{}.
\newblock \showarticletitle{A unified PAC-Bayesian framework for machine
  unlearning via information risk minimization}. In
  \bibinfo{booktitle}{\emph{MLSP}}. \bibinfo{pages}{1--6}.
\newblock


\bibitem[\protect\citeauthoryear{Kadhe, Halimi, Rawat, and Baracaldo}{Kadhe
  et~al\mbox{.}}{2023}]%
        {kadhe2023fairsisa}
\bibfield{author}{\bibinfo{person}{Swanand Kadhe}, \bibinfo{person}{Anisa
  Halimi}, \bibinfo{person}{Ambrish Rawat}, {and} \bibinfo{person}{Nathalie
  Baracaldo}.} \bibinfo{year}{2023}\natexlab{}.
\newblock \showarticletitle{Fair{SISA}: Ensemble Post-Processing to Improve
  Fairness of Unlearning in {LLM}s}. In \bibinfo{booktitle}{\emph{SoLaR}}.
\newblock


\bibitem[\protect\citeauthoryear{Karasuyama and Takeuchi}{Karasuyama and
  Takeuchi}{2009}]%
        {karasuyama2009multiple}
\bibfield{author}{\bibinfo{person}{Masayuki Karasuyama} {and}
  \bibinfo{person}{Ichiro Takeuchi}.} \bibinfo{year}{2009}\natexlab{}.
\newblock \showarticletitle{Multiple incremental decremental learning of
  support vector machines}.
\newblock \bibinfo{journal}{\emph{NIPS}}  \bibinfo{volume}{22}
  (\bibinfo{year}{2009}).
\newblock


\bibitem[\protect\citeauthoryear{Karasuyama and Takeuchi}{Karasuyama and
  Takeuchi}{2010}]%
        {karasuyama2010multiple}
\bibfield{author}{\bibinfo{person}{Masayuki Karasuyama} {and}
  \bibinfo{person}{Ichiro Takeuchi}.} \bibinfo{year}{2010}\natexlab{}.
\newblock \showarticletitle{Multiple incremental decremental learning of
  support vector machines}.
\newblock \bibinfo{journal}{\emph{IEEE Transactions on Neural Networks}}
  \bibinfo{volume}{21}, \bibinfo{number}{7} (\bibinfo{year}{2010}),
  \bibinfo{pages}{1048--1059}.
\newblock


\bibitem[\protect\citeauthoryear{Kassem, Mahmoud, and Saad}{Kassem
  et~al\mbox{.}}{2023}]%
        {kassem2023preserving}
\bibfield{author}{\bibinfo{person}{Aly Kassem}, \bibinfo{person}{Omar Mahmoud},
  {and} \bibinfo{person}{Sherif Saad}.} \bibinfo{year}{2023}\natexlab{}.
\newblock \showarticletitle{Preserving privacy through dememorization: An
  unlearning technique for mitigating memorization risks in language models}.
  In \bibinfo{booktitle}{\emph{EMNLP}}. \bibinfo{pages}{4360--4379}.
\newblock


\bibitem[\protect\citeauthoryear{Kearns}{Kearns}{1998}]%
        {kearns1998efficient}
\bibfield{author}{\bibinfo{person}{Michael Kearns}.}
  \bibinfo{year}{1998}\natexlab{}.
\newblock \showarticletitle{Efficient noise-tolerant learning from statistical
  queries}.
\newblock \bibinfo{journal}{\emph{JACM}} \bibinfo{volume}{45},
  \bibinfo{number}{6} (\bibinfo{year}{1998}), \bibinfo{pages}{983--1006}.
\newblock


\bibitem[\protect\citeauthoryear{Khan et~al\mbox{.}}{Khan
  et~al\mbox{.}}{2021}]%
        {khan2021knowledge}
\bibfield{author}{\bibinfo{person}{Mohammad Emtiyaz~E Khan} {et~al\mbox{.}}}
  \bibinfo{year}{2021}\natexlab{}.
\newblock \showarticletitle{Knowledge-adaptation priors}.
\newblock \bibinfo{journal}{\emph{NIPS}}  \bibinfo{volume}{34}
  (\bibinfo{year}{2021}), \bibinfo{pages}{19757--19770}.
\newblock


\bibitem[\protect\citeauthoryear{Kim, Lee, and Woo}{Kim et~al\mbox{.}}{2024}]%
        {kim2024layer}
\bibfield{author}{\bibinfo{person}{Hyunjune Kim}, \bibinfo{person}{Sangyong
  Lee}, {and} \bibinfo{person}{Simon~S Woo}.} \bibinfo{year}{2024}\natexlab{}.
\newblock \showarticletitle{Layer Attack Unlearning: Fast and Accurate Machine
  Unlearning via Layer Level Attack and Knowledge Distillation}. In
  \bibinfo{booktitle}{\emph{AAAI}}, Vol.~\bibinfo{volume}{38}.
  \bibinfo{pages}{21241--21248}.
\newblock


\bibitem[\protect\citeauthoryear{Kirkpatrick, Pascanu, Rabinowitz, Veness,
  Desjardins, Rusu, Milan, Quan, Ramalho, Grabska-Barwinska,
  et~al\mbox{.}}{Kirkpatrick et~al\mbox{.}}{2017}]%
        {kirkpatrick2017overcoming}
\bibfield{author}{\bibinfo{person}{James Kirkpatrick}, \bibinfo{person}{Razvan
  Pascanu}, \bibinfo{person}{Neil Rabinowitz}, \bibinfo{person}{Joel Veness},
  \bibinfo{person}{Guillaume Desjardins}, \bibinfo{person}{Andrei~A Rusu},
  \bibinfo{person}{Kieran Milan}, \bibinfo{person}{John Quan},
  \bibinfo{person}{Tiago Ramalho}, \bibinfo{person}{Agnieszka
  Grabska-Barwinska}, {et~al\mbox{.}}} \bibinfo{year}{2017}\natexlab{}.
\newblock \showarticletitle{Overcoming catastrophic forgetting in neural
  networks}.
\newblock \bibinfo{journal}{\emph{PNAS}} \bibinfo{volume}{114},
  \bibinfo{number}{13} (\bibinfo{year}{2017}), \bibinfo{pages}{3521--3526}.
\newblock


\bibitem[\protect\citeauthoryear{Koh et~al\mbox{.}}{Koh et~al\mbox{.}}{2017}]%
        {koh2017understanding}
\bibfield{author}{\bibinfo{person}{Pang~Wei Koh} {et~al\mbox{.}}}
  \bibinfo{year}{2017}\natexlab{}.
\newblock \showarticletitle{Understanding black-box predictions via influence
  functions}. In \bibinfo{booktitle}{\emph{ICML}}. \bibinfo{pages}{1885--1894}.
\newblock


\bibitem[\protect\citeauthoryear{Kullback et~al\mbox{.}}{Kullback
  et~al\mbox{.}}{1951}]%
        {KLFormula}
\bibfield{author}{\bibinfo{person}{S. Kullback} {et~al\mbox{.}}}
  \bibinfo{year}{1951}\natexlab{}.
\newblock \showarticletitle{{On Information and Sufficiency}}.
\newblock \bibinfo{journal}{\emph{The Annals of Mathematical Statistics}}
  \bibinfo{volume}{22}, \bibinfo{number}{1} (\bibinfo{year}{1951}),
  \bibinfo{pages}{79 -- 86}.
\newblock


\bibitem[\protect\citeauthoryear{Lee and Choi}{Lee and Choi}{2024}]%
        {lee2024learning}
\bibfield{author}{\bibinfo{person}{Sangyoon Lee} {and}
  \bibinfo{person}{Dae-Hyun Choi}.} \bibinfo{year}{2024}\natexlab{}.
\newblock \showarticletitle{Learning and Unlearning to Operate Profitable
  Secure Electric Vehicle Charging}.
\newblock \bibinfo{journal}{\emph{TII}} (\bibinfo{year}{2024}).
\newblock


\bibitem[\protect\citeauthoryear{Lee, Tanveer, Rahmani, Alinejad-Rokny,
  Khoshvaght, Zare, Alamdari, and Hosseinzadeh}{Lee et~al\mbox{.}}{2024}]%
        {lee2024sfgcn}
\bibfield{author}{\bibinfo{person}{Sang-Woong Lee}, \bibinfo{person}{Jawad
  Tanveer}, \bibinfo{person}{Amir~Masoud Rahmani}, \bibinfo{person}{Hamid
  Alinejad-Rokny}, \bibinfo{person}{Parisa Khoshvaght},
  \bibinfo{person}{Gholamreza Zare}, \bibinfo{person}{Pegah~Malekpour
  Alamdari}, {and} \bibinfo{person}{Mehdi Hosseinzadeh}.}
  \bibinfo{year}{2024}\natexlab{}.
\newblock \showarticletitle{SFGCN: Synergetic Fusion-based Graph Convolutional
  Networks Approach for Link Prediction in Social Networks}.
\newblock \bibinfo{journal}{\emph{Information Fusion}} (\bibinfo{year}{2024}),
  \bibinfo{pages}{102684}.
\newblock


\bibitem[\protect\citeauthoryear{Lei, Su, Cui, Yau, and Gu}{Lei
  et~al\mbox{.}}{2019}]%
        {lei2019geometric}
\bibfield{author}{\bibinfo{person}{Na Lei}, \bibinfo{person}{Kehua Su},
  \bibinfo{person}{Li Cui}, \bibinfo{person}{Shing-Tung Yau}, {and}
  \bibinfo{person}{Xianfeng~David Gu}.} \bibinfo{year}{2019}\natexlab{}.
\newblock \showarticletitle{A geometric view of optimal transportation and
  generative model}.
\newblock \bibinfo{journal}{\emph{Computer Aided Geometric Design}}
  \bibinfo{volume}{68} (\bibinfo{year}{2019}), \bibinfo{pages}{1--21}.
\newblock


\bibitem[\protect\citeauthoryear{Li, Hsu, Chen, and Marculescu}{Li
  et~al\mbox{.}}{2024a}]%
        {li2024machine}
\bibfield{author}{\bibinfo{person}{Guihong Li}, \bibinfo{person}{Hsiang Hsu},
  \bibinfo{person}{Chun-Fu Chen}, {and} \bibinfo{person}{Radu Marculescu}.}
  \bibinfo{year}{2024}\natexlab{a}.
\newblock \showarticletitle{Machine Unlearning for Image-to-Image Generative
  Models}. In \bibinfo{booktitle}{\emph{ICLR}}.
\newblock


\bibitem[\protect\citeauthoryear{Li, Zhao, Wu, Zhang, Li, and Wang}{Li
  et~al\mbox{.}}{2024b}]%
        {li2024towards}
\bibfield{author}{\bibinfo{person}{Xunkai Li}, \bibinfo{person}{Yulin Zhao},
  \bibinfo{person}{Zhengyu Wu}, \bibinfo{person}{Wentao Zhang},
  \bibinfo{person}{Rong-Hua Li}, {and} \bibinfo{person}{Guoren Wang}.}
  \bibinfo{year}{2024}\natexlab{b}.
\newblock \showarticletitle{Towards Effective and General Graph Unlearning via
  Mutual Evolution}. In \bibinfo{booktitle}{\emph{AAAI}},
  Vol.~\bibinfo{volume}{38}. \bibinfo{pages}{13682--13690}.
\newblock


\bibitem[\protect\citeauthoryear{Li, Chen, Zheng, Zhang, Han, Meng, and
  Wang}{Li et~al\mbox{.}}{2023}]%
        {li2023making}
\bibfield{author}{\bibinfo{person}{Yuyuan Li}, \bibinfo{person}{Chaochao Chen},
  \bibinfo{person}{Xiaolin Zheng}, \bibinfo{person}{Yizhao Zhang},
  \bibinfo{person}{Zhongxuan Han}, \bibinfo{person}{Dan Meng}, {and}
  \bibinfo{person}{Jun Wang}.} \bibinfo{year}{2023}\natexlab{}.
\newblock \showarticletitle{Making users indistinguishable: Attribute-wise
  unlearning in recommender systems}. In \bibinfo{booktitle}{\emph{MM}}.
  \bibinfo{pages}{984--994}.
\newblock


\bibitem[\protect\citeauthoryear{Li, Wang, and Cheng}{Li et~al\mbox{.}}{2021}]%
        {li2020online}
\bibfield{author}{\bibinfo{person}{Yuantong Li}, \bibinfo{person}{Chi-Hua
  Wang}, {and} \bibinfo{person}{Guang Cheng}.} \bibinfo{year}{2021}\natexlab{}.
\newblock \showarticletitle{Online Forgetting Process for Linear Regression
  Models}. In \bibinfo{booktitle}{\emph{AISTAT}}, Vol.~\bibinfo{volume}{130}.
  \bibinfo{pages}{217--225}.
\newblock


\bibitem[\protect\citeauthoryear{Lin, Zhang, Chen, Chen, and Susilo}{Lin
  et~al\mbox{.}}{2023}]%
        {lin2023erm}
\bibfield{author}{\bibinfo{person}{Shen Lin}, \bibinfo{person}{Xiaoyu Zhang},
  \bibinfo{person}{Chenyang Chen}, \bibinfo{person}{Xiaofeng Chen}, {and}
  \bibinfo{person}{Willy Susilo}.} \bibinfo{year}{2023}\natexlab{}.
\newblock \showarticletitle{Erm-ktp: Knowledge-level machine unlearning via
  knowledge transfer}. In \bibinfo{booktitle}{\emph{CVPR}}.
  \bibinfo{pages}{20147--20155}.
\newblock


\bibitem[\protect\citeauthoryear{Liu, Liu, et~al\mbox{.}}{Liu
  et~al\mbox{.}}{2022b}]%
        {liu2022continual}
\bibfield{author}{\bibinfo{person}{Bo Liu}, \bibinfo{person}{Qiang Liu},
  {et~al\mbox{.}}} \bibinfo{year}{2022}\natexlab{b}.
\newblock \showarticletitle{Continual Learning and Private Unlearning}.
\newblock \bibinfo{journal}{\emph{arXiv preprint arXiv:2203.12817}}
  (\bibinfo{year}{2022}).
\newblock


\bibitem[\protect\citeauthoryear{Liu, Ma, Yang, Wang, and Liu}{Liu
  et~al\mbox{.}}{2020b}]%
        {liu2020federated}
\bibfield{author}{\bibinfo{person}{Gaoyang Liu}, \bibinfo{person}{Xiaoqiang
  Ma}, \bibinfo{person}{Yang Yang}, \bibinfo{person}{Chen Wang}, {and}
  \bibinfo{person}{Jiangchuan Liu}.} \bibinfo{year}{2020}\natexlab{b}.
\newblock \showarticletitle{Federated unlearning}.
\newblock \bibinfo{journal}{\emph{arXiv preprint arXiv:2012.13891}}
  (\bibinfo{year}{2020}).
\newblock


\bibitem[\protect\citeauthoryear{Liu, Ma, Yang, Wang, and Liu}{Liu
  et~al\mbox{.}}{2021b}]%
        {liu2021federaser}
\bibfield{author}{\bibinfo{person}{Gaoyang Liu}, \bibinfo{person}{Xiaoqiang
  Ma}, \bibinfo{person}{Yang Yang}, \bibinfo{person}{Chen Wang}, {and}
  \bibinfo{person}{Jiangchuan Liu}.} \bibinfo{year}{2021}\natexlab{b}.
\newblock \showarticletitle{Federaser: Enabling efficient client-level data
  removal from federated learning models}. In
  \bibinfo{booktitle}{\emph{IWQOS}}. \bibinfo{pages}{1--10}.
\newblock


\bibitem[\protect\citeauthoryear{Liu, Wan, Wang, Zhang, Zhang, and Li}{Liu
  et~al\mbox{.}}{2022c}]%
        {liu2022forgetting}
\bibfield{author}{\bibinfo{person}{Wenyan Liu}, \bibinfo{person}{Juncheng Wan},
  \bibinfo{person}{Xiaoling Wang}, \bibinfo{person}{Weinan Zhang},
  \bibinfo{person}{Dell Zhang}, {and} \bibinfo{person}{Hang Li}.}
  \bibinfo{year}{2022}\natexlab{c}.
\newblock \showarticletitle{Forgetting fast in recommender systems}.
\newblock \bibinfo{journal}{\emph{arXiv preprint arXiv:2208.06875}}
  (\bibinfo{year}{2022}).
\newblock


\bibitem[\protect\citeauthoryear{Liu and Tsaftaris}{Liu and Tsaftaris}{2020}]%
        {liu2020have}
\bibfield{author}{\bibinfo{person}{Xiao Liu} {and} \bibinfo{person}{Sotirios~A
  Tsaftaris}.} \bibinfo{year}{2020}\natexlab{}.
\newblock \showarticletitle{Have you forgotten? A method to assess if machine
  learning models have forgotten data}. In \bibinfo{booktitle}{\emph{MICCAI}}.
  \bibinfo{pages}{95--105}.
\newblock


\bibitem[\protect\citeauthoryear{Liu, Fan, Chen, Liu, Ma, Wang, and Ma}{Liu
  et~al\mbox{.}}{2022a}]%
        {liu2022backdoor}
\bibfield{author}{\bibinfo{person}{Yang Liu}, \bibinfo{person}{Mingyuan Fan},
  \bibinfo{person}{Cen Chen}, \bibinfo{person}{Ximeng Liu},
  \bibinfo{person}{Zhuo Ma}, \bibinfo{person}{Li Wang}, {and}
  \bibinfo{person}{Jianfeng Ma}.} \bibinfo{year}{2022}\natexlab{a}.
\newblock \showarticletitle{Backdoor Defense with Machine Unlearning}.
\newblock \bibinfo{journal}{\emph{arXiv preprint arXiv:2201.09538}}
  (\bibinfo{year}{2022}).
\newblock


\bibitem[\protect\citeauthoryear{Liu, Ma, Liu, Liu, Jiang, Ma, Yu, and Ren}{Liu
  et~al\mbox{.}}{2020a}]%
        {liu2020learn}
\bibfield{author}{\bibinfo{person}{Yang Liu}, \bibinfo{person}{Zhuo Ma},
  \bibinfo{person}{Ximeng Liu}, \bibinfo{person}{Jian Liu},
  \bibinfo{person}{Zhongyuan Jiang}, \bibinfo{person}{Jianfeng Ma},
  \bibinfo{person}{Philip Yu}, {and} \bibinfo{person}{Kui Ren}.}
  \bibinfo{year}{2020}\natexlab{a}.
\newblock \showarticletitle{Learn to forget: Machine unlearning via neuron
  masking}.
\newblock \bibinfo{journal}{\emph{arXiv preprint arXiv:2003.10933}}
  (\bibinfo{year}{2020}).
\newblock


\bibitem[\protect\citeauthoryear{Liu, Ma, Yang, Liu, Ma, and Ren}{Liu
  et~al\mbox{.}}{2021a}]%
        {liu2021revfrf}
\bibfield{author}{\bibinfo{person}{Yang Liu}, \bibinfo{person}{Zhuo Ma},
  \bibinfo{person}{Yilong Yang}, \bibinfo{person}{Ximeng Liu},
  \bibinfo{person}{Jianfeng Ma}, {and} \bibinfo{person}{Kui Ren}.}
  \bibinfo{year}{2021}\natexlab{a}.
\newblock \showarticletitle{Revfrf: Enabling cross-domain random forest
  training with revocable federated learning}.
\newblock \bibinfo{journal}{\emph{TDSC}} (\bibinfo{year}{2021}).
\newblock


\bibitem[\protect\citeauthoryear{Liu, Xu, Yuan, Wang, and Li}{Liu
  et~al\mbox{.}}{2022d}]%
        {liu2022right}
\bibfield{author}{\bibinfo{person}{Yi Liu}, \bibinfo{person}{Lei Xu},
  \bibinfo{person}{Xingliang Yuan}, \bibinfo{person}{Cong Wang}, {and}
  \bibinfo{person}{Bo Li}.} \bibinfo{year}{2022}\natexlab{d}.
\newblock \showarticletitle{The Right to be Forgotten in Federated Learning: An
  Efficient Realization with Rapid Retraining}. In
  \bibinfo{booktitle}{\emph{INFOCOM}}. \bibinfo{pages}{1749--1758}.
\newblock


\bibitem[\protect\citeauthoryear{Liu, Jiang, Shen, Peng, Lam, Yuan, and
  Liu}{Liu et~al\mbox{.}}{2023}]%
        {liu2023survey}
\bibfield{author}{\bibinfo{person}{Ziyao Liu}, \bibinfo{person}{Yu Jiang},
  \bibinfo{person}{Jiyuan Shen}, \bibinfo{person}{Minyi Peng},
  \bibinfo{person}{Kwok-Yan Lam}, \bibinfo{person}{Xingliang Yuan}, {and}
  \bibinfo{person}{Xiaoning Liu}.} \bibinfo{year}{2023}\natexlab{}.
\newblock \showarticletitle{A survey on federated unlearning: Challenges,
  methods, and future directions}.
\newblock \bibinfo{journal}{\emph{CSUR}} (\bibinfo{year}{2023}).
\newblock


\bibitem[\protect\citeauthoryear{Lotfi, Mirzarezaee, Hosseinzadeh, and
  Seydi}{Lotfi et~al\mbox{.}}{2021}]%
        {lotfi2021detection}
\bibfield{author}{\bibinfo{person}{Serveh Lotfi}, \bibinfo{person}{Mitra
  Mirzarezaee}, \bibinfo{person}{Mehdi Hosseinzadeh}, {and}
  \bibinfo{person}{Vahid Seydi}.} \bibinfo{year}{2021}\natexlab{}.
\newblock \showarticletitle{Detection of rumor conversations in Twitter using
  graph convolutional networks}.
\newblock \bibinfo{journal}{\emph{Applied Intelligence}}  \bibinfo{volume}{51}
  (\bibinfo{year}{2021}), \bibinfo{pages}{4774--4787}.
\newblock


\bibitem[\protect\citeauthoryear{Lu, Welleck, Hessel, Jiang, Qin, West,
  Ammanabrolu, and Choi}{Lu et~al\mbox{.}}{2022}]%
        {lu2022quark}
\bibfield{author}{\bibinfo{person}{Ximing Lu}, \bibinfo{person}{Sean Welleck},
  \bibinfo{person}{Jack Hessel}, \bibinfo{person}{Liwei Jiang},
  \bibinfo{person}{Lianhui Qin}, \bibinfo{person}{Peter West},
  \bibinfo{person}{Prithviraj Ammanabrolu}, {and} \bibinfo{person}{Yejin
  Choi}.} \bibinfo{year}{2022}\natexlab{}.
\newblock \showarticletitle{Quark: Controllable text generation with reinforced
  unlearning}.
\newblock \bibinfo{journal}{\emph{NeurIPS}}  \bibinfo{volume}{35}
  (\bibinfo{year}{2022}), \bibinfo{pages}{27591--27609}.
\newblock


\bibitem[\protect\citeauthoryear{Mahadevan and Mathioudakis}{Mahadevan and
  Mathioudakis}{2021}]%
        {mahadevan2021certifiable}
\bibfield{author}{\bibinfo{person}{Ananth Mahadevan} {and}
  \bibinfo{person}{Michael Mathioudakis}.} \bibinfo{year}{2021}\natexlab{}.
\newblock \showarticletitle{Certifiable machine unlearning for linear models}.
\newblock \bibinfo{journal}{\emph{arXiv preprint arXiv:2106.15093}}
  (\bibinfo{year}{2021}).
\newblock


\bibitem[\protect\citeauthoryear{Mahadevan and Mathioudakis}{Mahadevan and
  Mathioudakis}{2022}]%
        {mahadevan2022certifiable}
\bibfield{author}{\bibinfo{person}{Ananth Mahadevan} {and}
  \bibinfo{person}{Michael Mathioudakis}.} \bibinfo{year}{2022}\natexlab{}.
\newblock \showarticletitle{Certifiable Unlearning Pipelines for Logistic
  Regression: An Experimental Study}.
\newblock \bibinfo{journal}{\emph{Machine Learning and Knowledge Extraction}}
  \bibinfo{volume}{4}, \bibinfo{number}{3} (\bibinfo{year}{2022}),
  \bibinfo{pages}{591--620}.
\newblock


\bibitem[\protect\citeauthoryear{Mantelero}{Mantelero}{2013}]%
        {magdziarczyk2019right}
\bibfield{author}{\bibinfo{person}{Alessandro Mantelero}.}
  \bibinfo{year}{2013}\natexlab{}.
\newblock \showarticletitle{The EU Proposal for a General Data Protection
  Regulation and the roots of the `right to be forgotten'}.
\newblock \bibinfo{journal}{\emph{Computer Law \& Security Review}}
  \bibinfo{volume}{29}, \bibinfo{number}{3} (\bibinfo{year}{2013}),
  \bibinfo{pages}{229--235}.
\newblock


\bibitem[\protect\citeauthoryear{Marchant, Rubinstein, and Alfeld}{Marchant
  et~al\mbox{.}}{2022}]%
        {marchant2022hard}
\bibfield{author}{\bibinfo{person}{Neil~G Marchant},
  \bibinfo{person}{Benjamin~IP Rubinstein}, {and} \bibinfo{person}{Scott
  Alfeld}.} \bibinfo{year}{2022}\natexlab{}.
\newblock \showarticletitle{Hard to Forget: Poisoning Attacks on Certified
  Machine Unlearning}. In \bibinfo{booktitle}{\emph{AAAI}},
  Vol.~\bibinfo{volume}{36}. \bibinfo{pages}{7691--7700}.
\newblock


\bibitem[\protect\citeauthoryear{Martens}{Martens}{2020}]%
        {martens2020new}
\bibfield{author}{\bibinfo{person}{James Martens}.}
  \bibinfo{year}{2020}\natexlab{}.
\newblock \showarticletitle{New insights and perspectives on the natural
  gradient method}.
\newblock \bibinfo{journal}{\emph{JMLR}} \bibinfo{volume}{21},
  \bibinfo{number}{1} (\bibinfo{year}{2020}), \bibinfo{pages}{5776--5851}.
\newblock


\bibitem[\protect\citeauthoryear{Masi, Wu, Hassner, and Natarajan}{Masi
  et~al\mbox{.}}{2018}]%
        {masi2018deep}
\bibfield{author}{\bibinfo{person}{Iacopo Masi}, \bibinfo{person}{Yue Wu},
  \bibinfo{person}{Tal Hassner}, {and} \bibinfo{person}{Prem Natarajan}.}
  \bibinfo{year}{2018}\natexlab{}.
\newblock \showarticletitle{Deep face recognition: A survey}. In
  \bibinfo{booktitle}{\emph{SIBGRAPI}}. \bibinfo{pages}{471--478}.
\newblock


\bibitem[\protect\citeauthoryear{McMahan, Moore, Ramage, Hampson, and
  y~Arcas}{McMahan et~al\mbox{.}}{2017}]%
        {mcmahan2017communication}
\bibfield{author}{\bibinfo{person}{Brendan McMahan}, \bibinfo{person}{Eider
  Moore}, \bibinfo{person}{Daniel Ramage}, \bibinfo{person}{Seth Hampson},
  {and} \bibinfo{person}{Blaise~Aguera y Arcas}.}
  \bibinfo{year}{2017}\natexlab{}.
\newblock \showarticletitle{Communication-efficient learning of deep networks
  from decentralized data}. In \bibinfo{booktitle}{\emph{AISTAT}}.
  \bibinfo{pages}{1273--1282}.
\newblock


\bibitem[\protect\citeauthoryear{Mehrabi, Morstatter, Saxena, Lerman, and
  Galstyan}{Mehrabi et~al\mbox{.}}{2021}]%
        {mehrabi2021survey}
\bibfield{author}{\bibinfo{person}{Ninareh Mehrabi}, \bibinfo{person}{Fred
  Morstatter}, \bibinfo{person}{Nripsuta Saxena}, \bibinfo{person}{Kristina
  Lerman}, {and} \bibinfo{person}{Aram Galstyan}.}
  \bibinfo{year}{2021}\natexlab{}.
\newblock \showarticletitle{A survey on bias and fairness in machine learning}.
\newblock \bibinfo{journal}{\emph{CSUR}} \bibinfo{volume}{54},
  \bibinfo{number}{6} (\bibinfo{year}{2021}), \bibinfo{pages}{1--35}.
\newblock


\bibitem[\protect\citeauthoryear{Mehta, Pal, Singh, and Ravi}{Mehta
  et~al\mbox{.}}{2022}]%
        {mehta2022deep}
\bibfield{author}{\bibinfo{person}{Ronak Mehta}, \bibinfo{person}{Sourav Pal},
  \bibinfo{person}{Vikas Singh}, {and} \bibinfo{person}{Sathya~N Ravi}.}
  \bibinfo{year}{2022}\natexlab{}.
\newblock \showarticletitle{Deep Unlearning via Randomized Conditionally
  Independent Hessians}. In \bibinfo{booktitle}{\emph{CVPR}}.
  \bibinfo{pages}{10422--10431}.
\newblock


\bibitem[\protect\citeauthoryear{Mercuri, Khraishi, Okhrati, Batra, Hamill,
  Ghasempour, and Nowlan}{Mercuri et~al\mbox{.}}{2022a}]%
        {mercuri2022unlearning}
\bibfield{author}{\bibinfo{person}{Salvatore Mercuri}, \bibinfo{person}{Raad
  Khraishi}, \bibinfo{person}{Ramin Okhrati}, \bibinfo{person}{Devesh Batra},
  \bibinfo{person}{Conor Hamill}, \bibinfo{person}{Taha Ghasempour}, {and}
  \bibinfo{person}{Andrew Nowlan}.} \bibinfo{year}{2022}\natexlab{a}.
\newblock \showarticletitle{An Introduction to Machine Unlearning}.
\newblock \bibinfo{journal}{\emph{arXiv preprint arXiv:2209.00939}}
  (\bibinfo{year}{2022}).
\newblock


\bibitem[\protect\citeauthoryear{Mercuri, Khraishi, Okhrati, Batra, Hamill,
  Ghasempour, and Nowlan}{Mercuri et~al\mbox{.}}{2022b}]%
        {mercuri2022introduction}
\bibfield{author}{\bibinfo{person}{Salvatore Mercuri}, \bibinfo{person}{Raad
  Khraishi}, \bibinfo{person}{Ramin Okhrati}, \bibinfo{person}{Devesh Batra},
  \bibinfo{person}{Conor Hamill}, \bibinfo{person}{Taha Ghasempour}, {and}
  \bibinfo{person}{Andrew Nowlan}.} \bibinfo{year}{2022}\natexlab{b}.
\newblock \showarticletitle{An introduction to machine unlearning}.
\newblock \bibinfo{journal}{\emph{arXiv preprint arXiv:2209.00939}}
  (\bibinfo{year}{2022}).
\newblock


\bibitem[\protect\citeauthoryear{Micaelli et~al\mbox{.}}{Micaelli
  et~al\mbox{.}}{2019}]%
        {micaelli2019zero}
\bibfield{author}{\bibinfo{person}{Paul Micaelli} {et~al\mbox{.}}}
  \bibinfo{year}{2019}\natexlab{}.
\newblock \showarticletitle{Zero-shot knowledge transfer via adversarial belief
  matching}.
\newblock \bibinfo{journal}{\emph{NIPS}}  \bibinfo{volume}{32}
  (\bibinfo{year}{2019}).
\newblock


\bibitem[\protect\citeauthoryear{Nam, Cha, Ahn, Lee, and Shin}{Nam
  et~al\mbox{.}}{2020}]%
        {nam2020learning}
\bibfield{author}{\bibinfo{person}{Junhyun Nam}, \bibinfo{person}{Hyuntak Cha},
  \bibinfo{person}{Sungsoo Ahn}, \bibinfo{person}{Jaeho Lee}, {and}
  \bibinfo{person}{Jinwoo Shin}.} \bibinfo{year}{2020}\natexlab{}.
\newblock \showarticletitle{Learning from failure: De-biasing classifier from
  biased classifier}.
\newblock \bibinfo{journal}{\emph{NIPS}}  \bibinfo{volume}{33}
  (\bibinfo{year}{2020}), \bibinfo{pages}{20673--20684}.
\newblock


\bibitem[\protect\citeauthoryear{Neel, Roth, and Sharifi-Malvajerdi}{Neel
  et~al\mbox{.}}{2021}]%
        {neel2021descent}
\bibfield{author}{\bibinfo{person}{Seth Neel}, \bibinfo{person}{Aaron Roth},
  {and} \bibinfo{person}{Saeed Sharifi-Malvajerdi}.}
  \bibinfo{year}{2021}\natexlab{}.
\newblock \showarticletitle{Descent-to-delete: Gradient-based methods for
  machine unlearning}. In \bibinfo{booktitle}{\emph{Algorithmic Learning
  Theory}}. \bibinfo{pages}{931--962}.
\newblock


\bibitem[\protect\citeauthoryear{Nguyen, Low, and Jaillet}{Nguyen
  et~al\mbox{.}}{2020}]%
        {nguyen2020variational}
\bibfield{author}{\bibinfo{person}{Quoc~Phong Nguyen}, \bibinfo{person}{Bryan
  Kian~Hsiang Low}, {and} \bibinfo{person}{Patrick Jaillet}.}
  \bibinfo{year}{2020}\natexlab{}.
\newblock \showarticletitle{Variational bayesian unlearning}.
\newblock \bibinfo{journal}{\emph{NIPS}}  \bibinfo{volume}{33}
  (\bibinfo{year}{2020}), \bibinfo{pages}{16025--16036}.
\newblock


\bibitem[\protect\citeauthoryear{Nguyen, Oikawa, Divakaran, Chan,
  et~al\mbox{.}}{Nguyen et~al\mbox{.}}{2022b}]%
        {nguyen2022markov}
\bibfield{author}{\bibinfo{person}{Quoc~Phong Nguyen}, \bibinfo{person}{Ryutaro
  Oikawa}, \bibinfo{person}{Dinil~Mon Divakaran}, \bibinfo{person}{Mun~Choon
  Chan}, {et~al\mbox{.}}} \bibinfo{year}{2022}\natexlab{b}.
\newblock \showarticletitle{Markov Chain Monte Carlo-Based Machine Unlearning:
  Unlearning What Needs to Be Forgotten}. In
  \bibinfo{booktitle}{\emph{ASIACCS}}. \bibinfo{pages}{351--363}.
\newblock


\bibitem[\protect\citeauthoryear{Nguyen}{Nguyen}{2019}]%
        {nguyen2019debunking}
\bibfield{author}{\bibinfo{person}{Thanh~Tam Nguyen}.}
  \bibinfo{year}{2019}\natexlab{}.
\newblock \emph{\bibinfo{title}{Debunking Misinformation on the Web: Detection,
  Validation, and Visualisation}}.
\newblock \bibinfo{thesistype}{Ph.D. Dissertation}. \bibinfo{school}{EPFL,
  Switzerland}.
\newblock


\bibitem[\protect\citeauthoryear{Nguyen, Huynh, Nguyen, Liew, Yin, and
  Nguyen}{Nguyen et~al\mbox{.}}{2022a}]%
        {nguyen2022survey}
\bibfield{author}{\bibinfo{person}{Thanh~Tam Nguyen},
  \bibinfo{person}{Thanh~Trung Huynh}, \bibinfo{person}{Phi~Le Nguyen},
  \bibinfo{person}{Alan Wee-Chung Liew}, \bibinfo{person}{Hongzhi Yin}, {and}
  \bibinfo{person}{Quoc Viet~Hung Nguyen}.} \bibinfo{year}{2022}\natexlab{a}.
\newblock \showarticletitle{A Survey of Machine Unlearning}.
\newblock \bibinfo{journal}{\emph{arXiv preprint arXiv:2209.02299}}
  (\bibinfo{year}{2022}).
\newblock


\bibitem[\protect\citeauthoryear{Nguyen, Nguyen, Nguyen, Vo, Jo, and
  Nguyen}{Nguyen et~al\mbox{.}}{2021}]%
        {nguyen2021judo}
\bibfield{author}{\bibinfo{person}{Thanh~Toan Nguyen},
  \bibinfo{person}{Thanh~Tam Nguyen}, \bibinfo{person}{Thanh~Thi Nguyen},
  \bibinfo{person}{Bay Vo}, \bibinfo{person}{Jun Jo}, {and}
  \bibinfo{person}{Quoc Viet~Hung Nguyen}.} \bibinfo{year}{2021}\natexlab{}.
\newblock \showarticletitle{Judo: Just-in-time rumour detection in streaming
  social platforms}.
\newblock \bibinfo{journal}{\emph{Information Sciences}}  \bibinfo{volume}{570}
  (\bibinfo{year}{2021}), \bibinfo{pages}{70--93}.
\newblock


\bibitem[\protect\citeauthoryear{Pan, Chien, and Milenkovic}{Pan
  et~al\mbox{.}}{2023}]%
        {pan2023unlearning}
\bibfield{author}{\bibinfo{person}{Chao Pan}, \bibinfo{person}{Eli Chien},
  {and} \bibinfo{person}{Olgica Milenkovic}.} \bibinfo{year}{2023}\natexlab{}.
\newblock \showarticletitle{Unlearning graph classifiers with limited data
  resources}. In \bibinfo{booktitle}{\emph{WWW}}. \bibinfo{pages}{716--726}.
\newblock


\bibitem[\protect\citeauthoryear{Panda and AP}{Panda and AP}{2023}]%
        {panda2023fast}
\bibfield{author}{\bibinfo{person}{Subhodip Panda} {and}
  \bibinfo{person}{Prathosh AP}.} \bibinfo{year}{2023}\natexlab{}.
\newblock \showarticletitle{FAST: Feature Aware Similarity Thresholding for
  Weak Unlearning in Black-Box Generative Models}.
\newblock \bibinfo{journal}{\emph{arXiv preprint arXiv:2312.14895}}
  (\bibinfo{year}{2023}).
\newblock


\bibitem[\protect\citeauthoryear{Pardau}{Pardau}{2018}]%
        {pardau2018california}
\bibfield{author}{\bibinfo{person}{Stuart~L Pardau}.}
  \bibinfo{year}{2018}\natexlab{}.
\newblock \showarticletitle{The California consumer privacy act: Towards a
  European-style privacy regime in the United States}.
\newblock \bibinfo{journal}{\emph{J. Tech. L. \& Pol'y}}  \bibinfo{volume}{23}
  (\bibinfo{year}{2018}), \bibinfo{pages}{68}.
\newblock


\bibitem[\protect\citeauthoryear{Parisi, Kemker, Part, Kanan, and
  Wermter}{Parisi et~al\mbox{.}}{2019}]%
        {parisi2019continual}
\bibfield{author}{\bibinfo{person}{German~I Parisi}, \bibinfo{person}{Ronald
  Kemker}, \bibinfo{person}{Jose~L Part}, \bibinfo{person}{Christopher Kanan},
  {and} \bibinfo{person}{Stefan Wermter}.} \bibinfo{year}{2019}\natexlab{}.
\newblock \showarticletitle{Continual lifelong learning with neural networks: A
  review}.
\newblock \bibinfo{journal}{\emph{Neural Networks}}  \bibinfo{volume}{113}
  (\bibinfo{year}{2019}), \bibinfo{pages}{54--71}.
\newblock


\bibitem[\protect\citeauthoryear{Parne, Puppaala, Bhupathi, and Patgiri}{Parne
  et~al\mbox{.}}{2021}]%
        {parne2021machine}
\bibfield{author}{\bibinfo{person}{Nishchal Parne}, \bibinfo{person}{Kyathi
  Puppaala}, \bibinfo{person}{Nithish Bhupathi}, {and} \bibinfo{person}{Ripon
  Patgiri}.} \bibinfo{year}{2021}\natexlab{}.
\newblock \showarticletitle{An Investigation on Learning, Polluting, and
  Unlearning the Spam Emails for Lifelong Learning}.
\newblock \bibinfo{journal}{\emph{arXiv preprint arXiv:2111.14609}}
  (\bibinfo{year}{2021}).
\newblock


\bibitem[\protect\citeauthoryear{Pearce, Leibfried, and Brintrup}{Pearce
  et~al\mbox{.}}{2020}]%
        {pearce2020uncertainty}
\bibfield{author}{\bibinfo{person}{Tim Pearce}, \bibinfo{person}{Felix
  Leibfried}, {and} \bibinfo{person}{Alexandra Brintrup}.}
  \bibinfo{year}{2020}\natexlab{}.
\newblock \showarticletitle{Uncertainty in neural networks: Approximately
  bayesian ensembling}. In \bibinfo{booktitle}{\emph{AISTATS}}.
  \bibinfo{pages}{234--244}.
\newblock


\bibitem[\protect\citeauthoryear{Peste, Alistarh, and Lampert}{Peste
  et~al\mbox{.}}{2021}]%
        {peste2021ssse}
\bibfield{author}{\bibinfo{person}{Alexandra Peste}, \bibinfo{person}{Dan
  Alistarh}, {and} \bibinfo{person}{Christoph~H Lampert}.}
  \bibinfo{year}{2021}\natexlab{}.
\newblock \showarticletitle{{SSSE}: Efficiently Erasing Samples from Trained
  Machine Learning Models}. In \bibinfo{booktitle}{\emph{NeurIPS 2021 Workshop
  Privacy in Machine Learning}}.
\newblock


\bibitem[\protect\citeauthoryear{Pochinkov and Schoots}{Pochinkov and
  Schoots}{2024}]%
        {pochinkov2024dissecting}
\bibfield{author}{\bibinfo{person}{Nicholas Pochinkov} {and}
  \bibinfo{person}{Nandi Schoots}.} \bibinfo{year}{2024}\natexlab{}.
\newblock \showarticletitle{Dissecting Language Models: Machine Unlearning via
  Selective Pruning}.
\newblock \bibinfo{journal}{\emph{arXiv preprint arXiv:2403.01267}}
  (\bibinfo{year}{2024}).
\newblock


\bibitem[\protect\citeauthoryear{Qiu, Wang, Xu, Cui, and Shen}{Qiu
  et~al\mbox{.}}{2023}]%
        {qiu2023fedcio}
\bibfield{author}{\bibinfo{person}{Hongyu Qiu}, \bibinfo{person}{Yongwei Wang},
  \bibinfo{person}{Yonghui Xu}, \bibinfo{person}{Lizhen Cui}, {and}
  \bibinfo{person}{Zhiqi Shen}.} \bibinfo{year}{2023}\natexlab{}.
\newblock \showarticletitle{FedCIO: Efficient Exact Federated Unlearning with
  Clustering, Isolation, and One-shot Aggregation}. In
  \bibinfo{booktitle}{\emph{BigData}}. \bibinfo{pages}{5559--5568}.
\newblock


\bibitem[\protect\citeauthoryear{Ramaswamy, Kim, and Russakovsky}{Ramaswamy
  et~al\mbox{.}}{2021}]%
        {ramaswamy2021fair}
\bibfield{author}{\bibinfo{person}{Vikram~V Ramaswamy},
  \bibinfo{person}{Sunnie~SY Kim}, {and} \bibinfo{person}{Olga Russakovsky}.}
  \bibinfo{year}{2021}\natexlab{}.
\newblock \showarticletitle{Fair attribute classification through latent space
  de-biasing}. In \bibinfo{booktitle}{\emph{CVPR}}.
  \bibinfo{pages}{9301--9310}.
\newblock


\bibitem[\protect\citeauthoryear{Ren, Zheng, Qin, and Liu}{Ren
  et~al\mbox{.}}{2020c}]%
        {ren2020adversarial}
\bibfield{author}{\bibinfo{person}{Kui Ren}, \bibinfo{person}{Tianhang Zheng},
  \bibinfo{person}{Zhan Qin}, {and} \bibinfo{person}{Xue Liu}.}
  \bibinfo{year}{2020}\natexlab{c}.
\newblock \showarticletitle{Adversarial attacks and defenses in deep learning}.
\newblock \bibinfo{journal}{\emph{Engineering}} \bibinfo{volume}{6},
  \bibinfo{number}{3} (\bibinfo{year}{2020}), \bibinfo{pages}{346--360}.
\newblock


\bibitem[\protect\citeauthoryear{Ren, Baird, Han, Zhang, and Schuller}{Ren
  et~al\mbox{.}}{2020a}]%
        {ren2020generating}
\bibfield{author}{\bibinfo{person}{Zhao Ren}, \bibinfo{person}{Alice Baird},
  \bibinfo{person}{Jing Han}, \bibinfo{person}{Zixing Zhang}, {and}
  \bibinfo{person}{Bj{\"o}rn Schuller}.} \bibinfo{year}{2020}\natexlab{a}.
\newblock \showarticletitle{Generating and protecting against adversarial
  attacks for deep speech-based emotion recognition models}. In
  \bibinfo{booktitle}{\emph{ICASSP}}. \bibinfo{pages}{7184--7188}.
\newblock


\bibitem[\protect\citeauthoryear{Ren, Han, Cummins, and Schuller}{Ren
  et~al\mbox{.}}{2020b}]%
        {ren2020enhancing}
\bibfield{author}{\bibinfo{person}{Zhao Ren}, \bibinfo{person}{Jing Han},
  \bibinfo{person}{Nicholas Cummins}, {and} \bibinfo{person}{Bj{\"o}rn~W
  Schuller}.} \bibinfo{year}{2020}\natexlab{b}.
\newblock \showarticletitle{Enhancing Transferability of Black-Box Adversarial
  Attacks via Lifelong Learning for Speech Emotion Recognition Models.}. In
  \bibinfo{booktitle}{\emph{INTERSPEECH}}. \bibinfo{pages}{496--500}.
\newblock


\bibitem[\protect\citeauthoryear{Ren, Nguyen, and Nejdl}{Ren
  et~al\mbox{.}}{2022}]%
        {ren2022prototype}
\bibfield{author}{\bibinfo{person}{Zhao Ren}, \bibinfo{person}{Thanh~Tam
  Nguyen}, {and} \bibinfo{person}{Wolfgang Nejdl}.}
  \bibinfo{year}{2022}\natexlab{}.
\newblock \showarticletitle{Prototype learning for interpretable respiratory
  sound analysis}. In \bibinfo{booktitle}{\emph{ICASSP}}.
  \bibinfo{pages}{9087--9091}.
\newblock


\bibitem[\protect\citeauthoryear{Romero, Barrio, and Belanche}{Romero
  et~al\mbox{.}}{2007}]%
        {romero2007incremental}
\bibfield{author}{\bibinfo{person}{Enrique Romero}, \bibinfo{person}{Ignacio
  Barrio}, {and} \bibinfo{person}{Llu{\'\i}s Belanche}.}
  \bibinfo{year}{2007}\natexlab{}.
\newblock \showarticletitle{Incremental and decremental learning for linear
  support vector machines}. In \bibinfo{booktitle}{\emph{ICANN}}.
  \bibinfo{pages}{209--218}.
\newblock


\bibitem[\protect\citeauthoryear{Roth and Pernkopf}{Roth and Pernkopf}{2018}]%
        {roth2018bayesian}
\bibfield{author}{\bibinfo{person}{Wolfgang Roth} {and} \bibinfo{person}{Franz
  Pernkopf}.} \bibinfo{year}{2018}\natexlab{}.
\newblock \showarticletitle{Bayesian neural networks with weight sharing using
  Dirichlet processes}.
\newblock \bibinfo{journal}{\emph{TPAMI}} \bibinfo{volume}{42},
  \bibinfo{number}{1} (\bibinfo{year}{2018}), \bibinfo{pages}{246--252}.
\newblock


\bibitem[\protect\citeauthoryear{Salem, Bhattacharya, Backes, Fritz, and
  Zhang}{Salem et~al\mbox{.}}{2020}]%
        {salem2020updates}
\bibfield{author}{\bibinfo{person}{Ahmed Salem}, \bibinfo{person}{Apratim
  Bhattacharya}, \bibinfo{person}{Michael Backes}, \bibinfo{person}{Mario
  Fritz}, {and} \bibinfo{person}{Yang Zhang}.} \bibinfo{year}{2020}\natexlab{}.
\newblock \showarticletitle{$\{$Updates-Leak$\}$: Data Set Inference and
  Reconstruction Attacks in Online Learning}. In
  \bibinfo{booktitle}{\emph{USENIX Security}}. \bibinfo{pages}{1291--1308}.
\newblock


\bibitem[\protect\citeauthoryear{Salem, Zhang, Humbert, Berrang, Fritz, and
  Backes}{Salem et~al\mbox{.}}{2019}]%
        {Salem0HBF019}
\bibfield{author}{\bibinfo{person}{Ahmed Salem}, \bibinfo{person}{Yang Zhang},
  \bibinfo{person}{Mathias Humbert}, \bibinfo{person}{Pascal Berrang},
  \bibinfo{person}{Mario Fritz}, {and} \bibinfo{person}{Michael Backes}.}
  \bibinfo{year}{2019}\natexlab{}.
\newblock \showarticletitle{ML-Leaks: Model and Data Independent Membership
  Inference Attacks and Defenses on Machine Learning Models}. In
  \bibinfo{booktitle}{\emph{NDSS}}.
\newblock


\bibitem[\protect\citeauthoryear{Sari, Samosir, Sahara, Agustina, and
  Anita}{Sari et~al\mbox{.}}{2020}]%
        {sari2020learning}
\bibfield{author}{\bibinfo{person}{WN Sari}, \bibinfo{person}{BS Samosir},
  \bibinfo{person}{N Sahara}, \bibinfo{person}{L Agustina}, {and}
  \bibinfo{person}{Y Anita}.} \bibinfo{year}{2020}\natexlab{}.
\newblock \showarticletitle{Learning mathematics “Asyik” with Youtube
  educative media}. In \bibinfo{booktitle}{\emph{Journal of Physics: Conference
  Series}}, Vol.~\bibinfo{volume}{1477}. \bibinfo{pages}{022012}.
\newblock


\bibitem[\protect\citeauthoryear{Sattler, Korjakow, Rischke, and Samek}{Sattler
  et~al\mbox{.}}{2021}]%
        {sattler2021fedaux}
\bibfield{author}{\bibinfo{person}{Felix Sattler}, \bibinfo{person}{Tim
  Korjakow}, \bibinfo{person}{Roman Rischke}, {and} \bibinfo{person}{Wojciech
  Samek}.} \bibinfo{year}{2021}\natexlab{}.
\newblock \showarticletitle{Fedaux: Leveraging unlabeled auxiliary data in
  federated learning}.
\newblock \bibinfo{journal}{\emph{TNNLS}} (\bibinfo{year}{2021}).
\newblock


\bibitem[\protect\citeauthoryear{Schelter}{Schelter}{2020}]%
        {schelter2020amnesia}
\bibfield{author}{\bibinfo{person}{Sebastian Schelter}.}
  \bibinfo{year}{2020}\natexlab{}.
\newblock \showarticletitle{``Amnesia'' - A Selection of Machine Learning
  Models That Can Forget User Data Very Fast}. In
  \bibinfo{booktitle}{\emph{CIDR}}.
\newblock


\bibitem[\protect\citeauthoryear{Schelter, Grafberger, and Dunning}{Schelter
  et~al\mbox{.}}{2021}]%
        {schelter2021hedgecut}
\bibfield{author}{\bibinfo{person}{Sebastian Schelter}, \bibinfo{person}{Stefan
  Grafberger}, {and} \bibinfo{person}{Ted Dunning}.}
  \bibinfo{year}{2021}\natexlab{}.
\newblock \showarticletitle{Hedgecut: Maintaining randomised trees for
  low-latency machine unlearning}. In \bibinfo{booktitle}{\emph{SIGMOD}}.
  \bibinfo{pages}{1545--1557}.
\newblock


\bibitem[\protect\citeauthoryear{Sekhari, Acharya, Kamath, and Suresh}{Sekhari
  et~al\mbox{.}}{2021}]%
        {sekhari2021remember}
\bibfield{author}{\bibinfo{person}{Ayush Sekhari}, \bibinfo{person}{Jayadev
  Acharya}, \bibinfo{person}{Gautam Kamath}, {and}
  \bibinfo{person}{Ananda~Theertha Suresh}.} \bibinfo{year}{2021}\natexlab{}.
\newblock \showarticletitle{Remember what you want to forget: Algorithms for
  machine unlearning}.
\newblock \bibinfo{journal}{\emph{NIPS}}  \bibinfo{volume}{34}
  (\bibinfo{year}{2021}), \bibinfo{pages}{18075--18086}.
\newblock


\bibitem[\protect\citeauthoryear{Shaik, Tao, Li, Xie, Cai, Zhu, and Li}{Shaik
  et~al\mbox{.}}{2024}]%
        {shaik2024framu}
\bibfield{author}{\bibinfo{person}{Thanveer Shaik}, \bibinfo{person}{Xiaohui
  Tao}, \bibinfo{person}{Lin Li}, \bibinfo{person}{Haoran Xie},
  \bibinfo{person}{Taotao Cai}, \bibinfo{person}{Xiaofeng Zhu}, {and}
  \bibinfo{person}{Qing Li}.} \bibinfo{year}{2024}\natexlab{}.
\newblock \showarticletitle{Framu: Attention-based machine unlearning using
  federated reinforcement learning}.
\newblock \bibinfo{journal}{\emph{TKDE}} (\bibinfo{year}{2024}).
\newblock


\bibitem[\protect\citeauthoryear{Shaik, Tao, Xie, Li, Zhu, and Li}{Shaik
  et~al\mbox{.}}{2023}]%
        {shaik2023exploring}
\bibfield{author}{\bibinfo{person}{Thanveer Shaik}, \bibinfo{person}{Xiaohui
  Tao}, \bibinfo{person}{Haoran Xie}, \bibinfo{person}{Lin Li},
  \bibinfo{person}{Xiaofeng Zhu}, {and} \bibinfo{person}{Qing Li}.}
  \bibinfo{year}{2023}\natexlab{}.
\newblock \showarticletitle{Exploring the landscape of machine unlearning: A
  comprehensive survey and taxonomy}.
\newblock \bibinfo{journal}{\emph{arXiv preprint arXiv:2305.06360}}
  (\bibinfo{year}{2023}).
\newblock


\bibitem[\protect\citeauthoryear{Shan, Wenger, Zhang, Li, Zheng, and Zhao}{Shan
  et~al\mbox{.}}{2020}]%
        {shan2020protecting}
\bibfield{author}{\bibinfo{person}{S Shan}, \bibinfo{person}{E Wenger},
  \bibinfo{person}{J Zhang}, \bibinfo{person}{H Li}, \bibinfo{person}{H Zheng},
  {and} \bibinfo{person}{BY Zhao}.} \bibinfo{year}{2020}\natexlab{}.
\newblock \showarticletitle{Protecting Personal Privacy against Una uthorized
  Deep Learning Models}. In \bibinfo{booktitle}{\emph{USENIX Security}}.
  \bibinfo{pages}{1--16}.
\newblock


\bibitem[\protect\citeauthoryear{Shao, Lin, Cao, and Luo}{Shao
  et~al\mbox{.}}{2024}]%
        {shao2024federated}
\bibfield{author}{\bibinfo{person}{Jiaqi Shao}, \bibinfo{person}{Tao Lin},
  \bibinfo{person}{Xuanyu Cao}, {and} \bibinfo{person}{Bing Luo}.}
  \bibinfo{year}{2024}\natexlab{}.
\newblock \showarticletitle{Federated Unlearning: a Perspective of Stability
  and Fairness}.
\newblock \bibinfo{journal}{\emph{arXiv preprint arXiv:2402.01276}}
  (\bibinfo{year}{2024}).
\newblock


\bibitem[\protect\citeauthoryear{Shibata, Irie, Ikami, and Mitsuzumi}{Shibata
  et~al\mbox{.}}{2021}]%
        {shibata2021learning}
\bibfield{author}{\bibinfo{person}{Takashi Shibata}, \bibinfo{person}{Go Irie},
  \bibinfo{person}{Daiki Ikami}, {and} \bibinfo{person}{Yu Mitsuzumi}.}
  \bibinfo{year}{2021}\natexlab{}.
\newblock \showarticletitle{Learning with Selective Forgetting.}. In
  \bibinfo{booktitle}{\emph{IJCAI}}, Vol.~\bibinfo{volume}{2}.
  \bibinfo{pages}{6}.
\newblock


\bibitem[\protect\citeauthoryear{Shintre et~al\mbox{.}}{Shintre
  et~al\mbox{.}}{2019}]%
        {shintre2019making}
\bibfield{author}{\bibinfo{person}{Saurabh Shintre} {et~al\mbox{.}}}
  \bibinfo{year}{2019}\natexlab{}.
\newblock \showarticletitle{Making machine learning forget}. In
  \bibinfo{booktitle}{\emph{Annual Privacy Forum}}. \bibinfo{pages}{72--83}.
\newblock


\bibitem[\protect\citeauthoryear{Shokri, Stronati, Song, and Shmatikov}{Shokri
  et~al\mbox{.}}{2017}]%
        {shokri2017membership}
\bibfield{author}{\bibinfo{person}{Reza Shokri}, \bibinfo{person}{Marco
  Stronati}, \bibinfo{person}{Congzheng Song}, {and} \bibinfo{person}{Vitaly
  Shmatikov}.} \bibinfo{year}{2017}\natexlab{}.
\newblock \showarticletitle{Membership inference attacks against machine
  learning models}. In \bibinfo{booktitle}{\emph{SP}}. \bibinfo{pages}{3--18}.
\newblock


\bibitem[\protect\citeauthoryear{Shwartz-Ziv and Tishby}{Shwartz-Ziv and
  Tishby}{2017}]%
        {shwartz2017opening}
\bibfield{author}{\bibinfo{person}{Ravid Shwartz-Ziv} {and}
  \bibinfo{person}{Naftali Tishby}.} \bibinfo{year}{2017}\natexlab{}.
\newblock \showarticletitle{Opening the black box of deep neural networks via
  information}.
\newblock \bibinfo{journal}{\emph{arXiv preprint arXiv:1703.00810}}
  (\bibinfo{year}{2017}).
\newblock


\bibitem[\protect\citeauthoryear{Singh and Anand}{Singh and Anand}{2017}]%
        {singh2017data}
\bibfield{author}{\bibinfo{person}{Abhijeet Singh} {and}
  \bibinfo{person}{Abhineet Anand}.} \bibinfo{year}{2017}\natexlab{}.
\newblock \showarticletitle{Data leakage detection using cloud computing}.
\newblock \bibinfo{journal}{\emph{IJECS}} \bibinfo{volume}{6},
  \bibinfo{number}{4} (\bibinfo{year}{2017}).
\newblock


\bibitem[\protect\citeauthoryear{Singh, Majumdar, Mittal, and Vatsa}{Singh
  et~al\mbox{.}}{2022}]%
        {singh2022anatomizing}
\bibfield{author}{\bibinfo{person}{Richa Singh}, \bibinfo{person}{Puspita
  Majumdar}, \bibinfo{person}{Surbhi Mittal}, {and} \bibinfo{person}{Mayank
  Vatsa}.} \bibinfo{year}{2022}\natexlab{}.
\newblock \showarticletitle{Anatomizing bias in facial analysis}. In
  \bibinfo{booktitle}{\emph{AAAI}}, Vol.~\bibinfo{volume}{36}.
  \bibinfo{pages}{12351--12358}.
\newblock


\bibitem[\protect\citeauthoryear{Sinha, Mandal, and Kankanhalli}{Sinha
  et~al\mbox{.}}{2024}]%
        {sinha2024multi}
\bibfield{author}{\bibinfo{person}{Yash Sinha}, \bibinfo{person}{Murari
  Mandal}, {and} \bibinfo{person}{Mohan Kankanhalli}.}
  \bibinfo{year}{2024}\natexlab{}.
\newblock \showarticletitle{Multi-Modal Recommendation Unlearning}.
\newblock \bibinfo{journal}{\emph{arXiv preprint arXiv:2405.15328}}
  (\bibinfo{year}{2024}).
\newblock


\bibitem[\protect\citeauthoryear{Sommer, Song, Wagh, and Mittal}{Sommer
  et~al\mbox{.}}{2020}]%
        {sommer2020towards}
\bibfield{author}{\bibinfo{person}{David~Marco Sommer}, \bibinfo{person}{Liwei
  Song}, \bibinfo{person}{Sameer Wagh}, {and} \bibinfo{person}{Prateek
  Mittal}.} \bibinfo{year}{2020}\natexlab{}.
\newblock \showarticletitle{Towards probabilistic verification of machine
  unlearning}.
\newblock \bibinfo{journal}{\emph{arXiv preprint arXiv:2003.04247}}
  (\bibinfo{year}{2020}).
\newblock


\bibitem[\protect\citeauthoryear{Sommer, Song, Wagh, and Mittal}{Sommer
  et~al\mbox{.}}{2022}]%
        {sommer2022athena}
\bibfield{author}{\bibinfo{person}{David~Marco Sommer}, \bibinfo{person}{Liwei
  Song}, \bibinfo{person}{Sameer Wagh}, {and} \bibinfo{person}{Prateek
  Mittal}.} \bibinfo{year}{2022}\natexlab{}.
\newblock \showarticletitle{Athena: Probabilistic Verification of Machine
  Unlearning}.
\newblock \bibinfo{journal}{\emph{Proc. Priv. Enhancing Technol.}}
  \bibinfo{volume}{2022}, \bibinfo{number}{3} (\bibinfo{year}{2022}),
  \bibinfo{pages}{268--290}.
\newblock


\bibitem[\protect\citeauthoryear{Tahiliani, Hassija, Chamola, and
  Guizani}{Tahiliani et~al\mbox{.}}{2021}]%
        {tahiliani2021machine}
\bibfield{author}{\bibinfo{person}{Aman Tahiliani}, \bibinfo{person}{Vikas
  Hassija}, \bibinfo{person}{Vinay Chamola}, {and} \bibinfo{person}{Mohsen
  Guizani}.} \bibinfo{year}{2021}\natexlab{}.
\newblock \showarticletitle{Machine Unlearning: Its Need and Implementation
  Strategies}. In \bibinfo{booktitle}{\emph{IC3}}. \bibinfo{pages}{241--246}.
\newblock


\bibitem[\protect\citeauthoryear{Tam, Weidlich, Thang, Yin, and Hung}{Tam
  et~al\mbox{.}}{2017}]%
        {nguyen2017retaining}
\bibfield{author}{\bibinfo{person}{Nguyen~Thanh Tam}, \bibinfo{person}{Matthias
  Weidlich}, \bibinfo{person}{Duong~Chi Thang}, \bibinfo{person}{Hongzhi Yin},
  {and} \bibinfo{person}{Nguyen Quoc~Viet Hung}.}
  \bibinfo{year}{2017}\natexlab{}.
\newblock \showarticletitle{Retaining Data from Streams of Social Platforms
  with Minimal Regret}. In \bibinfo{booktitle}{\emph{IJCAI}}.
  \bibinfo{pages}{2850--2856}.
\newblock


\bibitem[\protect\citeauthoryear{Tanha, Abdi, Samadi, Razzaghi, and
  Asadpour}{Tanha et~al\mbox{.}}{2020}]%
        {tanha2020boosting}
\bibfield{author}{\bibinfo{person}{Jafar Tanha}, \bibinfo{person}{Yousef Abdi},
  \bibinfo{person}{Negin Samadi}, \bibinfo{person}{Nazila Razzaghi}, {and}
  \bibinfo{person}{Mohammad Asadpour}.} \bibinfo{year}{2020}\natexlab{}.
\newblock \showarticletitle{Boosting methods for multi-class imbalanced data
  classification: an experimental review}.
\newblock \bibinfo{journal}{\emph{Journal of Big Data}} \bibinfo{volume}{7},
  \bibinfo{number}{1} (\bibinfo{year}{2020}), \bibinfo{pages}{1--47}.
\newblock


\bibitem[\protect\citeauthoryear{Tao, Wang, Pan, Yu, Cheng, and Wang}{Tao
  et~al\mbox{.}}{2024}]%
        {tao2024communication}
\bibfield{author}{\bibinfo{person}{Youming Tao}, \bibinfo{person}{Cheng-Long
  Wang}, \bibinfo{person}{Miao Pan}, \bibinfo{person}{Dongxiao Yu},
  \bibinfo{person}{Xiuzhen Cheng}, {and} \bibinfo{person}{Di Wang}.}
  \bibinfo{year}{2024}\natexlab{}.
\newblock \showarticletitle{Communication Efficient and Provable Federated
  Unlearning}.
\newblock \bibinfo{journal}{\emph{PVLDB}} \bibinfo{volume}{17},
  \bibinfo{number}{5} (\bibinfo{year}{2024}), \bibinfo{pages}{1119–1131}.
\newblock


\bibitem[\protect\citeauthoryear{Tarun, Chundawat, Mandal, and
  Kankanhalli}{Tarun et~al\mbox{.}}{2021}]%
        {tarun2021fast}
\bibfield{author}{\bibinfo{person}{Ayush~K Tarun}, \bibinfo{person}{Vikram~S
  Chundawat}, \bibinfo{person}{Murari Mandal}, {and} \bibinfo{person}{Mohan
  Kankanhalli}.} \bibinfo{year}{2021}\natexlab{}.
\newblock \showarticletitle{Fast yet effective machine unlearning}.
\newblock \bibinfo{journal}{\emph{arXiv preprint arXiv:2111.08947}}
  (\bibinfo{year}{2021}).
\newblock


\bibitem[\protect\citeauthoryear{Thudi, Deza, Chandrasekaran, and
  Papernot}{Thudi et~al\mbox{.}}{2022a}]%
        {thudi2022unrolling}
\bibfield{author}{\bibinfo{person}{Anvith Thudi}, \bibinfo{person}{Gabriel
  Deza}, \bibinfo{person}{Varun Chandrasekaran}, {and} \bibinfo{person}{Nicolas
  Papernot}.} \bibinfo{year}{2022}\natexlab{a}.
\newblock \showarticletitle{Unrolling sgd: Understanding factors influencing
  machine unlearning}. In \bibinfo{booktitle}{\emph{EuroS\&P}}.
  \bibinfo{pages}{303--319}.
\newblock


\bibitem[\protect\citeauthoryear{Thudi, Jia, Shumailov, and Papernot}{Thudi
  et~al\mbox{.}}{2022b}]%
        {thudi2021necessity}
\bibfield{author}{\bibinfo{person}{Anvith Thudi}, \bibinfo{person}{Hengrui
  Jia}, \bibinfo{person}{Ilia Shumailov}, {and} \bibinfo{person}{Nicolas
  Papernot}.} \bibinfo{year}{2022}\natexlab{b}.
\newblock \showarticletitle{On the necessity of auditable algorithmic
  definitions for machine unlearning}. In \bibinfo{booktitle}{\emph{USENIX
  Security}}. \bibinfo{pages}{4007--4022}.
\newblock


\bibitem[\protect\citeauthoryear{Thudi, Shumailov, Boenisch, and
  Papernot}{Thudi et~al\mbox{.}}{2022c}]%
        {thudi2022bounding}
\bibfield{author}{\bibinfo{person}{Anvith Thudi}, \bibinfo{person}{Ilia
  Shumailov}, \bibinfo{person}{Franziska Boenisch}, {and}
  \bibinfo{person}{Nicolas Papernot}.} \bibinfo{year}{2022}\natexlab{c}.
\newblock \showarticletitle{Bounding Membership Inference}.
\newblock \bibinfo{journal}{\emph{arXiv preprint arXiv:2202.12232}}
  (\bibinfo{year}{2022}).
\newblock


\bibitem[\protect\citeauthoryear{Tishby et~al\mbox{.}}{Tishby
  et~al\mbox{.}}{2000}]%
        {tishby2000information}
\bibfield{author}{\bibinfo{person}{Naftali Tishby} {et~al\mbox{.}}}
  \bibinfo{year}{2000}\natexlab{}.
\newblock \showarticletitle{The information bottleneck method}.
\newblock \bibinfo{journal}{\emph{arXiv preprint physics/0004057}}
  (\bibinfo{year}{2000}).
\newblock


\bibitem[\protect\citeauthoryear{Tishby and Zaslavsky}{Tishby and
  Zaslavsky}{2015}]%
        {tishby2015deep}
\bibfield{author}{\bibinfo{person}{Naftali Tishby} {and} \bibinfo{person}{Noga
  Zaslavsky}.} \bibinfo{year}{2015}\natexlab{}.
\newblock \showarticletitle{Deep learning and the information bottleneck
  principle}. In \bibinfo{booktitle}{\emph{ITW}}. \bibinfo{pages}{1--5}.
\newblock


\bibitem[\protect\citeauthoryear{Tiwary, Guha, Panda, et~al\mbox{.}}{Tiwary
  et~al\mbox{.}}{2023}]%
        {tiwary2023adapt}
\bibfield{author}{\bibinfo{person}{Piyush Tiwary}, \bibinfo{person}{Atri Guha},
  \bibinfo{person}{Subhodip Panda}, {et~al\mbox{.}}}
  \bibinfo{year}{2023}\natexlab{}.
\newblock \showarticletitle{Adapt then unlearn: Exploiting parameter space
  semantics for unlearning in generative adversarial networks}.
\newblock \bibinfo{journal}{\emph{arXiv preprint arXiv:2309.14054}}
  (\bibinfo{year}{2023}).
\newblock


\bibitem[\protect\citeauthoryear{Tsai, Lin, and Lin}{Tsai
  et~al\mbox{.}}{2014}]%
        {tsai2014incremental}
\bibfield{author}{\bibinfo{person}{Cheng-Hao Tsai}, \bibinfo{person}{Chieh-Yen
  Lin}, {and} \bibinfo{person}{Chih-Jen Lin}.} \bibinfo{year}{2014}\natexlab{}.
\newblock \showarticletitle{Incremental and decremental training for linear
  classification}. In \bibinfo{booktitle}{\emph{KDD}}.
  \bibinfo{pages}{343--352}.
\newblock


\bibitem[\protect\citeauthoryear{Tveit et~al\mbox{.}}{Tveit
  et~al\mbox{.}}{2003a}]%
        {tveit2003multicategory}
\bibfield{author}{\bibinfo{person}{Amund Tveit} {et~al\mbox{.}}}
  \bibinfo{year}{2003}\natexlab{a}.
\newblock \showarticletitle{Multicategory incremental proximal support vector
  classifiers}. In \bibinfo{booktitle}{\emph{KES}}. \bibinfo{pages}{386--392}.
\newblock


\bibitem[\protect\citeauthoryear{Tveit, Hetland, and Engum}{Tveit
  et~al\mbox{.}}{2003b}]%
        {tveit2003incremental}
\bibfield{author}{\bibinfo{person}{Amund Tveit}, \bibinfo{person}{Magnus~Lie
  Hetland}, {and} \bibinfo{person}{H{\aa}avard Engum}.}
  \bibinfo{year}{2003}\natexlab{b}.
\newblock \showarticletitle{Incremental and decremental proximal support vector
  classification using decay coefficients}. In
  \bibinfo{booktitle}{\emph{DaWaK}}. \bibinfo{pages}{422--429}.
\newblock


\bibitem[\protect\citeauthoryear{Ullah, Mai, Rao, Rossi, and Arora}{Ullah
  et~al\mbox{.}}{2021}]%
        {ullah2021machine}
\bibfield{author}{\bibinfo{person}{Enayat Ullah}, \bibinfo{person}{Tung Mai},
  \bibinfo{person}{Anup Rao}, \bibinfo{person}{Ryan~A Rossi}, {and}
  \bibinfo{person}{Raman Arora}.} \bibinfo{year}{2021}\natexlab{}.
\newblock \showarticletitle{Machine unlearning via algorithmic stability}. In
  \bibinfo{booktitle}{\emph{Conference on Learning Theory}}.
  \bibinfo{pages}{4126--4142}.
\newblock


\bibitem[\protect\citeauthoryear{Veale, Binns, and Edwards}{Veale
  et~al\mbox{.}}{2018}]%
        {veale2018algorithms}
\bibfield{author}{\bibinfo{person}{Michael Veale}, \bibinfo{person}{Reuben
  Binns}, {and} \bibinfo{person}{Lilian Edwards}.}
  \bibinfo{year}{2018}\natexlab{}.
\newblock \showarticletitle{Algorithms that remember: model inversion attacks
  and data protection law}.
\newblock \bibinfo{journal}{\emph{Philos. Trans. R. Soc. A}}
  \bibinfo{volume}{376}, \bibinfo{number}{2133} (\bibinfo{year}{2018}),
  \bibinfo{pages}{20180083}.
\newblock


\bibitem[\protect\citeauthoryear{Verd{\'u}}{Verd{\'u}}{2014}]%
        {verdu2014total}
\bibfield{author}{\bibinfo{person}{Sergio Verd{\'u}}.}
  \bibinfo{year}{2014}\natexlab{}.
\newblock \showarticletitle{Total variation distance and the distribution of
  relative information}. In \bibinfo{booktitle}{\emph{ITA}}.
  \bibinfo{pages}{1--3}.
\newblock


\bibitem[\protect\citeauthoryear{Villaronga, Kieseberg, and Li}{Villaronga
  et~al\mbox{.}}{2018}]%
        {villaronga2018humans}
\bibfield{author}{\bibinfo{person}{Eduard~Fosch Villaronga},
  \bibinfo{person}{Peter Kieseberg}, {and} \bibinfo{person}{Tiffany Li}.}
  \bibinfo{year}{2018}\natexlab{}.
\newblock \showarticletitle{Humans forget, machines remember: Artificial
  intelligence and the right to be forgotten}.
\newblock \bibinfo{journal}{\emph{Computer Law \& Security Review}}
  \bibinfo{volume}{34}, \bibinfo{number}{2} (\bibinfo{year}{2018}),
  \bibinfo{pages}{304--313}.
\newblock


\bibitem[\protect\citeauthoryear{Voigt and Von~dem Bussche}{Voigt and Von~dem
  Bussche}{2017}]%
        {voigt2017eu}
\bibfield{author}{\bibinfo{person}{Paul Voigt} {and} \bibinfo{person}{Axel
  Von~dem Bussche}.} \bibinfo{year}{2017}\natexlab{}.
\newblock \showarticletitle{The eu general data protection regulation (gdpr)}.
\newblock \bibinfo{journal}{\emph{A Practical Guide, 1st Ed., Cham: Springer
  International Publishing}} \bibinfo{volume}{10}, \bibinfo{number}{3152676}
  (\bibinfo{year}{2017}), \bibinfo{pages}{10--5555}.
\newblock


\bibitem[\protect\citeauthoryear{Wang, Yao, Shan, Li, Viswanath, Zheng, and
  Zhao}{Wang et~al\mbox{.}}{2019}]%
        {wang2019neural}
\bibfield{author}{\bibinfo{person}{Bolun Wang}, \bibinfo{person}{Yuanshun Yao},
  \bibinfo{person}{Shawn Shan}, \bibinfo{person}{Huiying Li},
  \bibinfo{person}{Bimal Viswanath}, \bibinfo{person}{Haitao Zheng}, {and}
  \bibinfo{person}{Ben~Y Zhao}.} \bibinfo{year}{2019}\natexlab{}.
\newblock \showarticletitle{Neural cleanse: Identifying and mitigating backdoor
  attacks in neural networks}. In \bibinfo{booktitle}{\emph{SP}}.
  \bibinfo{pages}{707--723}.
\newblock


\bibitem[\protect\citeauthoryear{Wang and Schelter}{Wang and Schelter}{2022}]%
        {wang2022efficiently}
\bibfield{author}{\bibinfo{person}{Benjamin~Longxiang Wang} {and}
  \bibinfo{person}{Sebastian Schelter}.} \bibinfo{year}{2022}\natexlab{}.
\newblock \showarticletitle{Efficiently Maintaining Next Basket Recommendations
  under Additions and Deletions of Baskets and Items}.
\newblock \bibinfo{journal}{\emph{arXiv preprint arXiv:2201.13313}}
  (\bibinfo{year}{2022}).
\newblock


\bibitem[\protect\citeauthoryear{Wang, Huai, and Wang}{Wang
  et~al\mbox{.}}{2023b}]%
        {wang2023inductive}
\bibfield{author}{\bibinfo{person}{Cheng-Long Wang}, \bibinfo{person}{Mengdi
  Huai}, {and} \bibinfo{person}{Di Wang}.} \bibinfo{year}{2023}\natexlab{b}.
\newblock \showarticletitle{Inductive graph unlearning}. In
  \bibinfo{booktitle}{\emph{USENIX}}. \bibinfo{pages}{3205--3222}.
\newblock


\bibitem[\protect\citeauthoryear{Wang, Lin, Chen, Yang, Tang, Zhang, and
  Yu}{Wang et~al\mbox{.}}{2024b}]%
        {wang2024towards}
\bibfield{author}{\bibinfo{person}{Hangyu Wang}, \bibinfo{person}{Jianghao
  Lin}, \bibinfo{person}{Bo Chen}, \bibinfo{person}{Yang Yang},
  \bibinfo{person}{Ruiming Tang}, \bibinfo{person}{Weinan Zhang}, {and}
  \bibinfo{person}{Yong Yu}.} \bibinfo{year}{2024}\natexlab{b}.
\newblock \showarticletitle{Towards Efficient and Effective Unlearning of Large
  Language Models for Recommendation}.
\newblock \bibinfo{journal}{\emph{arXiv preprint arXiv:2403.03536}}
  (\bibinfo{year}{2024}).
\newblock


\bibitem[\protect\citeauthoryear{Wang, Guo, et~al\mbox{.}}{Wang
  et~al\mbox{.}}{2022}]%
        {wang2022federated}
\bibfield{author}{\bibinfo{person}{Junxiao Wang}, \bibinfo{person}{Song Guo},
  {et~al\mbox{.}}} \bibinfo{year}{2022}\natexlab{}.
\newblock \showarticletitle{Federated unlearning via class-discriminative
  pruning}. In \bibinfo{booktitle}{\emph{WWW}}. \bibinfo{pages}{622--632}.
\newblock


\bibitem[\protect\citeauthoryear{Wang, Chen, Yuan, Zeng, Wong, and Yin}{Wang
  et~al\mbox{.}}{2023a}]%
        {wang2023kga}
\bibfield{author}{\bibinfo{person}{Lingzhi Wang}, \bibinfo{person}{Tong Chen},
  \bibinfo{person}{Wei Yuan}, \bibinfo{person}{Xingshan Zeng},
  \bibinfo{person}{Kam-Fai Wong}, {and} \bibinfo{person}{Hongzhi Yin}.}
  \bibinfo{year}{2023}\natexlab{a}.
\newblock \showarticletitle{{KGA}: A General Machine Unlearning Framework Based
  on Knowledge Gap Alignment}. In \bibinfo{booktitle}{\emph{ACL}}.
  \bibinfo{pages}{13264--13276}.
\newblock


\bibitem[\protect\citeauthoryear{Wang, Han, Yang, Zhu, Liu, and Sugiyama}{Wang
  et~al\mbox{.}}{2024a}]%
        {wang2024unlearning}
\bibfield{author}{\bibinfo{person}{Qizhou Wang}, \bibinfo{person}{Bo Han},
  \bibinfo{person}{Puning Yang}, \bibinfo{person}{Jianing Zhu},
  \bibinfo{person}{Tongliang Liu}, {and} \bibinfo{person}{Masashi Sugiyama}.}
  \bibinfo{year}{2024}\natexlab{a}.
\newblock \showarticletitle{Unlearning with Control: Assessing Real-world
  Utility for Large Language Model Unlearning}.
\newblock \bibinfo{journal}{\emph{arXiv preprint arXiv:2406.09179}}
  (\bibinfo{year}{2024}).
\newblock


\bibitem[\protect\citeauthoryear{Wang, Li, Wang, Tang, and Zhou}{Wang
  et~al\mbox{.}}{2009}]%
        {wang2009learning}
\bibfield{author}{\bibinfo{person}{Rui Wang}, \bibinfo{person}{Yong~Fuga Li},
  \bibinfo{person}{XiaoFeng Wang}, \bibinfo{person}{Haixu Tang}, {and}
  \bibinfo{person}{Xiaoyong Zhou}.} \bibinfo{year}{2009}\natexlab{}.
\newblock \showarticletitle{Learning your identity and disease from research
  papers: information leaks in genome wide association study}. In
  \bibinfo{booktitle}{\emph{CCS}}. \bibinfo{pages}{534--544}.
\newblock


\bibitem[\protect\citeauthoryear{Wang, Qinami, Karakozis, Genova, Nair, Hata,
  and Russakovsky}{Wang et~al\mbox{.}}{2020}]%
        {wang2020towards}
\bibfield{author}{\bibinfo{person}{Zeyu Wang}, \bibinfo{person}{Klint Qinami},
  \bibinfo{person}{Ioannis~Christos Karakozis}, \bibinfo{person}{Kyle Genova},
  \bibinfo{person}{Prem Nair}, \bibinfo{person}{Kenji Hata}, {and}
  \bibinfo{person}{Olga Russakovsky}.} \bibinfo{year}{2020}\natexlab{}.
\newblock \showarticletitle{Towards fairness in visual recognition: Effective
  strategies for bias mitigation}. In \bibinfo{booktitle}{\emph{CVPR}}.
  \bibinfo{pages}{8919--8928}.
\newblock


\bibitem[\protect\citeauthoryear{Warnecke, Pirch, Wressnegger, and
  Rieck}{Warnecke et~al\mbox{.}}{2021}]%
        {warnecke2021machine}
\bibfield{author}{\bibinfo{person}{Alexander Warnecke}, \bibinfo{person}{Lukas
  Pirch}, \bibinfo{person}{Christian Wressnegger}, {and}
  \bibinfo{person}{Konrad Rieck}.} \bibinfo{year}{2021}\natexlab{}.
\newblock \showarticletitle{Machine Unlearning of Features and Labels}.
\newblock \bibinfo{journal}{\emph{arXiv preprint arXiv:2108.11577}}
  (\bibinfo{year}{2021}).
\newblock


\bibitem[\protect\citeauthoryear{Wu, Zhu, and Mitra}{Wu et~al\mbox{.}}{2022c}]%
        {wu2022federated}
\bibfield{author}{\bibinfo{person}{Chen Wu}, \bibinfo{person}{Sencun Zhu},
  {and} \bibinfo{person}{Prasenjit Mitra}.} \bibinfo{year}{2022}\natexlab{c}.
\newblock \showarticletitle{Federated unlearning with knowledge distillation}.
\newblock \bibinfo{journal}{\emph{arXiv preprint arXiv:2201.09441}}
  (\bibinfo{year}{2022}).
\newblock


\bibitem[\protect\citeauthoryear{Wu, Souza, Zhang, Fifty, Yu, and
  Weinberger}{Wu et~al\mbox{.}}{2019}]%
        {wu2019simplifying}
\bibfield{author}{\bibinfo{person}{Felix Wu}, \bibinfo{person}{Amauri Souza},
  \bibinfo{person}{Tianyi Zhang}, \bibinfo{person}{Christopher Fifty},
  \bibinfo{person}{Tao Yu}, {and} \bibinfo{person}{Kilian Weinberger}.}
  \bibinfo{year}{2019}\natexlab{}.
\newblock \showarticletitle{Simplifying graph convolutional networks}. In
  \bibinfo{booktitle}{\emph{ICML}}. \bibinfo{pages}{6861--6871}.
\newblock


\bibitem[\protect\citeauthoryear{Wu, Hashemi, and Srinivasa}{Wu
  et~al\mbox{.}}{2022b}]%
        {wu2022puma}
\bibfield{author}{\bibinfo{person}{Ga Wu}, \bibinfo{person}{Masoud Hashemi},
  {and} \bibinfo{person}{Christopher Srinivasa}.}
  \bibinfo{year}{2022}\natexlab{b}.
\newblock \showarticletitle{PUMA: Performance Unchanged Model Augmentation for
  Training Data Removal}. In \bibinfo{booktitle}{\emph{AAAI}}.
\newblock


\bibitem[\protect\citeauthoryear{Wu, Guo, Wang, Hong, Zhang, and Ding}{Wu
  et~al\mbox{.}}{2022a}]%
        {9964015}
\bibfield{author}{\bibinfo{person}{Leijie Wu}, \bibinfo{person}{Song Guo},
  \bibinfo{person}{Junxiao Wang}, \bibinfo{person}{Zicong Hong},
  \bibinfo{person}{Jie Zhang}, {and} \bibinfo{person}{Yaohong Ding}.}
  \bibinfo{year}{2022}\natexlab{a}.
\newblock \showarticletitle{Federated Unlearning: Guarantee the Right of
  Clients to Forget}.
\newblock \bibinfo{journal}{\emph{IEEE Network}} \bibinfo{volume}{36},
  \bibinfo{number}{5} (\bibinfo{year}{2022}), \bibinfo{pages}{129--135}.
\newblock


\bibitem[\protect\citeauthoryear{Wu et~al\mbox{.}}{Wu et~al\mbox{.}}{2020a}]%
        {wu2020deltagrad}
\bibfield{author}{\bibinfo{person}{Yinjun Wu} {et~al\mbox{.}}}
  \bibinfo{year}{2020}\natexlab{a}.
\newblock \showarticletitle{Deltagrad: Rapid retraining of machine learning
  models}. In \bibinfo{booktitle}{\emph{ICML}}. \bibinfo{pages}{10355--10366}.
\newblock


\bibitem[\protect\citeauthoryear{Wu, Tannen, and Davidson}{Wu
  et~al\mbox{.}}{2020b}]%
        {wu2020priu}
\bibfield{author}{\bibinfo{person}{Yinjun Wu}, \bibinfo{person}{Val Tannen},
  {and} \bibinfo{person}{Susan~B Davidson}.} \bibinfo{year}{2020}\natexlab{b}.
\newblock \showarticletitle{Priu: A provenance-based approach for incrementally
  updating regression models}. In \bibinfo{booktitle}{\emph{SIGMOD}}.
  \bibinfo{pages}{447--462}.
\newblock


\bibitem[\protect\citeauthoryear{Wu, Zhu, Li, and He}{Wu et~al\mbox{.}}{2023}]%
        {wu2023deltaboost}
\bibfield{author}{\bibinfo{person}{Zhaomin Wu}, \bibinfo{person}{Junhui Zhu},
  \bibinfo{person}{Qinbin Li}, {and} \bibinfo{person}{Bingsheng He}.}
  \bibinfo{year}{2023}\natexlab{}.
\newblock \showarticletitle{Deltaboost: Gradient boosting decision trees with
  efficient machine unlearning}.
\newblock \bibinfo{journal}{\emph{SIGMOD}} \bibinfo{volume}{1},
  \bibinfo{number}{2} (\bibinfo{year}{2023}), \bibinfo{pages}{1--26}.
\newblock


\bibitem[\protect\citeauthoryear{Xu, Zhu, Zhang, Zhou, and Yu}{Xu
  et~al\mbox{.}}{2023}]%
        {heng2023unlearningsurvey}
\bibfield{author}{\bibinfo{person}{Heng Xu}, \bibinfo{person}{Tianqing Zhu},
  \bibinfo{person}{Lefeng Zhang}, \bibinfo{person}{Wanlei Zhou}, {and}
  \bibinfo{person}{Philip~S. Yu}.} \bibinfo{year}{2023}\natexlab{}.
\newblock \showarticletitle{Machine Unlearning: A Survey}.
\newblock \bibinfo{journal}{\emph{CSUR}} \bibinfo{volume}{56},
  \bibinfo{number}{1} (\bibinfo{year}{2023}), 36.
\newblock


\bibitem[\protect\citeauthoryear{Xu, Wu, Wang, and Jia}{Xu
  et~al\mbox{.}}{2024}]%
        {xu2024machine}
\bibfield{author}{\bibinfo{person}{Jie Xu}, \bibinfo{person}{Zihan Wu},
  \bibinfo{person}{Cong Wang}, {and} \bibinfo{person}{Xiaohua Jia}.}
  \bibinfo{year}{2024}\natexlab{}.
\newblock \showarticletitle{Machine unlearning: Solutions and challenges}.
\newblock \bibinfo{journal}{\emph{TETCI}} (\bibinfo{year}{2024}).
\newblock


\bibitem[\protect\citeauthoryear{Yamashita, Yamada, and Shibata}{Yamashita
  et~al\mbox{.}}{2023}]%
        {yamashita2023one}
\bibfield{author}{\bibinfo{person}{Tomoya Yamashita}, \bibinfo{person}{Masanori
  Yamada}, {and} \bibinfo{person}{Takashi Shibata}.}
  \bibinfo{year}{2023}\natexlab{}.
\newblock \showarticletitle{One-Shot Machine Unlearning with Mnemonic Code}.
\newblock \bibinfo{journal}{\emph{arXiv preprint arXiv:2306.05670}}
  (\bibinfo{year}{2023}).
\newblock


\bibitem[\protect\citeauthoryear{Yoon, Nam, Yun, Kim, and Ok}{Yoon
  et~al\mbox{.}}{2022}]%
        {yoon2022few}
\bibfield{author}{\bibinfo{person}{Youngsik Yoon}, \bibinfo{person}{Jinhwan
  Nam}, \bibinfo{person}{Hyojeong Yun}, \bibinfo{person}{Dongwoo Kim}, {and}
  \bibinfo{person}{Jungseul Ok}.} \bibinfo{year}{2022}\natexlab{}.
\newblock \showarticletitle{Few-Shot Unlearning by Model Inversion}.
\newblock \bibinfo{journal}{\emph{arXiv preprint arXiv:2205.15567}}
  (\bibinfo{year}{2022}).
\newblock


\bibitem[\protect\citeauthoryear{Yu, Zhang, Chen, Yin, and Liu}{Yu
  et~al\mbox{.}}{2021}]%
        {yu2021does}
\bibfield{author}{\bibinfo{person}{Da Yu}, \bibinfo{person}{Huishuai Zhang},
  \bibinfo{person}{Wei Chen}, \bibinfo{person}{Jian Yin}, {and}
  \bibinfo{person}{Tie-Yan Liu}.} \bibinfo{year}{2021}\natexlab{}.
\newblock \showarticletitle{How does data augmentation affect privacy in
  machine learning?}. In \bibinfo{booktitle}{\emph{AAAI}},
  Vol.~\bibinfo{volume}{35}. \bibinfo{pages}{10746--10753}.
\newblock


\bibitem[\protect\citeauthoryear{Yu, Seff, Zhang, Song, Funkhouser, and
  Xiao}{Yu et~al\mbox{.}}{2015}]%
        {yu2015lsun}
\bibfield{author}{\bibinfo{person}{Fisher Yu}, \bibinfo{person}{Ari Seff},
  \bibinfo{person}{Yinda Zhang}, \bibinfo{person}{Shuran Song},
  \bibinfo{person}{Thomas Funkhouser}, {and} \bibinfo{person}{Jianxiong Xiao}.}
  \bibinfo{year}{2015}\natexlab{}.
\newblock \showarticletitle{Lsun: Construction of a large-scale image dataset
  using deep learning with humans in the loop}.
\newblock \bibinfo{journal}{\emph{arXiv preprint arXiv:1506.03365}}
  (\bibinfo{year}{2015}).
\newblock


\bibitem[\protect\citeauthoryear{Yuan, Yin, Wu, Zhang, He, and Wang}{Yuan
  et~al\mbox{.}}{2023}]%
        {yuan2023federated}
\bibfield{author}{\bibinfo{person}{Wei Yuan}, \bibinfo{person}{Hongzhi Yin},
  \bibinfo{person}{Fangzhao Wu}, \bibinfo{person}{Shijie Zhang},
  \bibinfo{person}{Tieke He}, {and} \bibinfo{person}{Hao Wang}.}
  \bibinfo{year}{2023}\natexlab{}.
\newblock \showarticletitle{Federated unlearning for on-device recommendation}.
  In \bibinfo{booktitle}{\emph{WSDM}}. \bibinfo{pages}{393--401}.
\newblock


\bibitem[\protect\citeauthoryear{Zanella-B{\'e}guelin, Wutschitz, Tople,
  R{\"u}hle, Paverd, Ohrimenko, et~al\mbox{.}}{Zanella-B{\'e}guelin
  et~al\mbox{.}}{2020}]%
        {zanella2020analyzing}
\bibfield{author}{\bibinfo{person}{Santiago Zanella-B{\'e}guelin},
  \bibinfo{person}{Lukas Wutschitz}, \bibinfo{person}{Shruti Tople},
  \bibinfo{person}{Victor R{\"u}hle}, \bibinfo{person}{Andrew Paverd},
  \bibinfo{person}{Olga Ohrimenko}, {et~al\mbox{.}}}
  \bibinfo{year}{2020}\natexlab{}.
\newblock \showarticletitle{Analyzing information leakage of updates to natural
  language models}. In \bibinfo{booktitle}{\emph{SIGSAC}}.
  \bibinfo{pages}{363--375}.
\newblock


\bibitem[\protect\citeauthoryear{Zeng, Wang, Chen, Just, Jin, and Jia}{Zeng
  et~al\mbox{.}}{2021}]%
        {zeng2021learning}
\bibfield{author}{\bibinfo{person}{Yingyan Zeng}, \bibinfo{person}{Tianhao
  Wang}, \bibinfo{person}{Si Chen}, \bibinfo{person}{Hoang~Anh Just},
  \bibinfo{person}{Ran Jin}, {and} \bibinfo{person}{Ruoxi Jia}.}
  \bibinfo{year}{2021}\natexlab{}.
\newblock \showarticletitle{Learning to Refit for Convex Learning Problems}.
\newblock \bibinfo{journal}{\emph{arXiv preprint arXiv:2111.12545}}
  (\bibinfo{year}{2021}).
\newblock


\bibitem[\protect\citeauthoryear{Zhang, Chen, Cong, Guo, Liu, and Zhou}{Zhang
  et~al\mbox{.}}{2020}]%
        {zhang2020deep}
\bibfield{author}{\bibinfo{person}{Hao Zhang}, \bibinfo{person}{Bo Chen},
  \bibinfo{person}{Yulai Cong}, \bibinfo{person}{Dandan Guo},
  \bibinfo{person}{Hongwei Liu}, {and} \bibinfo{person}{Mingyuan Zhou}.}
  \bibinfo{year}{2020}\natexlab{}.
\newblock \showarticletitle{Deep autoencoding topic model with scalable hybrid
  Bayesian inference}.
\newblock \bibinfo{journal}{\emph{TPAMI}} \bibinfo{volume}{43},
  \bibinfo{number}{12} (\bibinfo{year}{2020}), \bibinfo{pages}{4306--4322}.
\newblock


\bibitem[\protect\citeauthoryear{Zhang, Bai, Huang, and Xu}{Zhang
  et~al\mbox{.}}{2022}]%
        {zhang2022machine}
\bibfield{author}{\bibinfo{person}{Peng-Fei Zhang}, \bibinfo{person}{Guangdong
  Bai}, \bibinfo{person}{Zi Huang}, {and} \bibinfo{person}{Xin-Shun Xu}.}
  \bibinfo{year}{2022}\natexlab{}.
\newblock \showarticletitle{Machine Unlearning for Image Retrieval: A
  Generative Scrubbing Approach}. In \bibinfo{booktitle}{\emph{MM}}.
  \bibinfo{pages}{237--245}.
\newblock


\bibitem[\protect\citeauthoryear{Zhang, Wang, Cheng, Sun, Zhang, and
  Xiao}{Zhang et~al\mbox{.}}{2023}]%
        {zhang2023machine}
\bibfield{author}{\bibinfo{person}{Xulong Zhang}, \bibinfo{person}{Jianzong
  Wang}, \bibinfo{person}{Ning Cheng}, \bibinfo{person}{Yifu Sun},
  \bibinfo{person}{Chuanyao Zhang}, {and} \bibinfo{person}{Jing Xiao}.}
  \bibinfo{year}{2023}\natexlab{}.
\newblock \showarticletitle{Machine unlearning methodology based on stochastic
  teacher network}. In \bibinfo{booktitle}{\emph{ADMA}}.
  \bibinfo{pages}{250--261}.
\newblock


\bibitem[\protect\citeauthoryear{Zhang, Chen, Jia, Zhang, Fan, Liu, Hong, Ding,
  and Liu}{Zhang et~al\mbox{.}}{2024a}]%
        {zhang2024defensive}
\bibfield{author}{\bibinfo{person}{Yimeng Zhang}, \bibinfo{person}{Xin Chen},
  \bibinfo{person}{Jinghan Jia}, \bibinfo{person}{Yihua Zhang},
  \bibinfo{person}{Chongyu Fan}, \bibinfo{person}{Jiancheng Liu},
  \bibinfo{person}{Mingyi Hong}, \bibinfo{person}{Ke Ding}, {and}
  \bibinfo{person}{Sijia Liu}.} \bibinfo{year}{2024}\natexlab{a}.
\newblock \showarticletitle{Defensive Unlearning with Adversarial Training for
  Robust Concept Erasure in Diffusion Models}.
\newblock \bibinfo{journal}{\emph{arXiv preprint arXiv:2405.15234}}
  (\bibinfo{year}{2024}).
\newblock


\bibitem[\protect\citeauthoryear{Zhang, Zhang, Yao, Jia, Liu, Liu, and
  Liu}{Zhang et~al\mbox{.}}{2024b}]%
        {zhang2024unlearncanvas}
\bibfield{author}{\bibinfo{person}{Yihua Zhang}, \bibinfo{person}{Yimeng
  Zhang}, \bibinfo{person}{Yuguang Yao}, \bibinfo{person}{Jinghan Jia},
  \bibinfo{person}{Jiancheng Liu}, \bibinfo{person}{Xiaoming Liu}, {and}
  \bibinfo{person}{Sijia Liu}.} \bibinfo{year}{2024}\natexlab{b}.
\newblock \showarticletitle{Unlearncanvas: A stylized image dataset to
  benchmark machine unlearning for diffusion models}.
\newblock \bibinfo{journal}{\emph{arXiv preprint arXiv:2402.11846}}
  (\bibinfo{year}{2024}).
\newblock


\bibitem[\protect\citeauthoryear{Zhu, Han, Yao, Xu, Niu, and Sugiyama}{Zhu
  et~al\mbox{.}}{2024}]%
        {zhu2024decoupling}
\bibfield{author}{\bibinfo{person}{Jianing Zhu}, \bibinfo{person}{Bo Han},
  \bibinfo{person}{Jiangchao Yao}, \bibinfo{person}{Jianliang Xu},
  \bibinfo{person}{Gang Niu}, {and} \bibinfo{person}{Masashi Sugiyama}.}
  \bibinfo{year}{2024}\natexlab{}.
\newblock \showarticletitle{Decoupling the Class Label and the Target Concept
  in Machine Unlearning}.
\newblock \bibinfo{journal}{\emph{arXiv preprint arXiv:2406.08288}}
  (\bibinfo{year}{2024}).
\newblock


\bibitem[\protect\citeauthoryear{Zhu, Li, and Hu}{Zhu et~al\mbox{.}}{2023}]%
        {zhu2023heterogeneous}
\bibfield{author}{\bibinfo{person}{Xiangrong Zhu}, \bibinfo{person}{Guangyao
  Li}, {and} \bibinfo{person}{Wei Hu}.} \bibinfo{year}{2023}\natexlab{}.
\newblock \showarticletitle{Heterogeneous federated knowledge graph embedding
  learning and unlearning}. In \bibinfo{booktitle}{\emph{WWW}}.
  \bibinfo{pages}{2444--2454}.
\newblock


\bibitem[\protect\citeauthoryear{Zou and Schiebinger}{Zou and
  Schiebinger}{2018}]%
        {zou2018ai}
\bibfield{author}{\bibinfo{person}{James Zou} {and} \bibinfo{person}{Londa
  Schiebinger}.} \bibinfo{year}{2018}\natexlab{}.
\newblock \bibinfo{title}{AI can be sexist and racist -- it's time to make it
  fair}.
\newblock
\newblock


\end{thebibliography}

%\printbibliography

%\appendix

\end{document}